\PassOptionsToPackage{dvipsnames}{xcolor}

\documentclass{article} 
\usepackage[utf8]{inputenc}
\usepackage[T1]{fontenc}   

\usepackage[normalem]{ulem}
\usepackage{booktabs}
\usepackage{pifont}
\usepackage{natbib}
\usepackage{xcolor}
\usepackage{algorithmic}
\usepackage{algorithm}
\usepackage{caption}
\usepackage{footnote}

\newlength{\defbaselineskip}
\newcommand{\xmark}{\ding{55}}
\makeatletter
\newcommand{\printfnsymbol}[1]{%
  \textsuperscript{\@fnsymbol{#1}}%
}
\makeatother

\usepackage{amsmath,amsfonts,bm}

\def\eqref#1{equation~\ref{#1}}

\def\1{\bm{1}}

\DeclareMathAlphabet{\mathsfit}{\encodingdefault}{\sfdefault}{m}{sl}
\SetMathAlphabet{\mathsfit}{bold}{\encodingdefault}{\sfdefault}{bx}{n}

\DeclareMathOperator*{\argmax}{arg\,max}

\newtheorem{lemma}{Lemma}

\usepackage{hyperref}
\usepackage{url}
\usepackage{graphicx}
\usepackage{thm-restate}
\usepackage{multirow}
\usepackage{subfigure}
\usepackage{lineno}
\usepackage{mathtools}
\usepackage{caption}
\usepackage[export]{adjustbox}
\usepackage{amssymb}
\usepackage{authblk}

\usepackage{enumitem}

\newcommand{\figsize}{0.7}

\newcommand{\figsizeapp}{0.9}
\newcommand{\figsizepme}{0.8}
\newcommand{\figsizeadvection}{0.7}
\newcommand{\tabsize}{0.7}

\allowdisplaybreaks

\newcommand{\probconservnosp}{\textsc{ProbConserv}} 
\newcommand{\probconserv}{\textsc{ProbConserv}~}
\newcommand{\physnpnosp}{\textsc{ProbConserv-ANP}} 
\newcommand{\physnp}{\textsc{ProbConserv-ANP}~}
\newcommand{\ANP}{\textsc{ANP}~}
\newcommand{\ANPnosp}{\textsc{ANP}}
\newcommand{\HCANP}{\textsc{HardC-ANP}~}
\newcommand{\HCANPnosp}{\textsc{HardC-ANP}}
\newcommand{\SCANP}{\textsc{SoftC-ANP}~}
\newcommand{\SCANPnosp}{\textsc{SoftC-ANP}}

\newcommand{\distnorm}{\mathcal{N}} 
\newcommand{\dnorm}{p_{\distnorm}}
\providecommand{\keywords}[1]{\textit{Keywords:} #1}
\begin{document}
\title{Learning Physical Models that Can Respect Conservation Laws}

\author[a]{Derek Hansen\footnote{Equal contribution.}\thanks{Work completed during an internship at AWS AI Labs.}}
\author[b]{Danielle C. Maddix\printfnsymbol{1}\footnote{Correspondence to:
Danielle C. Maddix <dmmaddix@amazon.com>.}}
\author[b]{Shima Alizadeh}
\author[b]{Gaurav Gupta}
\author[c]{Michael W. Mahoney}
\affil[a]{Dept. of Statistics, Univ. of Michigan (1085 S University Ave., Ann Arbor, MI 48109, US)}
\affil[b]{AWS AI Labs (2795 Augustine Dr, Santa Clara, CA 95054, US)}
\affil[c]{Amazon Supply Chain Optimization Technologies (7 West 34th St., NY, NY 10001, US)}
\affil[ ]{{\texttt{dereklh@umich.edu}, \  \{\texttt{dmmaddix, alizshim, gauravaz, zmahmich}\}\texttt{@amazon.com}}}
\date{}

\maketitle

\begin{abstract}
Recent work in scientific machine learning (SciML) has focused on incorporating partial differential equation (PDE) information into the learning process. 
Much of this work has focused on relatively ``easy'' PDE operators (e.g., elliptic and parabolic), with less emphasis on relatively ``hard'' PDE operators (e.g., hyperbolic). Within numerical PDEs, the latter problem class requires control of a type of volume element or conservation constraint, which is known to be challenging. Delivering on the promise of SciML requires seamlessly incorporating both types of problems into the learning process. To address this issue, we propose \probconservnosp, a framework for incorporating conservation constraints into a generic SciML architecture. To do so, \probconserv combines the integral form of a conservation law with a Bayesian update. We provide a detailed analysis of \probconserv on learning with the Generalized Porous Medium Equation (GPME), a widely-applicable parameterized family of PDEs that illustrates the qualitative properties of both easier and harder PDEs. \probconserv is effective for easy GPME variants, performing well with state-of-the-art competitors; and for harder GPME variants it outperforms other approaches that do not guarantee volume conservation. \probconserv  seamlessly enforces physical conservation constraints, maintains probabilistic uncertainty quantification (UQ), and deals well with shocks and heteroscedasticities. In each case, it achieves superior predictive performance on downstream~tasks.
\end{abstract}


\section{Introduction}

Conservation laws are ubiquitous in science and engineering, where they are used to model physical phenomena ranging from heat transfer to wave propagation to fluid flow dynamics, and beyond.  
These laws can be expressed in two complementary ways: 
in a differential form; or in an integral form.
They are  most commonly expressed as partial differential equations (PDEs) in a differential form, 
$$
u_t + \nabla \cdot F(u) = 0 ,
$$
for an unknown $u$ and a nonlinear flux function $F(u)$. 
This differential form of the conservation law can be integrated over a spatial domain $\Omega$ using the divergence theorem to result in an integral form of the conservation law, 
$$
U_t= -\int_{\Gamma} F(u) \cdot n d\Gamma, 
$$
where $U = \int_{\Omega} u(t,x)d\Omega$, $\Gamma$ denotes the boundary of $\Omega$ and $n$ denotes the outward unit normal vector. 
As examples: 
in the case of heat transfer, $u$ denotes the temperature, and $U$ the conserved energy of system; and 
in the case of porous media flow, $u$ denotes the density, and $U$ the conserved mass of the porous media.

Global conservation states that the rate of change in time of the conserved quantity $U$  over a domain $\Omega$ is given by 
the 
flux across the boundary $\Gamma$ of the domain. 
Local conservation arises naturally in the numerical solution of PDEs.
Traditional numerical methods (e.g., finite differences, finite elements, and finite volume methods) have been developed to solve PDEs numerically, with finite volume methods being designed for (and being particularly well-suited for) conservation laws 
\citep{leveque1990numerical, leveque2002, leveque}.  
Finite volume methods divide the domain $\Omega$ into control volumes and apply the integral form locally. 
They enforce that the time derivative of the cell-averaged unknown is equal to the difference between the in-flux and out-flux over the control volume.
(This local conservation---so-called since the out-flux that leaves one cell equals the in-flux that enters a neighboring cell---can be used to guarantee global conservation over the whole domain.) 
This numerical approach should be contrasted with finite difference methods, which use the differential form directly, and which are thus not guaranteed to satisfy the conservation condition.

This discussion is relevant for machine learning (ML) since there has been an interest recently in Scientific ML (SciML) in incorporating the physical knowledge or physical constraints into neural network (NN) training. 
A popular example of this is the so-called Physics-Informed Neural Networks (PINNs) \citep{Raissi19}.
This approach uses a NN to approximate the PDE solution by incorporating the differential form of the PDE into the loss function, basically as a soft constraint or regularization term. 
Other data-driven approaches, including DeepONet \citep{deeponet} and Neural Operators (NOs) \citep{liFourierNeuralOperator2021b, guptaMultiwaveletbasedOperatorLearning2021}, train on simulations and aim to learn the underlying function map from initial conditions or PDE coefficients to the solution. 
Other methods such as Physics-Informed Neural Operator (PINO) attempt to make the data-driven Fourier Neural Operator (FNO) ``physics-informed,'' again by adding the differential form into the supervised loss function as a soft constraint regularization term \citep{PINO_2021, goswami22}.  

Challenges and limitations for SciML of this soft constraint approach on model training were recently identified~\citep{krishnapriyanCharacterizingPossibleFailure2021b,cacm_edwards22}. The basic issue is that, unlike numerical finite volume methods, these ML and SciML methods do \emph{not} guarantee that the physical property of conservation is satisfied. This is a consequence of the fact that the Lagrange dual form of the constrained optimization problem does not in general satisfy the constraint.
This results in very weak control on the physical conservation property, resulting in non-physical solutions that violate the governing conservation law.  

In this work, we frame the problem of learning physical models that can respect conservation laws via a ``finite-volume lens'' from scientific computing.  
This permits us to use the integral form of the governing conservation law to enforce conservation conditions for a range of SciML problems. 
In particular, 
for a wide range of initial and boundary conditions,
we can express the integral form as a time-varying linear constraint that is compatible with existing ML pipelines. 
This permits us to propose a two-step framework.
In the first step, we use an ML model with a mean and variance estimate to compute a predictive distribution for the solution at specified target points.
Possible methods for this step include:
classic estimation methods (e.g., Gaussian Processes \citep{rasmussen2006}); 
methods designed to exploit the complementary strengths of classical methods and NN methods (e.g., Neural Processes \citep{kimAttentiveNeuralProcesses2019}); as well as 
computing ensembles of NN models (to compute empirical estimates of means and variances). 
In the second step, we apply a discretization of the integral form of the constraint as a Bayesian update in order to enforce the physical conservation constraint on the black-box unconstrained output. 
We illustrate our framework, 
\probconservnosp, by using an Attentive Neural Process (ANP) \citep{kimAttentiveNeuralProcesses2019} as the probabilistic deep learning model in the first step paired with a global conservation constraint in the second step. 
In more detail, the following are our main contributions: 
\begin{itemize}[noitemsep,topsep=0pt] 
\item 
\textit{Integral form for conservation}.
We propose to use the integral form of the governing conservation law via finite volume methods, rather than the commonly used differential form, to enforce conservation subject to a specified noise parameter. 
Through an ablation study, 
    we show that adding the differential form of the PDE as a soft constraint to the loss function does not enforce conservation in the underlying unconstrained ML model.
\item 
\textit{Strong control on the conservation constraint}.
By using the integral form, we are able to enforce conservation via linear probabilistic constraints, which can be made arbitrarily binding or sharp by reducing the variance term $\sigma_G^2$. In particular, by adjusting $\sigma_G^2$, one can balance satisfying conservation with predictive metrics (e.g., MSE), with \probconserv obtaining exact conservation when $\sigma_G^2=0$. 
\item 
\textit{Effective for ``easy'' to ``hard'' PDEs}.
We evaluate on a parametric family of PDEs, which permits us to explore ``easy" parameter regimes  as well as ``medium" and ``hard" parameter regimes.
We find that our method and the baselines do well for ``easy'' problems (although baselines sometimes have issues even with ``easy'' problems, and even for ``easy'' problems their solutions may not be conservative), but we do seamlessly better as we go to ``harder'' problems, with a $5\times$ improvement in MSE. 
\item 
\textit{Uncertainty Quantification (UQ) and downstream tasks}.
We provide theoretical guarantees that \probconserv increases predictive log-likelihood (LL) compared to the original black-box ML model.
 Empirically, we show that \probconserv consistently improves LL, which takes into account both prediction accuracy and well-calibrated uncertainty.  
 On ``hard''  problems, this improved control on uncertainty leads to better insights on downstream shock position detection tasks.
\end{itemize}

There is a large body of related work, too much to summarize here; see \autoref{sec:related_works} for a summary.

\section{A Probabilistic Approach to Conservation Law Enforcement}

In this section, we present our framework, \probconservnosp, for learning physical models that can respect conservation laws.
Our approach centers around the following two sources of information: 
an unconstrained ML algorithm that makes mean and variance predictions; and 
a conservation constraint (in the form of \autoref{eqn:global_conserv_mb} below) that comes from knowledge of the underlying physical system.
See Algorithm~\ref{alg:conserv} for details of our approach.
In the first step, we compute a set of mean and variance estimates for the unconstrained model.
In the second step, we use those mean and variance estimates to compute an update that respects the conservation law. 
The update rule has a natural probabilistic interpretation in terms of uncertainty quantification, and it can be used to satisfy the conservation constraint to a user-specified tolerance level.
As this tolerance goes to zero, our method gracefully converges to a limiting solution that satisfies conservation exactly (see Theorem \ref{thm:constraint_to_zero} below).

\begin{algorithm}[t]
\begin{algorithmic}
  \STATE {\textbf{Input:}} {
    Constraint matrix $G$, constraint value $b$, non-zero noise $\sigma_G$ and input points $(t_1, x_1), \dots (t_N, x_N)$
  }
  \STATE {\textbf{Step 1}: 
  Calculate black-box prediction over output grid: ${\mu, \Sigma = f_\theta((t_1, x_1), \dots (t_N, x_N); D)}$ }
  \STATE {\textbf{Step 2}}: 
  Calculate 
  $\tilde \mu$ and $\tilde \Sigma$ according to \autoref{eqn:updated_mean_var}.
    \STATE {\textbf{Output:}}{
    $\tilde \mu, \tilde \Sigma$
  }
\end{algorithmic}
   \caption{\probconserv}
   \label{alg:conserv}
\end{algorithm}

\subsection{Integral Form of Conservation Laws as a Linear Constraint}
\label{subsec:linear_form}

Here, we first derive the integral form of a governing conservation law from the corresponding differential form (a la finite volume methods), and we then show how this integral form can be expressed as a linear constraint (for PDEs with specific initial and boundary conditions, even for certain nonlinear differential PDE operators) for a broad class of real-world problems.

Consider the differential form of the governing equation:
\begin{equation}
\begin{drcases}
    & \mathcal{F} u(t,x) = 0, \,\, x \in \Omega, \\ 
    & u(0,x) = h(x), \,\, x \in \Omega, \\
    & u(t,x) = g(t,x), \,\, x \in \Gamma,
    \label{eq:gov_eqtn}
\end{drcases}, \forall\,t \geq 0,
\end{equation}
where $\Gamma$ denotes the boundary of the domain $\Omega$, $h(x)$ the initial condition, and $g(t,x)$ the Dirichlet boundary condition. 
Recently popular SciML methods, e.g., PINNs \citep{Raissi19}, PINOs \citep{PINO_2021, goswami22}, focus on incorporating this form of the constraint into the NN training procedure.
In particular, the differential form of the PDE $\mathcal{F}u(t,x)$ could be added as a soft constraint to the loss function $\mathcal{L}$, as follows:
\[
 \min_{\theta} \mathcal{L}(u) + \lambda \|\mathcal{F}u\|,
\]
where $\mathcal{L}$ denotes a loss function measuring the error of the NN approximated solution relative to the known initial and boundary conditions (and potentially any observed solution samples), $\theta$ denotes the NN parameters, and $\lambda$ denotes a penalty or regularization parameter.

For conservation laws, the differential form is given as:
\begin{equation}
    \mathcal{F}u = u_t + \nabla \cdot F(u),
    \label{eq:diff_form}
\end{equation} 
for some given nonlinear flux function $F(u)$. The corresponding integral form of a conservation law is given~as:
\begin{equation}
\begin{aligned} 
 \int_{\Omega} u(t,x)d\Omega = \int_{\Omega} h(x)d\Omega - \int_0^t \int_{\Gamma} F(u) \cdot n d\Gamma dt .
   \end{aligned} 
    \label{eq:int_form}
\end{equation}
See Appendix~\ref{app:differential_to_integral_form} for a derivation.

In one-dimension, the boundary integral of the flux can be computed analytically, as the difference of the flux in and out of the domain: 
\begin{equation}
\begin{aligned}
    \underbrace{\int_{\Omega} u(t,x)d\Omega}_{\mathcal{G}u(t,x)}  = \underbrace{\int_{\Omega} h(x)d\Omega + \int_{0}^{t}
    (F_{\text{in}} - F_{\text{out}})dt}_{b(t)},
     \label{eqn:global_conserv_mb}
\end{aligned}
\end{equation} 
where  $\Omega = [x_0, x_N]$, $F_{\text{in}} = F(u, t, x_0)|_{u=g(t,x_0)}$, and $F_{\text{out}} = F(u, t, x_N)|_{u=g(t,x_N)}$. 
In two and higher dimensions, we do not have an analytic expression, but one can approximate this boundary integral as the sum over the spatial dimensions of the difference of the in and out fluxes on the boundary in that dimension.
This methodology is well-developed within finite volume discretization methods, and
we leave this extension to future~work.

In many applications (including those we consider), by using the prescribed physical boundary condition $u(t,x) = g(t,x)$ for $x \in \Gamma$, it holds that the in and out fluxes on the boundary do \emph{not} depend on $u$, and instead they only depend on $t$. 
This is known as a \emph{boundary flux linearity assumption} since, when it holds, one can use a simple linear constraint to enforce the conservation law. 
This assumption holds for a broad class of problems---even including nonlinear conservation laws with nonlinear PDE operators $\mathcal{F}$ (See \autoref{app:exact_sol} for the initial/boundary conditions, exact solutions, exact linear global conservation constraints and \autoref{tab:linear_conserv} for a summary).
In these cases, \autoref{eqn:global_conserv_mb} results in the following linear constraint equation:
\begin{equation}
    \mathcal{G}u(t,x) 
    = \int_{\Omega} u(t,x)d\Omega 
    = b(t),
    \label{eq:linear_constraint}
\end{equation}
which can be used to enforce global conservation. 
See Appendix~\ref{app:integral_discret} for details on how this integral equation can be discretized into a matrix equation.

In other applications, of course, the flux linearity assumption along the boundary of the domain will not hold.
For example, the flux may not be known and/or the boundary condition may depend on $u(t,x)$.
In these cases, we will not be able to not apply \autoref{eq:linear_constraint} directly.
However, nonlinear least squares methods may still be used to enforce the conservation constraint.
This methodology is also well-developed, and we leave this extension to future~work.

\subsection{Step 1: Unconstrained Probability Distribution}

In Step 1 of \probconservnosp,
we use a supervised black-box ML model to infer the mean $\mu$ and covariance $\Sigma$ of the unknown function $u$ from observed data $D$.
For example, $D$ can include values of the function $u$ observed at a small set of points.
Over a set of
$N$ input points $(t_1, x_1), \dots, (t_N, x_N)$, 
the probability distribution of $u \coloneqq [u(t_1, x_1), \dots u(t_N, x_N)] \in \mathbb{R}^{N}$ conditioned on data $D$ has mean ${\mu \coloneqq \mathbb E(u \vert D)}$ and covariance ${\Sigma \coloneqq \text{Cov}(u \vert D)}$ given by the black-box model $f_\theta$, i.e.,
\begin{equation}
    \mu, \Sigma = f_\theta \left((t_1, x_1), \dots, (t_N, x_N); D \right).
  \label{eqn:unconstrained_output}
\end{equation}
This framework is general, and there are possible choices for the model in \autoref{eqn:unconstrained_output}.
Gaussian Processes \citep{rasmussen2006} are a natural choice, assuming that one has chosen an appropriate mean and kernel function for the specific problem. 
The ANP model \citep{kimAttentiveNeuralProcesses2019}, which uses a transformer architecture to encode the mean and covariance, is another choice.
A third option is to perform repeated runs, e.g., with different initial seeds, of non-probabilistic black-box NN models to compute empirical estimates of mean and variance parameters. 

\subsection{Step 2: Enforcing Conservation Constraint} \label{subsec:prob_constraints}

In Step 2 of \probconservnosp, we incorporate a discretized and probabilistic form of the constraint given in \autoref{eq:linear_constraint}:
\begin{equation} \label{eqn:normal_constraint}
\begin{split}
b &= Gu + \sigma_G \epsilon,
\end{split}
\end{equation}
where $G$ denotes a matrix approximating the linear operator $\mathcal G$ (see Appendix~\ref{app:integral_discret}), $b$ denotes a vector of observed constraint values, and $\epsilon$ denotes a noise term, where each component has unit variance.
The parameter $\sigma_G \ge 0$ controls how much the conservation constraint can be violated (see \autoref{app:noise_param} for details), with $\sigma_G = 0$ enforcing exact adherence.  
Step 2 outputs the following updated mean $\tilde \mu$ and covariance $\tilde \Sigma$ that respect conservation, given as:
\begin{subequations}
\label{eqn:updated_mean_var}
\begin{align}
   \label{eqn:updated_mean_var_MEAN}
   \tilde \mu &= \mu - \Sigma G^T (\sigma_G^2 I + G \Sigma G^T)^{-1} (G\mu - b), \\
   \label{eqn:updated_mean_var_VAR}
   \tilde \Sigma &= \Sigma - \Sigma G^T (\sigma_G^2 I + G \Sigma G^T)^{-1} G \Sigma,
\end{align}
\end{subequations}
 where $\mu$ and $\Sigma$ denote the mean and covariance matrix, respectively, from Step 1 (\autoref{eqn:unconstrained_output}).

The update rule given in \autoref{eqn:updated_mean_var} can be justified from two complementary perspectives.
    From a Bayesian probabilistic perspective, \autoref{eqn:updated_mean_var} is the posterior mean and covariance of the predictive distribution of $u$ after incorporating the information given by the conservation constraint via \autoref{eqn:normal_constraint}.   From an optimization perspective, \autoref{eqn:updated_mean_var} is the solution to a least-squares problem that places a binding inequality constraint on the conserved quantity $G\tilde \mu$ (i.e., $\|G \tilde \mu - b\|_2 \le c$ for some ${c \in (0, \|G\mu - b\|_2)}$).
See \autoref{app:deriv_mean_cov_limit} for more details on these two complementary perspectives.

We emphasize that, for $\sigma_G > 0$, the final solution does not satisfy $G \tilde \mu = b$ exactly.
Adherence to the constraint can be gracefully controlled by shrinking $\sigma_G$. 
Specifically, if we consider a monotonic decreasing sequence of constraint values $\sigma_{G, n} \downarrow 0$, then the corresponding sequence of posterior means $\tilde \mu_n$ is well-behaved, and the limiting solution can be calculated.
This is shown in the following theorem.
\begin{restatable}{thm}{constrainttozero}
\label{thm:constraint_to_zero}
  Let $\mu$ and $\Sigma$ be the mean and covariance of $u$ obtained at the end of Step 1.
  Let $\sigma_{G, n} \downarrow 0$ be a monotonic decreasing sequence of constraint values and let $\tilde \mu_n$ be the corresponding posterior mean at the end of Step 2 shown in \autoref{eqn:updated_mean_var}.
  Then:
  \begin{enumerate}[noitemsep,topsep=0pt]
    \item The sequence $\tilde \mu_{n}$ converges to a limit $\tilde \mu^\star$ monotonically; i.e., $\| \tilde \mu_n - \tilde \mu^\star \|_{\Sigma^{-1}} \downarrow 0$. 
    \item The limiting mean $\tilde \mu^\star$ is the solution to a {constrained} least-squares problem: $\mathrm{argmin}_{y} \|y - \mu\|_{\Sigma^{-1}}$ subject to $Gy = b$.
    \item The sequence $G \tilde \mu_{n}$ converges to $b$ in $L_2$; i.e., $\| G\tilde \mu_n - b \|_{2} \downarrow 0$. 
    \end{enumerate}
    Moreover, if the conservation constraint $Gu = b$ holds exactly for the true solution $u$, then: 
    \begin{enumerate}[noitemsep,topsep=0pt, resume]
  \item The distance between the true solution $u$ and the posterior mean $\tilde \mu_n$ decreases as $\sigma_{G, n} \to 0$, i.e., ${\|\tilde \mu_n - u\|_{\Sigma^{-1}} \downarrow \|\tilde \mu^\star - u\|_{\Sigma^{-1}}}$.
    \item For sufficiently small $\sigma_{G, n}$, the log-likelihood $\text{LL}(u; \tilde \mu_n, \tilde \Sigma_n)$  is greater than $\text{LL}(u; \mu, \Sigma)$ and increases as $\sigma_{G, n} \to 0$.
\end{enumerate}
\end{restatable}

See \autoref{app:limiting_sol} for a proof of Theorem \ref{thm:constraint_to_zero}. 
Importantly, Theorem \ref{thm:constraint_to_zero} holds for any mean and covariance estimates $\mu, \Sigma$, whether they come from a Gaussian Process, ANP, or repeated runs of a black-box NN. It also shows that we are guaranteed to improve in log-likelihood (LL), which we also verify in the empirical results (see \autoref{app:noise_param}).

We should also emphasize that, in addition to conservation, \autoref{eqn:normal_constraint} can incorporate other inductive biases, based on knowledge of the underlying PDE.  To take but one practically-useful example, one typically desires a solution that is free of artificial high-frequency oscillations.
This smoothing can be accomplished by penalizing large absolute values of the second derivative
via a second order central finite difference discretization in the matrix $\tilde{G}$ (see Appendix~\ref{app:artificial_diff}).

\section{Empirical Results} \label{sec:experiments}
\begin{figure*}[h]
    \centering
    \vskip 0.1in
    \subfigure[easy: Diffusion equation ($k=1$)]{\label{subfig:heat_eqn}\includegraphics[width=0.32\linewidth, height=4cm]{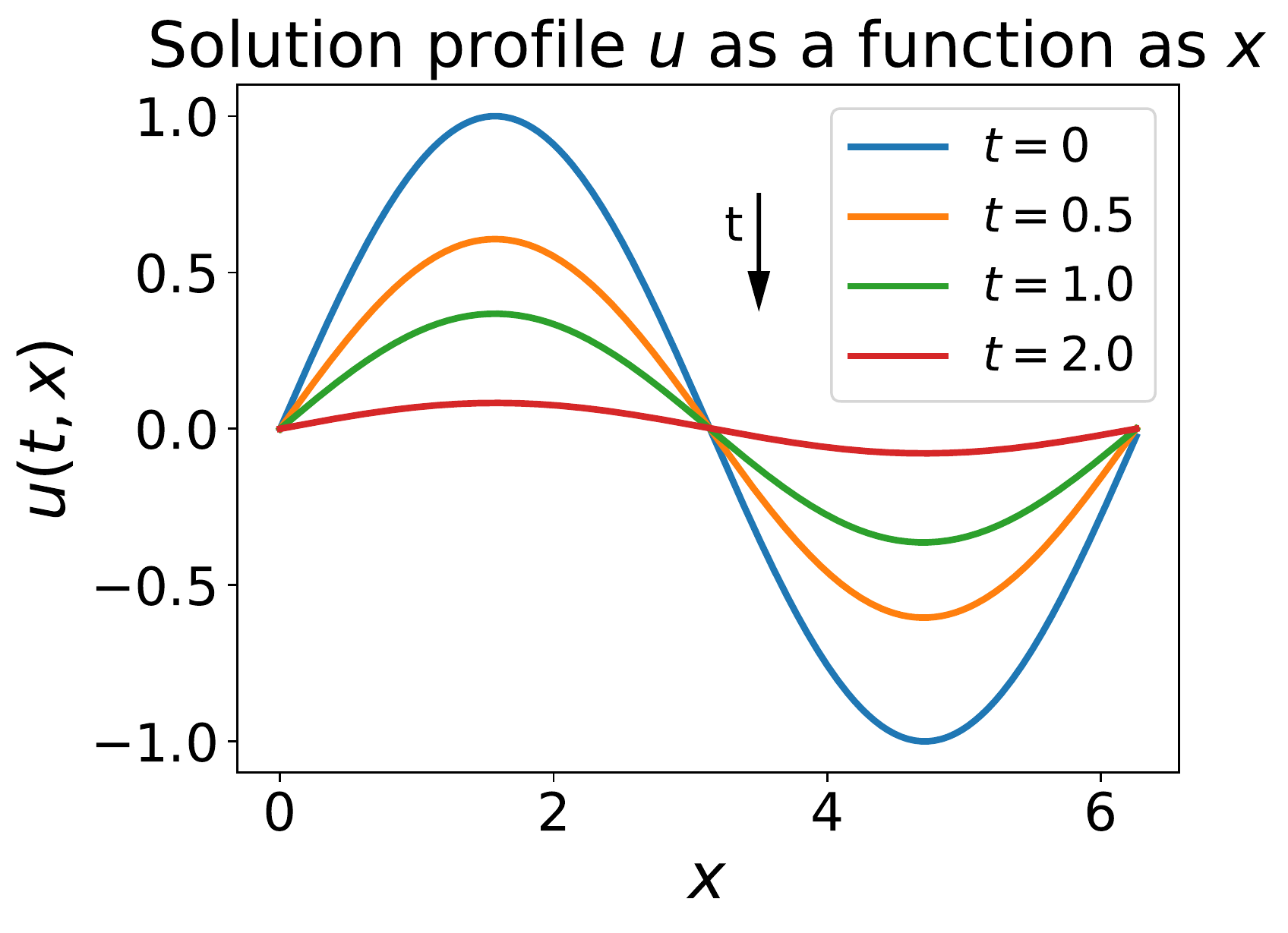}}
    \centering
    \subfigure[medium: PME ($k(u)=u^3$)]{\label{subfig:PME}\includegraphics[width=0.32\linewidth, height=4cm]{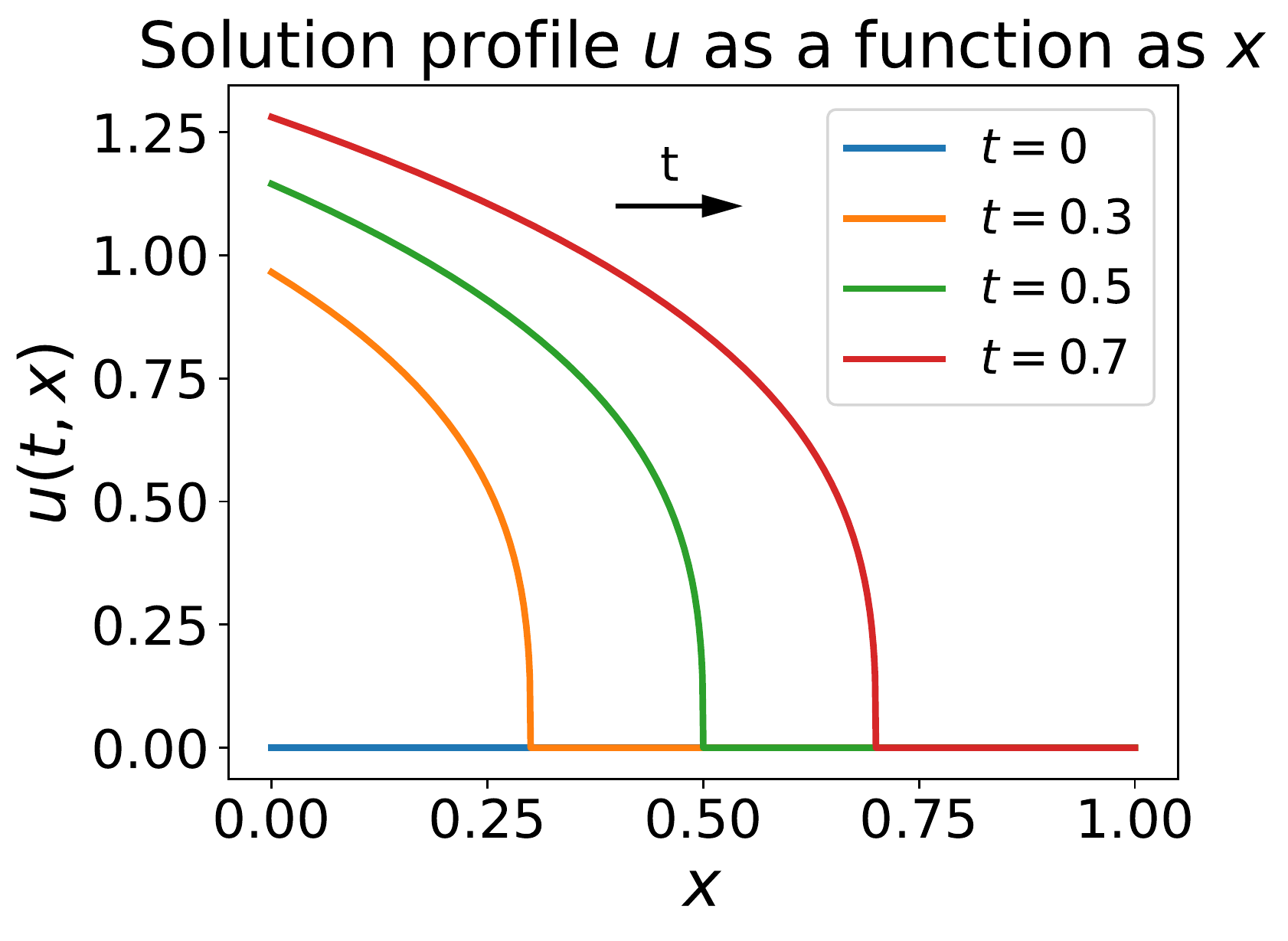}}
    \subfigure[hard: Stefan (discont. $k(u)$)]{\label{subfig:Stefan}\includegraphics[width=0.32\linewidth, height=4cm]{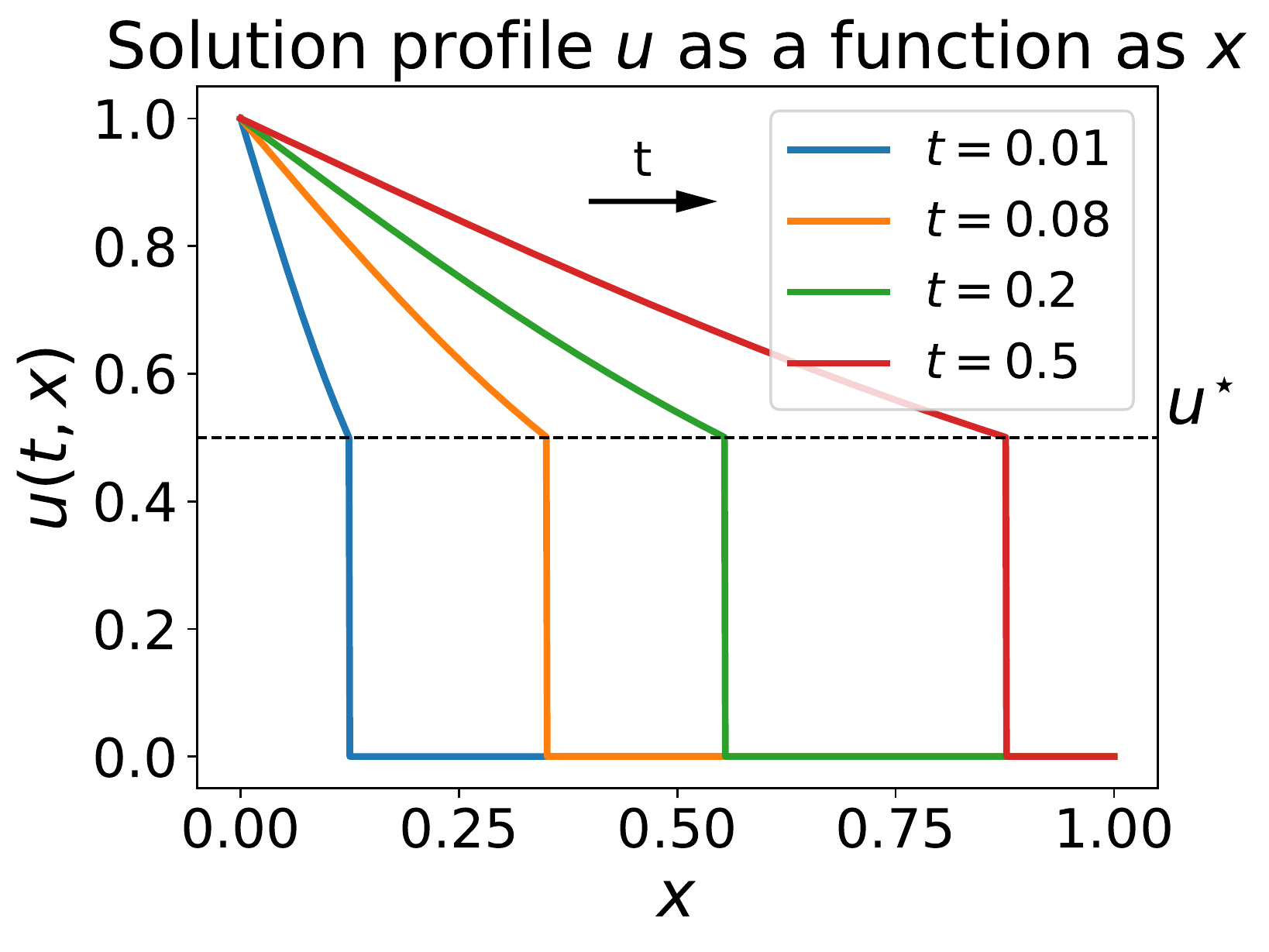}}
    \caption{Illustration of the ``easy-to-hard'' paradigm for PDEs, for the GPME family of conservation equations: 
    (a) ``easy'' parabolic smooth (diffusion equation) solutions, with constant parameter $k(u)=k\equiv1$; 
    (b) ``medium'' degenerate parabolic PME solutions, with nonlinear monomial coefficient $k(u) = u^m$, with parameter $m=3$ here; and  
    (c) ``hard'' hyperbolic-like (degenerate parabolic) sharp solutions (Stefan equation) with nonlinear step-function coefficient $k(u) = \1_{u \ge u^{\star}}$, where $\1_{\mathcal{E}}$ is an indicator function for event $\mathcal{E}$.
    }
    \label{fig:para_to_degenerate}
    \vskip -0.1in
\end{figure*}

In this section, we provide an empirical evaluation
to illustrate the main aspects of our proposed framework \probconservnosp. 
We choose 
the \ANP model \citep{kimAttentiveNeuralProcesses2019} as our black-box, data-driven model in Step~1, and we refer to this instantiation of our framework as \physnpnosp.\footnote{The code is available at \url{https://github.com/amazon-science/probconserv}.}
Unless otherwise stated, we use the limiting solution described in \autoref{eqn:updated_mean_var}, with $\sigma_G = 0$, so that 
conservation is enforced exactly through the integral form of the PDE.
We organize our empirical results around the following questions:
\begin{enumerate}
[noitemsep,topsep=0pt] 
    \item 
   \textit{Integral vs. differential form?} 
        \item
    \textit{Strong control on the enforcement of the conservation constraint?}
    \item 
    \textit{``Easy'' to ``hard'' PDEs?} 
    \item 
    \textit{Uncertainty Quantification (UQ) for downstream tasks?} 
\end{enumerate}

\paragraph{Generalized Porous Medium Equation.} 
The parametric Generalized Porous Medium Equation (GPME) is a \emph{family} of conservation equations, parameterized by a nonlinear coefficient $k(u)$.
It has been used in applications ranging from underground flow transport to nonlinear heat transfer to water desalination and beyond \citep{vazquez2007}. The GPME is given as: 
\begin{equation}
u_t - \nabla \cdot (k(u) \nabla u) = 0,
\label{eqn:gpme}
\end{equation}
where $F(u) = -k(u)\nabla u$ is a nonlinear flux function, and where the parameter $k=k(u)$ can be varied.  
Even though the GPME is nonlinear in general, for specific initial and boundary conditions, it has closed form self-similar solutions \citep{vazquez2007, maddix2018_harmonic_avg, maddix2018temp_oscill}.
This enables ease of evaluation by comparing each competing method to ground truth solutions.

By varying the parameter $k(u)$ in the GPME family, one can obtain PDE problems with widely-varying difficulties, from ``easy'' (where finite element and finite difference methods perform well) to ``hard'' (where finite volume methods are needed), and exhibiting many of the qualitative properties of smooth/easy parabolic to sharp/hard hyperbolic PDEs. See \autoref{fig:para_to_degenerate} for an illustration. 
In particular: 
the Diffusion equation is parabolic, linear and smooth, and represents an ``easy'' case (Sec.~\ref{scn:empirical_easy});
the Porous Medium Equation (PME) has a solution that becomes sharper (as $m \ge 1$, for $k(u)=u^m$, increases), and represents an ``intermediate'' or ``medium'' case (Sec.~\ref{scn:empirical_medium}); and
the Stefan equation has a solution that becomes discontinuous, and represents a ``hard'' case (Sec.~\ref{scn:empirical_hard}). 

We consider these three instances of the GPME (Diffusion, PME, Stefan) that represent increasing levels of difficulty. In particular, the challenging Stefan test case illustrates the importance of developing methods that satisfy conservation conditions on ``hard'' problems, with non-smooth and even discontinuous solutions, as well as for downstream tasks, e.g., the estimation of the shock 
position over time.
This is important, given the well-known inductive bias that many ML methods have toward smooth/continuous behavior.

See \autoref{app:overview_gpme} for more on the GPME; 
see \autoref{sec:detailed_experiments} for details on the \physnp model schematic (\autoref{fig:schematic}), model training, data generation and the ANP; and 
see \autoref{sxn:app-additional-empirical-results} for additional empirical results on the GPME and hyperbolic conservation laws.

\paragraph{Baselines.} 
We compare our results 
to the following baselines:

\begin{itemize}
[noitemsep,topsep=0pt] 
  \item 
  \ANPnosp: Base unconstrained ANP \citep{kimAttentiveNeuralProcesses2019}, trained to minimize the negative evidence lower bound (ELBO):
        \[
        \mathcal L = - \mathbb E_{D, u\sim p} \mathbb E_{z \sim q_{\phi}} \log p_{\theta} (u, z| D) - \log q_{\phi}(z | u, D),
        \]
    where $q_\phi$ denotes the variational distribution of the data used for training, and $p_\theta$ denotes the generative model.
    The ANP learns a global latent representation $z$ that captures uncertainty in global parameters, which influences the prediction of the reference solution $u$.
    At inference time, the distribution of $u$ given $z$ ($p_\theta (u \vert z, D)$) outputs a mean and diagonal covariance for Step~1. 
    \item 
    \SCANPnosp: In this ``Physics-Informed'' Neural Process ablation, we include a soft constrained PDE in the loss function, as is done with PINNs \citep{Raissi19}, to obtain:
    \[ 
    \mathcal{L} + \lambda \mathbb E_{z \sim q_{\phi}}\|\mathcal F\mu_z\|_2^2,
    \]
    where $\mathcal F$ denotes the underlying PDE differential form in \autoref{eq:gov_eqtn}, $\mu_z$ denotes the output mean of the ANP, and $\lambda$ denotes a hyperparameter controlling the relative strength of the penalty.
    (See Appendix \ref{app:scanp_lambda} for details on the hyperparameter tuning of $\lambda$.) 
\item 
\HCANPnosp: In this hard-constrained Neural Process ablation, we project the ANP mean to the nearest solution in $L_2$ satisfying the integral form of conservation constraint.
This method is inspired by the approach taken in \citet{hard_constraints} that projects the output of a neural network onto the nearest solution satisfying a linear PDE system.
\HCANP is an alternative to Step 2 that solves the following constrained least-squares problem:
\[
\begin{aligned}
\mu_{HC} &= \mathrm{argmin}_u \|u - \mu\|_2^2~ \text{s.t.}~ Gu = b \\
&= \mu - G^T (G G^T)^{-1} (G \mu- b).
\end{aligned}
\]
\HCANP is equivalent to the limiting solution of the mean of \probconserv as $\sigma_G \to 0$ in \autoref{eqn:updated_mean_var_MEAN}, if the variance from Step~1 is fixed to be the same for each point, i.e., $\Sigma = I$.
\end{itemize}

\paragraph{Evaluation.}
At test time, we select a value of the PDE parameter $\alpha$ that lies within the range of PDE parameters used during training (i.e., $\alpha \in \mathcal A$). For each value of $\alpha$, we generate multiple independent draws of $(D_i, u_i, b_i)$ in the same manner as the training data.
For a given prediction of the mean $\mu$ and covariance $\Sigma$ at a particular time-index $t_j$ in the training window, we report the following prediction metrics: conservation error $(\text{CE}(\mu) = (G\mu - b)_{t_j}$); predictive log-likelihood  $(\text{LL}(u; \mu, \Sigma) =  {-\frac 1 {2M} \|u_{t_j,\cdot} -  \mu_{t_j,\cdot}\|_{\Sigma^{-1}_{t_j}} - \frac{1}{2M} \sum_i \log \sigma^2_{t_j, i} - \log 2 \pi})$; and mean-squared error $(\text{MSE}(u, \mu) = \frac{1}{M} \|u_{t_j, \cdot} - \mu_{t_j, \cdot}\|^2_2)$,
where $M$ denotes the number of spatial points and $\sigma^2_{t_j,\cdot}$ denotes 
the diagonal of 
$\Sigma_{t_j} \in \mathbb{R}^{M\times M}$. We report the average of each metric over $n_{\text{test}}=50$ independent runs. Our convention for bolding the CE metric is binary on whether conservation is satisfied exactly or not. For the LL and MSE metrics, we bold the methods whose mean metric is within one standard deviation of the best mean metric. 

\subsection{Diffusion Equation: Constant $k$}
\label{scn:empirical_easy}

\begin{figure}[t]
\centering
\includegraphics[width=\figsize\linewidth]{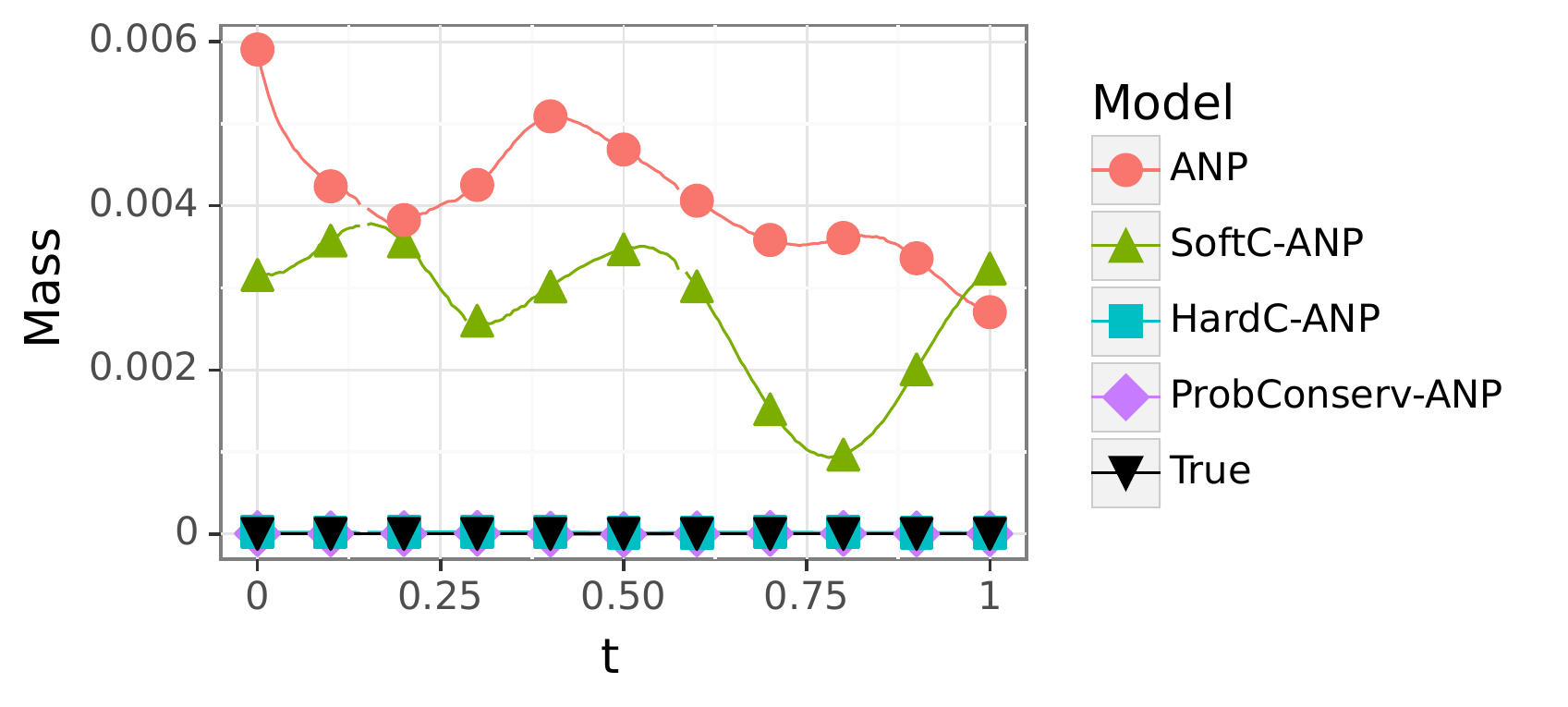}
\vskip 0.1in
    \caption{
   The total mass $U(t) = \int_\Omega u(t,x) d\Omega$ 
   as a function of time $t$ for the (``easy'') diffusion equation with constant diffusivity coefficient $k \in \mathcal{A} = [1,5]$ and test-time parameter value $k= 1$. The true $U(t)$ is zero at all times since there is zero net flux from the domain boundaries and mass cannot be created or destroyed on the interior. 
   Both \physnp and \HCANP satisfy conservation of mass exactly. The other baselines violate conservation 
   and result in a non-physical  
   mass profile over time. \ANP and \SCANP are not even zero at time $t=0$. 
   }
    \label{fig:heat_conserv_time}
    \vskip -0.1in
\end{figure}
The diffusion equation is the simplest non-trivial form of the GPME, with constant diffusivity coefficient $k(u) = k > 0$ (see Figure \ref{subfig:heat_eqn}). We train on values of $k \in \mathcal A = [1, 5]$.  The diffusion equation is also known as the heat equation, where in that application the PDE parameter $k$ denotes the conductivity and the total conserved quantity denotes the energy. In our empirical evaluations, we use the diffusion equation notation, and refer to the conserved quantity as the mass. 

\begin{table}[t]
   \caption{
     Mean and standard error for CE $\times 10^{-3}$ (should be zero), LL (higher is better) and MSE $\times 10^{-4}$ (lower is better) over $n_{\text{test}}=50$ runs for the (``easy") diffusion equation at time $t=0.5$ with variable diffusivity constant $k$ parameter in the range $\mathcal{A}=[1,5]$ and test-time parameter value $k=1$.
     }\label{tab:heat_vary_k}
     \vskip 0.1in
     \centering
\resizebox{\tabsize\linewidth}{!}{
    \begin{tabular}{c|lll}
        & CE                & LL                     & MSE                     \\
\midrule
\ANP    & 4.68 (0.10)       & $2.72$ (0.02)          & $1.71$ (0.41)           \\
\SCANP  & 3.47 (0.17)       & $2.40$ (0.02)          & $2.24$ (0.78)           \\
\HCANP  & \textbf{0} (0.00) & $\textbf{3.08}$ (0.04) &  $\mathbf{1.37}$ (0.33) \\
\physnp & \textbf{0} (0.00) & {2.74} (0.02)   & \textbf{1.55} (0.33)    \\
    \end{tabular}
    }
    \vskip -0.1in
\end{table}

\autoref{fig:heat_conserv_time}
illustrates that the unconstrained \ANP solution violates conservation by allowing mass to enter and exit the system over time. 
Physically, there is no in-flux or out-flux on the boundary of the domain, and thus the true total mass of the system $U(t) = \int_\Omega u(t,x) d\Omega$ is zero at all times.
Surprisingly, even incorporating the differential form of the conservation law as a soft constraint into the training loss via \SCANP violates conservation and the violation occurs even at $t=0$.

Enforcing conservation as a hard constraint in our \physnp model and \HCANP guarantees that the system total mass is zero, and also leads to improved predictive performance for both methods. In particular, \autoref{tab:heat_vary_k} shows that these methods exactly obtain the lowest MSE and the highest LL.
The success of these two approaches that enforce the integral form of the conservation law exactly, along with the failure of \SCANP that penalizes the differential form, demonstrates that physical knowledge must be properly incorporated into the learning process to improve predictive accuracy. \autoref{fig:heat_time_plot} in Appendix~\ref{app:heat_extra} illustrates that these conservative methods perform well on this ``easy'' case since the uncertainty from the \ANP is relatively homoscedastic throughout the solution space; that is, the estimated errors are mostly the same size, and the constant variance assumption in \HCANP holds reasonably well. 

\begin{table*}[t]
    \centering
     \caption{Mean and standard error for CE $\times 10^{-3}$ (should be zero), LL (higher is better) and MSE $\times 10^{-4}$ (lower is better) over $n_{\text{test}}=50$ runs for the (``medium'') PME at time $t=0.5$ with variable $m$ parameter in the range $\mathcal{A} = [0.99, 6]$.
     For test-time parameter $m=1$, where conservation by the unconstrained \ANP is violated the most, \physnp leads to a substantial $\mathbf{5.5 \times}$ improvement in MSE and log-likelihood. For test-time parameters $m=3, 6$, the MSE for \physnp increases due to the error concentrated at the sharper boundary while the desired log-likelihood and conservation metrics improve.
     }
    \vskip 0.1in
    \resizebox{1\linewidth}{!}{
    \begin{tabular}{c|lll|lll|lll}
      &        $m=1$  & & &     $m=3$ & & &$m=6$ \\
                      & CE         & LL                   & MSE                  & CE         & LL                   & MSE                  & CE         & LL                   & MSE \\                   
\midrule
\ANP                  & $6.67$ (0.39)     & $3.49$ (0.01)        & $0.94$ (0.09)        & $-1.23$ (0.29)   & $3.67$  (0.00)       & $1.90$ (0.04)        & $-2.58$ (0.23)    & $3.81$ (0.01)        & \textbf{7.67} (0.09) \\         
\SCANP                & $5.62$ (0.35)    & $3.11$ (0.01)        & $1.11$ (0.14)        & $-0.65$ (0.30)   & $3.46$  (0.00)       & $2.06$ (0.03)        & $-3.03$ (0.26)    & $3.49$ (0.00)        & $7.82$ (0.09)    \\      
\HCANP                & \textbf{0} (0.00) & 3.16  (0.04)         & 0.43   (0.04)        & \textbf{0} (0.00) & $3.44$  (0.03)       & \textbf{1.86} (0.03) & \textbf{0} (0.00) & 3.40 (0.05)          & \textbf{7.61} (0.09)   \\
\physnp               & \textbf{0} (0.00) & \textbf{3.56} (0.01) & \textbf{0.17} (0.02) & \textbf{0} (0.00) & \textbf{3.68} (0.00) & 2.10 (0.07)          & \textbf{0} (0.00) & \textbf{3.83} (0.01) & 10.4 (0.04)  \\ 
    \end{tabular}
}
     \label{tab:pme_vary_m}
     \vskip -0.1in
\end{table*}

\subsection{Porous Medium Equation (PME): $k(u) = u^m$}
\label{scn:empirical_medium}
The Porous Medium Equation (PME) is a subclass of the GPME in which the coefficient,
$k(u)=u^m,$
is nonlinear and smooth (see Figure \ref{subfig:PME}). The PME is known to be degenerate parabolic, with different behaviors depending on the value of $m$.
We train on values of $m \in \mathcal A = [0.99, 6]$.

\autoref{tab:pme_vary_m} compares the CE, MSE, and LL results for $m=1,3,6$.
These three values of $m$ reflect ``easy,'' ``medium,'' and ``hard'' scenarios, respectively, as the solution profile becomes sharper.
Despite achieving relatively low MSE for $m=1$, the \ANP model violates conservation the most. The error profiles as a function of $x$ in \autoref{fig:pme_error} in Appendix~\ref{app:scanp_lambda} illustrate the cause: 
the \ANP consistently overestimates the solution to the left of the shock.
Enforcing conservation consistently fixes this bias, leading to errors that are distributed around $0$.
Our \physnp method results in an $\approx 82\%$ improvement in MSE, and \HCANP results in an $\approx 54\%$ improvement over the \ANPnosp.
Since \HCANP shifts every point equally, it induces a negative bias in the zero (degeneracy) region of the domain, leading to a non-physical~solution.

For $m=3,6$, while the MSE for \physnp increases compared to the \ANPnosp, the LL for \physnp improves. The increase in LL for \physnp indicates that the uncertainty is better calibrated as a whole. \autoref{fig:pme_error} in Appendix~\ref{app:scanp_lambda} illustrates that \physnp reduces the errors to the left of the shock point while increasing the error immediately to the right of it. This error increase is penalized more in the $L_2$ norm, which leads to an increase in MSE. The LL metric improves because our \physnp model takes into account the estimated variance at each point.
It is expected that the largest uncertainty occurs at the sharpest part of the solution, since that is the area with the largest gradient. This region is more difficult to be captured as the shock interface becomes sharper when $m$ is increased. 

For control on the enforcement of conservation constraint, see \autoref{fig:pme_pred_wrt_precision} in \autoref{app:noise_param}, where we show empirically that the log likelihood is always increasing, as stated in Theorem \ref{thm:constraint_to_zero}. Note that there are optimal values of $\sigma_G^2$, in which case the MSE can be better optimized.

\subsection{Stefan Problem: Discontinuous Nonlinear $k(u)$}
\label{scn:empirical_hard}
\begin{table}[h]
\centering
     \caption{Mean and standard error for CE $\times 10^{-2}$ (should be zero), LL (higher is better), and MSE $\times 10^{-3}$ (lower is better) over $n_{\text{test}}=50$ runs for the (``hard'') Stefan variant of the GPME at time $t=0.05$. 
     Each model is trained with the parameter $u^{\star}$ in the range $\mathcal{A}=[0.55, 0.7]$ and test-time parameter value \(u^\star = 0.6\). \physnp leads to an increase in log-likelihood and a $\mathbf{3 \times}$ decrease in MSE. 
     }
    \label{tab:stefan_vary_u_star}
     \vskip 0.1in
    \resizebox{\tabsize\linewidth}{!}{
    \begin{tabular}{c|lll}
         & CE           & LL                              & MSE \\
\midrule
\ANP     & -1.30 (0.01)        & 3.53 (0.00)            & 5.38 (0.01)       \\ 
\SCANP   & -1.72 (0.04)        & \textbf{3.57} (0.01) & 6.81 (0.15)      \\   
\HCANP   & \textbf{0} (0.00)   & 2.33 (0.06)          & 5.18 (0.02)      \\  
\physnp  & \textbf{0} (0.00)   & \textbf{3.56} (0.00) & \textbf{1.89} (0.01) \\ 

    \end{tabular}
    }
     \vskip -0.1in
\end{table}

The most challenging case of the GPME is the Stefan problem.
In this case, the coefficient $k(u)$ is a discontinuous nonlinear step function $k(u) = \1_{u \ge u^{\star}}$, where $\1_{\mathcal{E}}$ denotes an indicator function for event  $_{\mathcal{E}}$ and 
$u^{\star} \in \mathbb{R}_+$. 
The solution is degenerate parabolic and develops a moving shock over time (see Figure \ref{subfig:Stefan}).
We train on values of $u^\star \in \mathcal{A} = [0.55, 0.7]$ and evaluate the predictive performances of each model at $u^\star = 0.6$.

\begin{figure}[h]
\vskip 0.1in
\begin{center}
\subfigure[Solution profile.]{
\includegraphics[width=\figsize\linewidth]{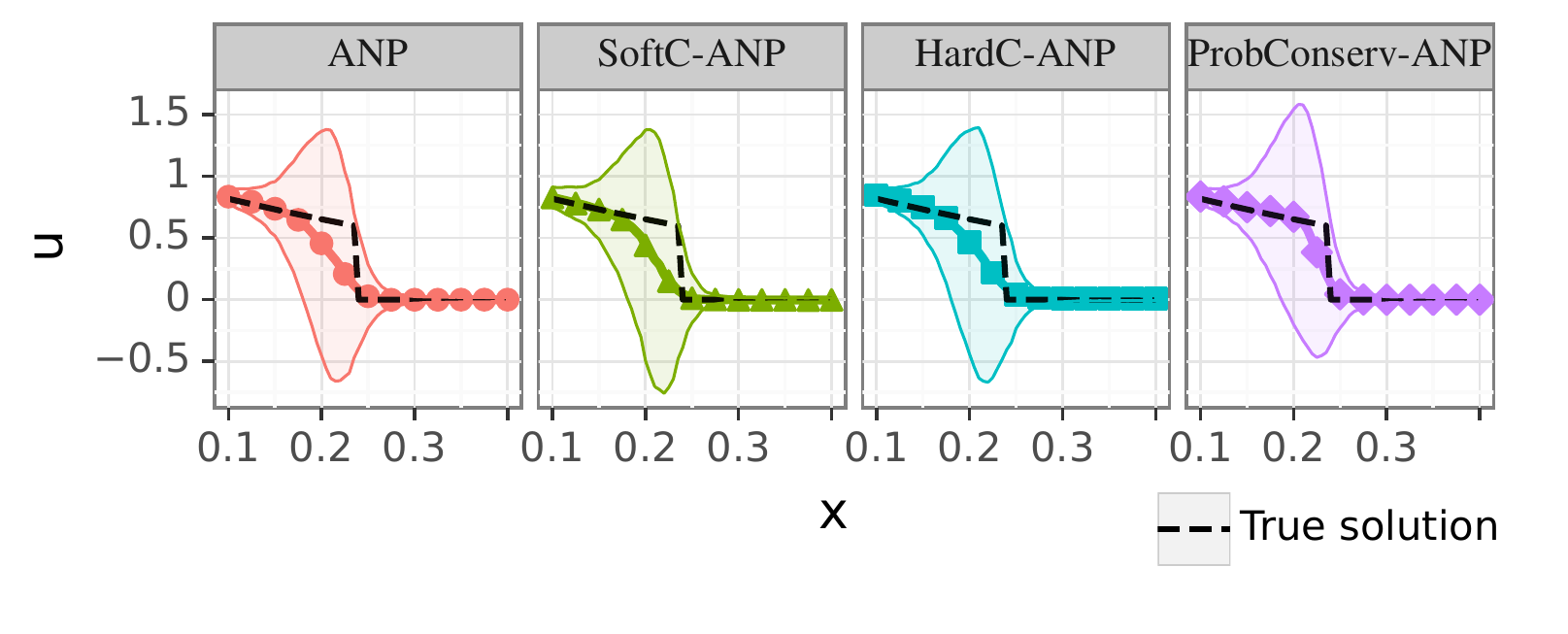}
\label{subfig:stefan_solution_profile}
}

\subfigure[Posterior of the shock position.]{
\includegraphics[width=\figsize\linewidth]{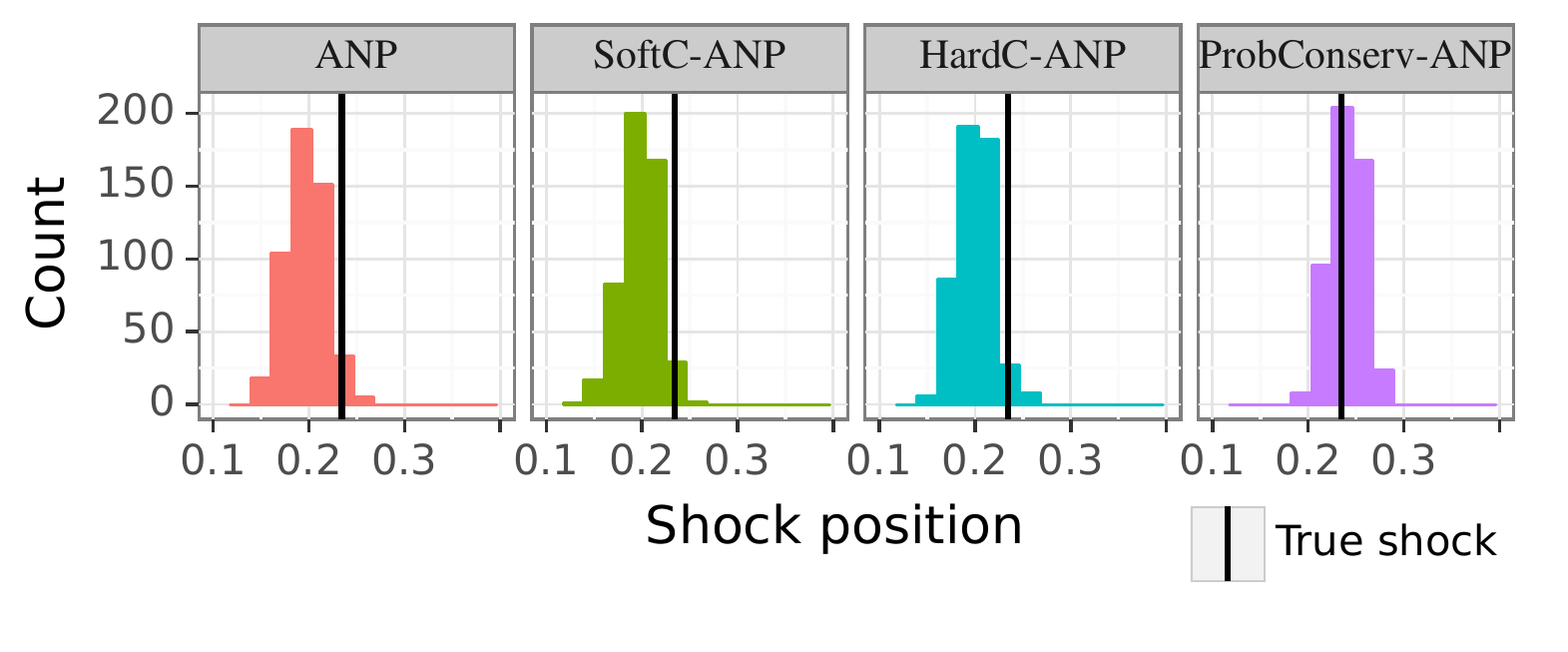}\label{fig:stefan_shock_indomain}
}
\vskip -0.1in
\end{center}
\caption{(a) Stefan solution profiles at time $t = 0.05$ with training parameter values $u^{\star} \in \mathcal{A}=[0.55, 0.7]$ and test-time parameter $u^\star = 0.6$. %
\physnp results in a sharper solution profile and the solution is mean-centered around the shock position.
(b) The corresponding histogram of the posterior of the shock position computed as the mean plus or minus 3 standard deviations. \physnp reduces the level of underestimation and the induced negative bias at the shock interface to result in more accurate shock position prediction.
}
\label{fig:stefan_sol_prof_indomain}
\vskip -0.1in
\end{figure}

Unlike the PME test case, 
where the degeneracy point ($x^*(t)=t$) is the same for each value of $m$, the shock position for the Stefan problem depends on the parameter $u^\star$ (See \autoref{fig:gpme_effect_parameters} in \autoref{app:exact_sol}).
This makes the problem more challenging for the \ANPnosp, as it can no longer memorize the shock position.
On this ``harder'' problem, 
the unconstrained ANP violates the physical property of conservation by an order of magnitude larger in CE than in the ``easier'' diffusion and PME cases.
By enforcing conservation of mass, \physnp results in substantial $\approx 65\%$ improvement in MSE (\autoref{tab:stefan_vary_u_star}).  In addition, Figure \ref{subfig:stefan_solution_profile} shows that the solution profiles associated with \ANP and the other baselines are smoothed and deviate more from the true solution than the solution profile of our \physnp model. 
Similar to our previous two case studies, adding the differential form of the PDE via \SCANP does not lead to a conservative solution (see \autoref{fig:stefan_cons} in Appendix \ref{app:stefan_extra}). 
In fact, \autoref{tab:stefan_vary_u_star} shows that surprisingly, conservation is violated more by \SCANP than with the \ANPnosp, with a corresponding increase in MSE.
These results demonstrate that physics-based constraints, e.g., conservation laws need be incorporated carefully (via finite volume based ideas) into ML-based models.

\autoref{tab:stefan_vary_u_star} shows that the LL for \physnp increases only slightly, compared to that of the \ANP (3.56 vs 3.53), and it is slightly less than \SCANPnosp.
Figure \ref{subfig:stefan_solution_profile} shows that  enforcing conservation of mass creates a small upward bias in the left part of the solution profile for $x \in [0, 0.2]$.
Since the variance coming from the ANP is smaller in that region, this bias is heavily penalized in the LL.
This bias is worse for \HCANPnosp, which assumes an identity covariance matrix and ignores the uncertainty estimates from the ANP.
\HCANP adds more noticeable upward bias to the $x \in [0, 0.2]$ region, and it even adds bias to the zero-density region to the right of the shock.
Compared to \physnpnosp, \HCANP only leads to a slight reduction in MSE (3\%) and a much lower LL (2.33).
This shows the benefit of using the uncertainty quantification from the ANP in our \physnp model for this challenging heteroscedastic case.

\paragraph{Downstream Task: Shock Point Estimation.}

While quantifying predictive performance in terms of MSE or LL is useful in ML, these metrics are typically not of direct interest to practitioners. To this end, we consider the downstream task of shock point estimation, which is an important problem in fluids, climate, 
and other areas.
The shock position for the Stefan problem \(x^\star (t)\) depends on the parameter $u^\star$. Hence, for a given function at test-time, the shock position \(x^\star (t)\) is unknown and must be predicted from the estimated solution profile.

We define the shock point at time $t$ as the first spatial point (left-to-right) where the function equals~zero:
\begin{equation}
x^\star (t) = \inf_x \{u(t, x) = 0\}.
\end{equation}
On a discrete grid,
we approximate the infimum using the minimum.
The advantage of a probabilistic approach is that we can directly quantify the uncertainty of $x^\star (t)$ by drawing samples from the posterior distributions of our \physnp model and the baselines.

Figure \ref{fig:stefan_shock_indomain} shows the corresponding histograms of the posterior of the shock position. We see that our \physnp posterior is centered around the true shock value.
By underestimating the solution profile, the \ANP misses the true shock position wide to the left, as do the other baselines \SCANP and \HCANPnosp. Remarkably, neither adding the differential form as a soft constraint (\SCANPnosp) nor projecting to the nearest conservative solution in $L_2$ (\HCANPnosp) helps with the task of shock position estimation. 
This result highlights that both capturing the physical conservation constraint and using statistical uncertainty estimates in our \physnp model are necessary on challenging problems with shocks, especially when the shock position is unknown.

\section{Conclusion}


We have formulated the problem of learning physical models that can respect conservation laws from the finite volume perspective, by writing the governing conservation law in integral form rather than the commonly-used (in SciML) differential form.  
This permits us to incorporate the global integral form of the conservation law as a linear constraint into black-box ML models; and 
this in turn permits us to develop a two-step framework that first trains a black-box probabilistic ML model, and then constrains the output using a probabilistic constraint of the linear integral form.
Our approach leads to improvements (in MSE, LL, etc.) for a range of ``easy'' to ``hard'' parameterized PDE problems.
Perhaps more interestingly, our unique approach of using uncertainty quantification to enforce physical constraints leads to improvements in challenging shock point estimation problems.
Future extensions include support for local conservation in finite volume methods, where the same linear constraint approach can be taken by computing the fluxes as latent variables; imposing boundary conditions as linear constraints \citep{saad_bc_2022}; and extension to other physical constraints, including nonlinear constraints, e.g., enstrophy in 2D and helicity in 3D, and inequality constraints, e.g., entropy \citep{osti_1395816}. 

\section*{Acknowledgments}
 
 Derek Hansen acknowledges support from the National Science Foundation Graduate Research Fellowship Program under grant no. 1256260. Any opinions, findings, and conclusions or recommendations expressed in this material are those of the author(s) and do not necessarily reflect the views of the National Science Foundation.
 The authors would also like to thank Margot Gerritsen and Yuyang Wang for their support.

\bibliography{derek_intern_project}
\bibliographystyle{icml2023/icml2023}

\newpage
\appendix
\onecolumn

\section{Related Works}
\label{sec:related_works}

Our method involves combining in a novel way ideas from several different literatures.
As such, there is a large body of related work, each of which approaches the problems we consider from somewhat different perspectives.
Here, we summarize some of the most related. \autoref{tab:related_work} provides an overview of the comparisons of these methods.

\begin{table*}[h]
\caption{Summary of different properties of numerical and SciML methods for physical systems. 
    } 
    \vskip 0.1in
\resizebox{1\linewidth}{!}{
    \begin{tabular}{lcccccc}
    \multirow{2}{*}{Method} &
    \multirow{2}{*}{Conservative} &
    \multirow{2}*{UQ} &    \multirow{2}{*}{\begin{tabular}[c]{@{}c@{}}Inference with different\\Initial Conditions\end{tabular}} & \multirow{2}{*}{\begin{tabular}[c]{@{}c@{}}Inference with different \\ PDE coefficients\end{tabular}} & \multirow{2}{*}{\begin{tabular}[c]{@{}c@{}}Resolution\\independent\end{tabular}} \\\\
    \midrule
       Numerical methods
        & \checkmark & \xmark & \xmark & \xmark & \xmark \\
    PINNs &
     \xmark & \xmark & \xmark & \xmark & \checkmark\\
    Neural Operators 
    & \xmark & \xmark & \checkmark &\checkmark & \checkmark\\
    Conservative ML models 
    & \checkmark & \xmark & \checkmark & \xmark & \xmark \\
    \midrule
    \probconserv (our approach) &\checkmark &\checkmark &\checkmark & \checkmark & \checkmark \\
    \end{tabular}
    }
    \label{tab:related_work}
     \vskip -0.1in
\end{table*}

\subsection{Numerical Methods} Numerical methods aim to approximate the solution to partial differential equations (PDEs) by first discretizing the spatial domain $\Omega$ into $N$ gridpoints $\{x_i\}_{i=1}^N$ with spatial step size $\Delta x$. Then, at each time step, we  integrate the resulting semi-discrete ODE in time with temporal step size 
$\Delta t$ to iteratively compute the solution at final time $T$, i.e., $\{u(T, x_i)\}_{i=1}^N$. By the Lax Equivalence theorem for linear problems, convergence to the true solution, i.e., the norm of the error tending to zero,
can be proven to occur when
$\Delta t, \Delta x \rightarrow 0$ ($N \rightarrow \infty$) 
for methods that are both stable and consistent \citep{leveque}. A limitation of numerical methods is that to obtain higher accuracy, fine mesh resolutions must be used, which can be computationally expensive in higher dimensions.  In addition, for changes in PDE parameters, the simulations need to be re-run. These classical methods are also deterministic, and they do not provide uncertainty quantification.
\paragraph{Finite Volume Methods.} Finite volume methods are designed for conservation laws. These methods divide the domain into control volumes, where the integral form of the governing equation is solved \citep{leveque1990numerical, leveque2002}. By solving the integral form at each control volume, these methods enforce flux continuity, i.e., that the out-flux of one cell is equal to the in-flux of its neighbor. This results in local conservation, which guarantees global conservation over the entire domain. \citet{maddix2018_harmonic_avg} show that the degenerate parabolic Generalized Porous Medium Equation (GPME) has presented challenges for classical averaged-based finite volume methods, e.g., arithmetic and harmonic averaging. These numerical artifacts include artificial temporal oscillations, and locking or lagging of the shock position. To eliminate these artifacts on the more challenging Stefan problem, \citet{maddix2018temp_oscill} show that information about the shock location needs to be incorporated into the scheme to satisfy the Rankine-Hugoniot condition. Other complex methods that explicitly track the front, e.g., front-tracking methods \citep{ALRAWAHI2002471, li_2003} and level set methods \citep{seth88} that implicitly model the interface as a signed distance function, have also been applied to the Stefan problem for modeling crystallization  \citep{SETHIAN1992231, CHEN19978}.

\paragraph{Reduced Order Models (ROMs).}
Reduced Order Models (ROMs) have been a popular alternative to full order model numerical PDE simulations for computational efficiency.  ROMs aim to approximate the solution in a lower dimensional subspace by computing the proper orthogonal decomposition (POD) basis using the singular value decomposition (SVD). Similar to deep learning models, there is no way to enforce that unconstrained ROMs are conservative and non-oscillatory.  \citet{osti_1395816} investigate enforcing conservative, entropy and total variation diminishing (TVD) constraints for ROMs as constrained nonlinear least squares problems. These methods are coined ``structure preserving''  ROMs via physics-based constraints \citep{Sargsyan2016DimensionalityHA}.

\subsection{Scientific Machine Learning (SciML) Models}
Here we describe the recent work in using ML models to solve PDEs. At a high-level, these works can be divided into three categories: 1. Physics-Informed Neural Networks (PINNs), which aim to incorporate PDE information as a soft constraint in the loss function; 2. Neural Operators, which aim to learn the solution mapping from PDE coefficients or initial conditions to solutions; and 3. Hard-constrained conservative ML models, which aim to incorporate different types of constraints to enforce conservation into the architecture. 
\paragraph{Physics-informed ML Methods.}
Physics-informed neural networks (PINNs) \citep{Raissi19} parameterize the solution to PDEs with a neural network (NN). These methods impose physical knowledge into neural networks by adding the differential form of the PDE to the loss function as a soft constraint or regularizer. Purely data-driven approaches include DeepONet \citep{deeponet} and Neural Operators (NOs) \citep{liNeuralOperatorGraph2020a, liFourierNeuralOperator2021b,guptaMultiwaveletbasedOperatorLearning2021}, which aim to learn the underlying function map from initial conditions or PDE coefficients to the solution. Learning this mapping enables these methods to be resolution independent, i.e., train on a coarse resolution and perform inference on a finer resolution. These methods only use PDE knowledge implicitly by training on simulations.  
The Physics-Informed Neural Operator (PINO) attempts to address that the physics are not directly enforced in the model by making the data-driven Fourier Neural Operator (FNO) ``physics-informed.'' To do so, they again add the differential form into the supervised loss function as a soft constraint regularization term \citep{PINO_2021, goswami22}.

Recently \citet{krishnapriyanCharacterizingPossibleFailure2021b,cacm_edwards22} identified  several challenges and limitations for SciML of this soft constraint approach on the training procedure for several PDEs with large parameter values. 
In particular, \citet{krishnapriyanCharacterizingPossibleFailure2021b} show that the sharp and non-smooth loss surface created by adding the PDE directly as a regularizer can be more difficult to optimize.    
Relatedly, PINO has been shown to perform worse than the base FNO without the differential form of the PDE as a soft constraint in the loss \citep{PINO_2021, saad_bc_2022}.
Motivated by these observations, 
\citet{hard_constraints} propose a solution for linear PDEs that enforces the differential form of the PDE as a hard constraint; and \citet{shashank22} propose another solution using an adaptive update of collocation points. 
In addition, \citet{PINNs_neural_tangent} examine training issues associated with the spectral bias in PINNs 
\citep{NEURIPS2018_5a4be1fa}. 
\citet{cacm_edwards22} discusses the broader-scale impacts of these results for the SciML field, and motivates the need for better solutions that capture the underlying continuous physics.

\paragraph{Machine Learning Models for Conservation Laws.}
Enforcing the PDE as a soft constraint gives very weak control on the physical conservation property, resulting in non-physical solutions that can violate governing conservation law. 
 \citet{choi} aim to satisfy conservation by adding the continuity equation as a soft regularizer via the PINNs approach, and they show that this does not improve performance.  
To try to remedy this, \citet{MAO2020112789, cpinn_2022} propose conservative PINNs (cPINNs) for conservation laws, which aim to enforce flux continuity, i.e., the out-flux of one cell equals the in-flux of the neighboring cell, for a type of local conservation. 
Again, however, this condition on the flux is added to the loss function as a regularization term, i.e., as a soft constraint in a Lagrange dual form, and so the conservation condition is in general not exactly satisfied.

Motivated by the importance of satisfying conservation laws in climate applications, \citet{boltonApplicationsDeepLearning2019, zannaDataDrivenEquation2020, beuclerEnforcingAnalyticConstraints2021} have proposed building known linear physical constraints directly into deep learning architectures.
\citet{beuclerEnforcingAnalyticConstraints2021} propose a model that forces the output of a neural network into the null space of the constraint matrix.
While the solution exactly satisfies the constraints, the constraints depend on the resolution of the data, and they are an approximation to the true physical quantity that needs to be constrained. 
Surprisingly, \citet{beuclerEnforcingAnalyticConstraints2021} also finds that the reconstruction error is not always improved with adding constraints. Other methods to enforce conservation include the following. \citet{sturm2022} enforce the flux continuity equation  in the last layer of the neural network to model the balance of atoms.
\citet{muller2022} enforce conservation by encoding symmetries using Noether's theorem.
\citet{richterpowell2022neural} propose so-called Neural Conservation Laws, to enforce conservation by design by using parametizations of deep neural networks similar to the approaches in \citet{hard_constraints, sturm2022, muller2022}.  In particular, \citet{richterpowell2022neural} use a change of variables that combines time and space derivatives into the divergence operator to create a  divergence-free model, 
and they then use auto-differentiation similar to the Neural ODEs approach \citep{NEURIPS2018_node}.  
This optimize-then-discretize approach has been shown to have related difficulties \citep{learn_cont_physics, ott2021resnet, dis_then_opt}.

\section{Derivation of the Integral Form of a Conservation Law} 
\label{app:differential_to_integral_form}

To obtain the integral form of a conservation law, given in \autoref{eq:int_form} as:
\begin{equation}
\begin{aligned} 
 \int_{\Omega} u(t,x)d\Omega = \int_{\Omega} h(x)d\Omega - \int_0^t \int_{\Gamma} F(u) \cdot n d\Gamma dt,
   \end{aligned} 
    \label{eq:int_form_app}
\end{equation}
we first integrate the differential form of the conservation law, given in \autoref{eq:diff_form} as:
\begin{equation}
    \mathcal{F}u = u_t + \nabla \cdot F(u),
    \label{eq:diff_form_app}
\end{equation} over the spatial domain $\Omega$.
From this, we obtain an expression for the rate of change in time of the total conserved quantity in terms of the fluxes on the boundary, given as: 
\begin{subequations}
\label{eqn:deriv_diff_to_int_form}
\begin{align}
    \frac{d}{dt} \int_{\Omega} u(t,x)d\Omega 
    &= \int_{\Omega} u_t(t,x)d\Omega \label{eqn:time_deriv} \\ 
    &= - \int_{\Omega} \nabla \cdot F(u)d\Omega \\
    &= -\int_{\Gamma} (F(u) \cdot n) d\Gamma \label{eqn:rhs},
\end{align}
\end{subequations}
where the last step is obtained by applying the divergence theorem to the flux term, and $n$ is the outward pointing unit normal on the boundary $\Gamma$.

We then integrate \autoref{eqn:deriv_diff_to_int_form} over the temporal domain $[0,t]$. 
Doing this to \autoref{eqn:time_deriv} yields:
\[
\int_0^t\int_{\Omega} u_t(t,x)d\Omega = \int_{\Omega} u(t,x)d\Omega
- \int_{\Omega} u(0,x)d\Omega,
\] 
where $u(0,x) = h(x)$ denotes the initial condition. 
By equating this quantity to the temporal integral of the right hand side of \autoref{eqn:rhs}, we obtain the corresponding integral form of a conservation law:
\begin{equation*}
\begin{aligned} 
 \int_{\Omega} u(t,x)d\Omega = \int_{\Omega} h(x)d\Omega - \int_0^t \int_{\Gamma} F(u) \cdot n d\Gamma dt ,
   \end{aligned} 
\end{equation*}
which is \autoref{eq:int_form_app}.

\section{Exact Solutions and Linear Conservation Constraints for Conservation Laws}
\label{app:exact_sol}

In this section, we provide the exact solutions to a wide range of conservation laws:
\begin{equation}
\begin{drcases}
    &  \underbrace{u_t + \nabla \cdot F(u)}_{\mathcal{F}u} = 0, \,\, x \in \Omega, \\ 
    & u(0,x) = h(x), \\
    & u(t,x) = g(t,x), \,\, x \in \Gamma,
    \label{eq:gov_eqtn_app}
\end{drcases}, \forall\,t \geq 0,
\end{equation}
for general nonlinear flux $F(u)$, nonlinear differential operator $\mathcal{F}$, initial condition $h(x)$ and prescribed boundary conditions on the boundary $\Gamma$ of the spatial domain $\Omega$. 
These exact solutions are used to generate the solution samples for the training data in the experiment \autoref{sec:experiments}.  

The integral form of the conservation law in \autoref{eqn:global_conserv_mb} is given as:
\begin{equation}
\begin{aligned}
    \underbrace{\int_{\Omega} u(t,x)d\Omega}_{\mathcal{G}u(t,x)}  = \underbrace{\int_{\Omega} h(x)d\Omega + \int_{0}^{t}
    (F_{\text{in}} - F_{\text{out}})dt}_{b(t)},
     \label{eqn:global_conserv_mb_app}
\end{aligned}
\end{equation} 
where  $\Omega = [x_0, x_N]$, $F_{\text{in}} = F(u, t, x_0)|_{u=g(t,x_0)}$, $F_{\text{out}} = F(u, t, x_N)|_{u=g(t,x_N)}$ and $g(t,x)$ is the prescribed Dirichlet boundary condition in \autoref{eq:gov_eqtn_app}. 
We provide the exact formulation of our linear constraint $\mathcal{G}u(t,x)=b(t)$. \autoref{tab:linear_conserv} provides a summary, showing that our boundary flux linearity assumption holds for a broad class of problems---even including nonlinear conservation laws with nonlinear PDE operators $\mathcal{F}$.

\begin{table}[h]
\centering
\caption{Classification of PDE conservation laws ranging from ``easy'' to ``hard'', and corresponding total time-varying conserved value $b(t)$ in the integral form of \autoref{eqn:global_conserv_mb_app} for specified flux function $F(u)$, initial and boundary conditions $h(x)$ and $g(t,x)$, respectively in \autoref{eq:gov_eqtn_app}. See Section \ref{subsect:stefan} for the value of the constant $c_1 \in \mathbb{R}$.}\label{tbl:classification_pde_laws}
\vskip 0.1in
\resizebox{1\linewidth}{!}{
\begin{tabular} {llllllll}
\toprule
PDE & Type & $F(u)$& $h(x)$ & $g(t,x)$ & $\Omega$ &$ \Gamma$ &$b(t)$ \\ 
\midrule
\multirow{2}{*}{\begin{tabular}[c]{@{}c@{}} Diffusion \end{tabular}}  & \multirow{2}{*}{\begin{tabular}[c]{@{}c@{}}Linear parabolic \\(``easy'')\end{tabular}}  & \multirow{2}{*}{\begin{tabular}[c]{@{}c@{}} $-k\nabla u$, $k \in \mathbb{R}_+$ \end{tabular}} & \multirow{2}{*}{\begin{tabular}[c]{@{}c@{}} $\sin(x)$ \end{tabular}}  & \multirow{2}{*}{\begin{tabular}[c]{@{}c@{}} $\{0,0\}$\end{tabular}} & \multirow{2}{*}{\begin{tabular}[c]{@{}c@{}} $[0, 2\pi]$\end{tabular}} &\multirow{2}{*}{\begin{tabular}[c]{@{}c@{}} $\{0, 2\pi\}$\end{tabular}} & \multirow{2}{*}{\begin{tabular}[c]{@{}c@{}} $0$\end{tabular}}  \\\\
\midrule
PME & {\begin{tabular}[c]{@{}c@{}}Nonlinear degenerate \\ parabolic (``medium'')\end{tabular}}  & $-u^m \nabla u$, $m \in \mathbb{Z}_{+}$ & 0 & $\{(mt)^{1/m}, 0\}$ & $[0,1]$ & $\{0,1\}$ & $\frac{m^{1+1/m}}{m+1}t^{1+1/m}$ \\
\midrule
Stefan &{\begin{tabular}[c]{@{}c@{}}Nonlinear degenerate \\
parabolic (``hard'')\end{tabular}}& 
\(
    \begin{dcases}
    -\nabla u, \hspace{0.1cm} u \ge u^{\star} \\
    0, \hspace{0.1cm} \text{otherwise}
    \end{dcases}
\), $u^{\star} \in \mathbb{R}_+$ & $0$ & $\{1, 0\}$ & $[0,1]$ & $\{0,1\}$ & $2c_1\sqrt{t/\pi}$ \\ 
\midrule
\midrule
Advection &{\begin{tabular}[c]{@{}c@{}}Linear hyperbolic \\(``medium'')\end{tabular}}   & $\beta u$, $\beta \in \mathbb{R}_+$ & 
\(
    \begin{dcases}
    1, \hspace{0.1cm} x \le 0.5 \\
    0, \hspace{0.1cm} \text{otherwise}
    \end{dcases}
    \nonumber
\) & $\{1,0\}$ & $[0,1]$ & $\{0,1\}$ & $\frac{1}{2} + \beta t$
\\
\midrule
Burgers' &  {\begin{tabular}[c]{@{}c@{}}Nonlinear \\ hyperbolic (``hard'')\end{tabular}}  & $\frac{1}{2}u^2$  &  
\(
    \begin{dcases}
    -ax, \hspace{0.1cm} x \le 0, a \in \mathbb{R}_+ \\
    0, \hspace{0.1cm} \text{otherwise}
    \end{dcases}
    \nonumber
\) & $\{a,0\}$ & $[-1,1]$ & $\{-1,1\}$ & $(a/2)(1+at)$ \\
\bottomrule
\end{tabular}
}
\label{tab:linear_conserv}
\vskip -0.1in
\end{table}

\subsection{GPME Family of Conservation Laws}
\begin{figure*}[h]
\vskip 0.1in
    \centering
    \subfigure[$k$: Diffusion equation at $t=1$ (``easy'').]{\label{fig:heat_eqn_diff_k}\includegraphics[width=0.32\linewidth, height=3cm]{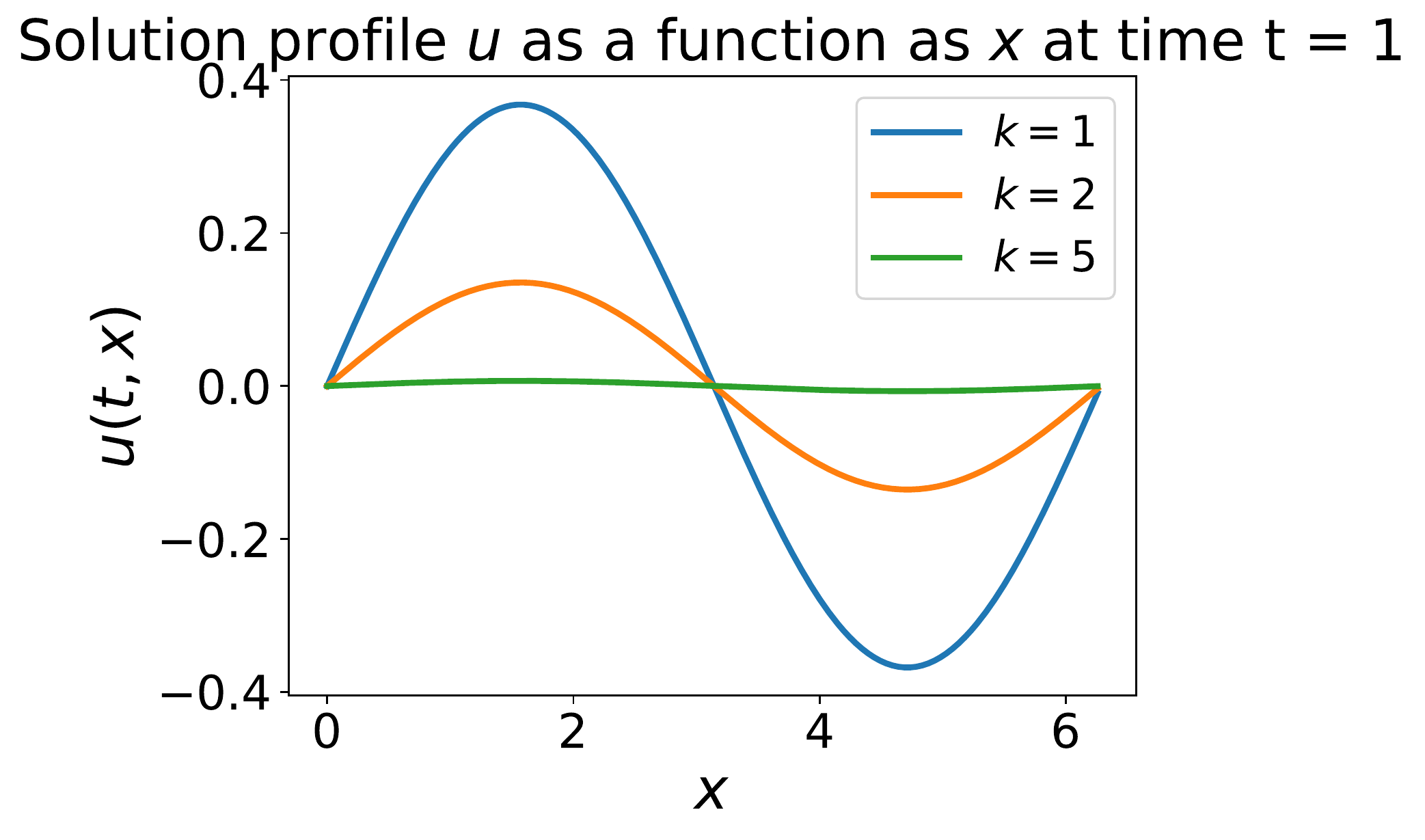}}
    \centering
    \subfigure[$m$: PME at $t=0.5$ (``medium'').]{\label{fig:effect_m}\includegraphics[width=0.32\linewidth, height=3cm]{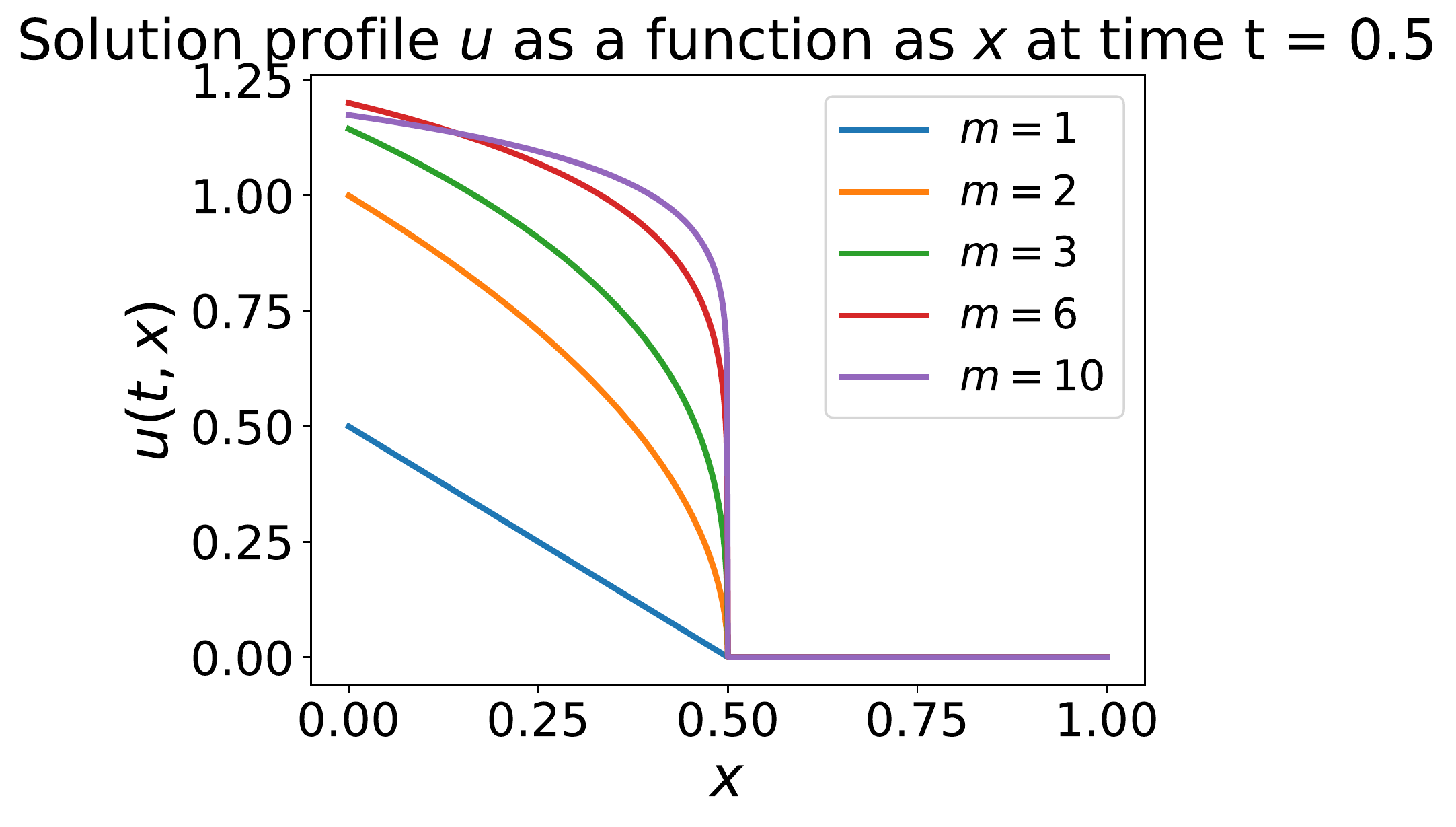}}
    \subfigure[$u^{\star}$: Stefan at $t=0.08$ (``hard'').]{\label{fig:effect_u_star}\includegraphics[width=0.32\linewidth, height=3cm]{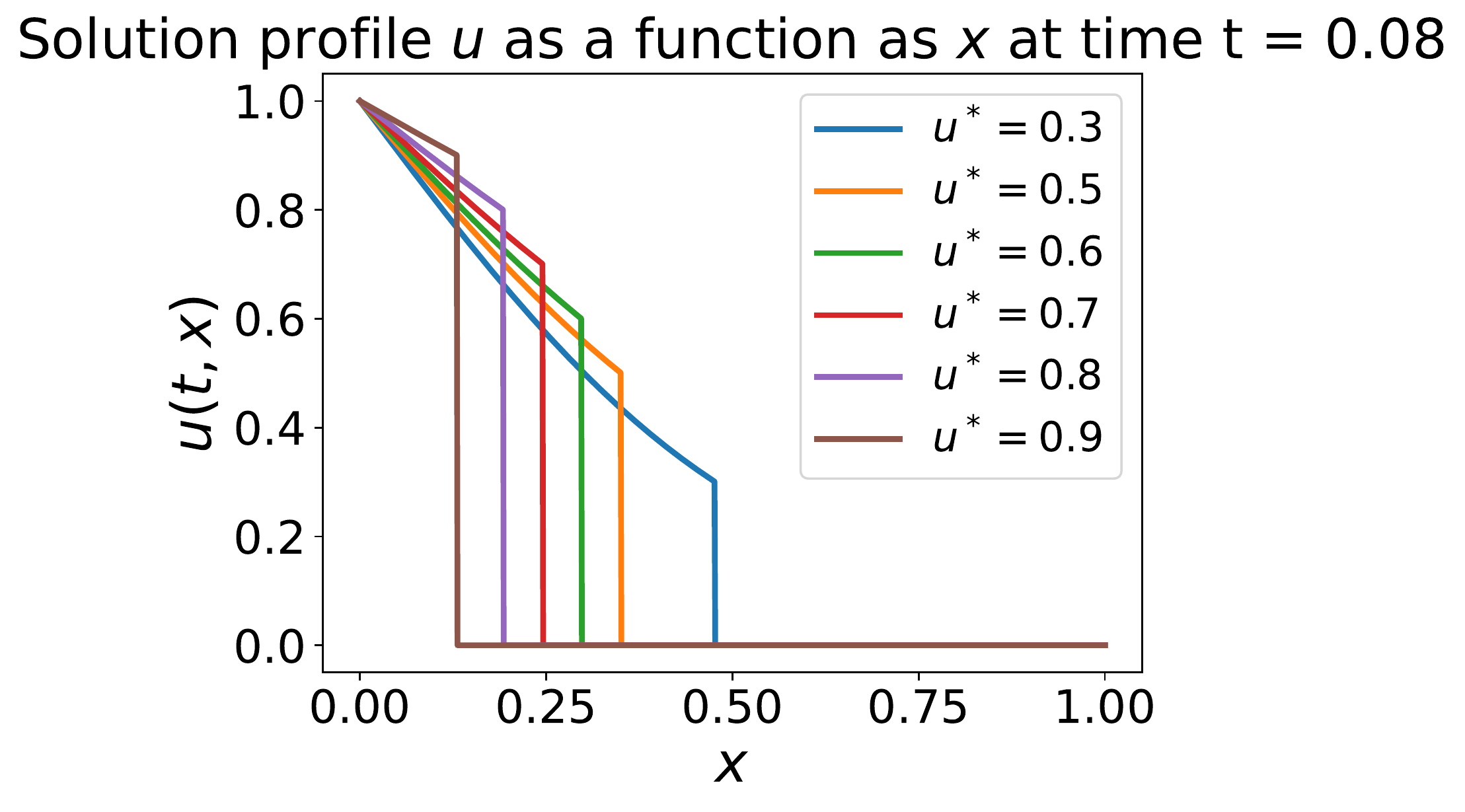}}
    \caption{Effect of PDE parameters on the three ``easy'' to ``hard'' instances of the GPME at fixed time $t$.}
    \vskip -0.1in
    \label{fig:gpme_effect_parameters}
\end{figure*}

In this subsection, we consider the (degenerate) parabolic GPME family of conservation laws given in \autoref{eqn:gpme} as:
\[
u_t - \nabla \cdot (k(u) \nabla u) = 0,
\]
with flux $F(u) = -k(u)\nabla u$. \autoref{fig:gpme_effect_parameters} shows the effects of the various PDE parameters $k(u)$ at a fixed time $t$ on the solution on three instances of the GPME ranging from the ``easy'' to ``hard'' cases, i.e., the diffusion equation, PME and Stefan, respectively.

\subsubsection{Diffusion Equation}
The heat or diffusion equation is a simple linear parabolic PDE with constant coefficient $k(u) = k$, which represents an ``easy'' task.  Figure \ref{fig:heat_eqn_diff_k} illustrates the effect of the constant diffusivity (conductivity) parameter $k$ on solutions to the diffusion (heat) equation.  For larger values of $k$, we see that the solution more quickly dissipates toward the constant smooth zero steady state.

\paragraph{Exact Solution.} We use the same diffusion test problem from \citet{krishnapriyanCharacterizingPossibleFailure2021b} 
with the following initial condition and periodic boundary conditions: 
\[
    \begin{aligned}
    u(0,x) &= h(x) = \sin(x), \forall x \in \Omega = [0, 2\pi], \\
    u(t,0) &= u(t, 2\pi), \forall t \ \in [0,T],
    \end{aligned}
\]
respectively.
The exact solution is given as 
\[
 u(t,x) = FT^{-1}(FT(h(x))e^{-kn^2t}),
\]
where $FT$ denotes the Fourier transform, and $n$ denotes the frequency in the Fourier domain.

\paragraph{Global Conservation.} 
The total mass (energy) is constant and zero over all time, since there is no in or out flux to the domain. Then, \autoref{eqn:global_conserv_mb_app} reduces to the following linear homogeneous system:

\begin{equation}
\mathcal{G}u(t,x) = \int _{x_0}^{x_N} u(t,x)  dx = 0 = b(t).
\end{equation}

To derive the above relation, we see by using separation of variables that the solution $u(t,x) = \sin(x)T(t)$ is a damped sine curve over time. The flux $F(u) = -k \nabla u = -\cos(x)T(t)$, where $T(t)$ denotes a decaying exponential function. Then, the integral form in \autoref{eqn:global_conserv_mb_app} is given as: 
\[
\begin{aligned}
 \int_{\Omega} u(t,x)d\Omega  &=  \int_{\Omega} h(x)d\Omega + \int_{0}^{t}[F(u, t,x_0=0) - F(u,t,x_N=2\pi)]dt \\
 &=\int_{0}^{2\pi} \sin(x)d\Omega - k\int_{0}^{t}[\cos(0)T(t) - \cos(2\pi)T(t)]dt = 0,
 \end{aligned}
\]
by periodicity.

\subsubsection{Porous Medium Equation}
In the Porous Medium Equation (PME), the nonlinearity and small values of the coefficient $k(u)=u^m$, for $m \ge 1$, cause challenges for current state-of-the-art SciML baselines as well as classical numerical methods on this degenerate parabolic equation. The difficulty increases as the exponent $m$ increases, 
and the solution forms sharper corners.
In particular, the solution gradient is finite for $m = 1$, and it approaches infinity near the front for $m > 1$. 
Figure \ref{fig:effect_m} illustrates the effect of the parameter $m$ on the solution, with solutions for $m > 1$ being sharper, and having a different profile than those for the piecewise linear solution for $m=1$.
\paragraph{Exact Solution.} We test the locking problem (TLP) of the PME from \citet{LIPNIKOV2016111, maddix2018_harmonic_avg} with the following initial and growing in time Dirichlet left boundary conditions for some final time $T \le 1$:
\[
    \begin{aligned}
    u(0,x) &=h(x)=0, \forall x\in \Omega = [0,1], \\
    u(t,0) &= g(t, 0) = (mt)^{1/m}, \forall t \in [0,T], \\
     u(t,1) &= g(t, 1) = 0, \forall t \in [0,T], \\
    \end{aligned}
    \nonumber
\]
respectively. 
The exact solution is given as:
\begin{equation}
    u(t,x) = (m(t-x)_+)^{1/m}.
\end{equation}

\paragraph{Global Conservation.}
We write the specific form of the linear conservation constraint in \autoref{eqn:global_conserv_mb_app} for the PME as:

\begin{equation}
\mathcal{G}u(t,x) = \int _{x_0}^{x_N} u(t,x)  dx = \frac{m^{1+1/m}}{m+1}t{^{1+1/m}} = b(t),
\end{equation}
by using the fact that the total mass of the initial condition is zero, and that $u(t, x_N=1)=0$ on the right boundary for $t \le x_N=1$.

Global conservation is driven by the in-flux at the growing in left boundary, where 
$$F_{\text{in}} = F(u, t, x_0)|_{u=g(t,x_0), x= x_0} = -g(t,x_0)^m\nabla u|_{x= x_0} = -mt\nabla u|_{x= x_0} . $$ 
The boundary flux at the right boundary is 0, since we assume that the shock is contained in the domain and $t > x$, hence $u(t,1) = 0$ and 
$$F_{\text{out}} =  F(u, t, x_N)|_{u=g(t,x_N), x= x_N} =  -g(t,x_N)^m\nabla u|_{x= x_N} = 0 . $$ 
The first integral on the righthand side in \autoref{eqn:global_conserv_mb_app} consisting of the initial mass is 0, since $h(x)=0$, and we are left only with the in-flux term:

\[
\begin{aligned}
 \int_{\Omega} u(t,x)d\Omega  &=  \int_{0}^{t}F_{\text{in}}(t)dt \\
 &=-\int_{0}^{t}g(t,x_0=0)^m\nabla u|_{x= x_0}dt\\
 &=\int_{0}^{t}(mt)(mt)^{1/m-1}dt \\
 &=m^{1/m}\int_{0}^{t}t^{1/m}dt \\
 & = \frac{m^{1+1/m}}{m+1}t^{1+1/m},
 \end{aligned}
\]
where $\nabla u|_{x= x_0} = -(m(t-x))^{1/m-1}|_{x= x_0} = -(mt)^{1/m-1}$.

\subsubsection{Stefan Problem}
\label{subsect:stefan}
The Stefan problem is the most challenging problem in the GPME degenerate parabolic family of conservation equations since the coefficient $k(u)$ is a nonlinear step function of the unknown $u$, given as:
\begin{equation}
k(u) = 
\begin{cases}
k_{\max}, \hspace{0.1cm} u \ge u^{\star}, \\
k_{\min}, \hspace{0.1cm} u < u^{\star},
\end{cases}
\label{eqn:stefan_app_1}
\end{equation}
for constants $k_{\max}, k_{\min} \in \mathbb{R}$ and $u(t, x^*(t)) = u^* \in \mathbb{R}_+$ for shock position $x^*(t)$.  In this problem, the solution is a shock or moving interface with a finite speed of propagation that does not dissipate over time. 
Figure \ref{fig:effect_u_star} illustrates the effect of the parameter $u^{\star}$ on the solution and shock position, with smaller values of $u^{\star}$ resulting in a faster shock speed.

\paragraph{Exact Solution.} We use the Stefan test case from \citet{vandermeer2016, maddix2018temp_oscill} with $k_{\max}=1$, $k_{\min}=0$ in \autoref{eqn:stefan_app_1}, and the following initial and Dirichlet boundary conditions for some final time $T$:
\[
\begin{aligned}
    u(0,x) &=h(x)=0, \forall x\in \Omega = [0,1], \\
    u(t,0) &= g(t,0) = 1, \forall t \in [0,T], \\
    u(t,1) &= g(t, 1) = 0, \forall t \in [0,T], \\
\end{aligned}
\]
respectively. 
The exact solution is given as:
\begin{equation}
 u(t,x) = \1_{u \ge u^\star} \left(1-c_1\Phi[x/(2\sqrt{k_{\text{max}}t})] \right),
\end{equation}
where $\1_{\mathcal{E}}$ denotes an indicator function for event $\mathcal{E}$, $\Phi(x) = \text{erf}(x)=\int_0^x\phi(y)dy$ denotes the error function with $\phi(y) = (2/\sqrt{\pi})\exp(-y^2)$, and constant $c_1 = (1-u^*) / \Phi[\alpha/(2\sqrt{k_{\max}})]$. A nonlinear solve for $\tilde{\alpha}$: 
$(1-u^*)/\sqrt{\pi} = u^*\Phi(\tilde{\alpha})\tilde{\alpha}\exp(\tilde{\alpha}^2)$,
is used to compute $\alpha = 2\sqrt{k_{\max}}\tilde{\alpha}$.  The exact shock position is $x^*(t) = \alpha \sqrt{t}$.  

\paragraph{Global Conservation.}
We write the linear $\mathcal{G}$ conservation constraint in \autoref{eqn:global_conserv_mb_app} for the Stefan equation as:
\begin{equation}
\mathcal{G}u(t,x) = \int _{x_0}^{x_N} u(t,x)  dx = 2c_1\sqrt{\frac{k_{\max}t}{\pi}} = b(t).
\end{equation}
We use the fact that the solution is monotonically non-increasing to compute the coefficient values at the boundaries, i.e., $u(t, x_0) \ge u^{\star} \ge u(t, x_N)$, where $0 = x_0 \le x^{\star} \le x_N =1$ and $x^*(t)$ denotes the shock position. It follows that $k(u(t, x_0)) = k_{\max}$ and $k(u(t, x_N)) = 0$. Then the out-flux $F_{\text{out}} = k(u(t, x_N))\nabla u = 0$. The first integral on the righthand side of \autoref{eqn:global_conserv_mb_app} consisting of the initial mass is 0, since $h(x)=0$, and we are left only with the in-flux term as follows:
\[
\begin{aligned}
 \int_{\Omega} u(t,x)d\Omega  &=  \int_{0}^{t}F_{\text{in}}(t)dt \\
 &=-k_{\max}\int_0^t\nabla u|_{x= x_0}dt\\
 &=c_1\sqrt{\frac{k_{\max}}{\pi}}\int_0^t t^{-1/2}dt\\
 &=2c_1\sqrt{\frac{k_{\max}t}{\pi}},
 \end{aligned}
\]
where $\nabla u|_{x= x_0} = -c_1\Phi'[x_0/(2\sqrt{k_{\max}}t)]/(2\sqrt{k_{\max}}t) = -c_1/\sqrt{\pi k_{\max}t}\exp[x_0^2/(4k_{\max}t)] = -c_1/\sqrt{\pi k_{\max}t}$ for $x_0=0$.

\subsection{Hyperbolic Conservation Laws}
In this section, we consider hyperbolic conservation laws, where solutions exhibit shocks and smooth initial conditions self-sharpen over time \citep{leveque1990numerical, leveque2002}.
\subsubsection{Linear Advection}
The linear advection (convection) equation: 
\begin{equation}
    u_t + \beta u_x = 0,
    \label{eqn:advection}
\end{equation}
is a hyperbolic conservation law with flux $F(u) = \beta u$, where a fluid with density $u$ is transported or advected by some constant velocity $\beta \in \mathbb{R}$. For larger values of $\beta$, the shock moves faster.

\paragraph{Exact Solution.} Here we consider the test case with the following initial and boundary conditions:
\[
\begin{aligned}
    u(0,x) &=h(x)=\1_{x \le 0.5}, \forall x\in \Omega = [0,1], \\
    u(t,0) &= g(t,0) = 1, \forall t \in [0,T], \\
    u(t,1) &= g(t,1) = 0, \forall t \in [0,T], \\
\end{aligned}
\]
respectively, and $\1_{\mathcal{E}}$ denotes an indicator function for event $\mathcal{E}$.  Note that the linear advection (convection) problem is also studied in \citet{krishnapriyanCharacterizingPossibleFailure2021b} with smooth $h(x) = \sin(x)$ and periodic boundary conditions. Here we consider the more challenging case, where the initial condition is already a shock.

In our case, the exact solution,
\[
u(t,x) = h(x-\beta t),
\] is simply the initial condition shifted to the right, which is a shock wave traveling to the right with speed $\beta > 0$. 
\paragraph{Global Conservation.}  We write the linear conservation constraint in \autoref{eqn:global_conserv_mb_app} for linear advection  as:
\begin{equation}
\mathcal{G}u(t,x) = \int _{x_0}^{x_N} u(t,x)  dx = \frac{1}{2} + \beta t = b(t).
\end{equation}
The out-flux $F_{\text{out}} = u(t,1)= g(t,1) =0$, by the fixed right Dirichlet boundary condition, and we are left with the following terms:
\[
\begin{aligned}
 \int_{\Omega} u(t,x)d\Omega  &=  \int_{\Omega} h(x) dx + \int_{0}^{t}F_{\text{in}}(t)dt \\
 &=\int_0^{0.5} dx + \beta \int_0^t u(t,0) dt\\
 &=\frac{1}{2} + \beta t,
 \end{aligned}
\]
by using the Dirichlet boundary condition $u(t,0) = g(t,0)=1$ in the second term in the last step. We see that the time rate of change in total mass is constant over time.

\subsubsection{Burgers' Equation}
Burgers' Equation, given as: 
\begin{equation}
    u_t + \frac{1}{2}(u^2)_x = 0,
    \label{eqn:burgers}
\end{equation}
is a commonly used nonlinear hyperbolic conservation law with flux $F(u) = \frac{1}{2}u^2$.
Among other things, it is used in traffic modeling. 

\paragraph{Exact Solution.} We consider the test case from \citet{osti_1395816}, where $a=1$, with the following initial and boundary conditions:
\[
\begin{aligned}
    u(0,x) &= h(x) = \begin{cases}
a, \hspace{0.1cm} x \le -1, \\
-ax, \hspace{0.1cm} -1 \le x \le 0,\\
0, \hspace{0.1cm} x \ge 0,
\end{cases}
    \forall x\in \Omega = [-1,1], \\
    u(t,-1) &= g(t, -1) = a, \forall t \in [0,T], \\
    u(t,1) &= g(t, 1) = 0, \forall t \in [0,T], \\
\end{aligned}
\]
respectively for constant, positive parameter slope $a \ge 1$. For larger values of $a$, the slope of the initial condition is steeper, and a shock is formed faster.

We write the nonlinear Burgers' \autoref{eqn:burgers} in non-conservative form as 
\[
    u_t + uu_x = 0.
\]
We see that this is the advection \autoref{eqn:advection}  with speed $\beta=u$. Hence, similarly the exact solution is given by $u(t,x) = h(x-ut)$ when the characteristics curves do not intersect, by using the method of characteristics \citep{evans2010partial}. 
We then obtain the following solution:
\[
\begin{aligned}
    u(t,x)  = \begin{cases}
a, \hspace{0.1cm} x-ut \le -1, \\
-a(x-ut), \hspace{0.1cm} -1 \le x-ut \le 0,\\
0, \hspace{0.1cm} x-ut \ge 0.
\end{cases}
\end{aligned}
\]
We use the second case to solve this implicit equation explicitly for $u$, i.e., $u = -a(x-ut) \iff u = \frac{-ax}{1-at}$. Then $x-ut = \frac{x}{1-at}$, where the denominator $1-at > 0$ for $t < 1/a$. We then solve the inequalities and  substitute this in to obtain:
\[
\begin{aligned}
    u(t,x)  = \begin{cases}
a, \hspace{0.75cm} x \le at-1, \\
\frac{ax}{at-1}, \hspace{0.45cm} at-1 \le x \le 0,  \\
0, \hspace{0.75cm} x \ge 0,
\end{cases}
\end{aligned}
\]
for $ 0 \le t < 1/a$. We see that as time increases the linear part of the solution self-sharpens with a steeper slope until the characteristics intersect at breaking time 
\[
t_b = \frac{-1}{\inf_x h'(x)} = 1/a,
\]
and a shock is formed. This is known as the waiting time phenomenon \citep{maddix2018temp_oscill}.  
  The rightward moving shock forms with weak solution given as:
\[
\begin{aligned}
    u(t,x)  = \begin{cases}
a, \hspace{0.1cm} x \le \frac{1}{2}(at-1),\\
0, \hspace{0.1cm} x \ge \frac{1}{2}(at-1),
\end{cases}
\end{aligned}
\]
for $t \ge 1/a$.
The shock speed $x'(t)$ is given by the Rankine-Hugoniot (RH) condition \citep{evans2010partial}. The RH condition simplifies for Burgers' Equation as follows: 
\[
x'(t) = \frac{f(u_R) - f(u_L)}{u_R-u_L} = \frac{1}{2}\frac{u_R^2 - u_L^2}{u_R-u_L}=\frac{1}{2}\frac{(u_R - u_L)(u_R+u_L)}{u_R-u_L} 
 = \frac{u_R + u_L}{2} = \frac{a}{2},
 \]
 where $u_L = a$ denotes the solution value to the left of the shock and $u_R = 0$ denotes the solution value to the right of the shock. Lastly, to obtain the shock position $x(t)$, we solve the simple ODE $x'(t) = a/2$ with initial condition $x(t_b=1/a) = 0$ to obtain $x(t) = \frac{at}{2} + c$, where $x(1/a) = \frac{1}{2} +c = 0$, and so $c=-\frac{1}{2}$. This results in $x(t) = \frac{1}{2}(at-1)$, as desired.

\paragraph{Global Conservation.}
We write the linear conservation constraint in \autoref{eqn:global_conserv_mb_app} for Burgers' equation~as:
\begin{equation}
\mathcal{G}u(t,x) = \int _{x_0}^{x_N} u(t,x)  dx = \frac{a}{2}(1 + at) = b(t).
\end{equation}
The out-flux is $F_{\text{out}} = \frac{1}{2}u(t,1)^2 = \frac{1}{2}g(t,1)^2 = 0$, by the fixed right Dirichlet boundary condition, and we are left with the following terms:
\[
\begin{aligned}
 \int_{\Omega} u(t,x)d\Omega  &=  \int_{\Omega} h(x) dx + \int_{0}^{t}F_{\text{in}}(t)dt \\
 &=-a\int_{-1}^{0} x dx + \frac{1}{2}\int_0^t u(t,-1)^2 dt\\
 &=\frac{a}{2}(1 + at),
 \end{aligned}
\]
by using the Dirichlet boundary condition $u(t,-1) = g(t,-1)=a$ in the second term in the last step. We again see that the time rate of change in total mass is constant over time.

\section{Discretizations of the Integral Operator $\mathcal{G}$ for Conservation and Additional Linear Constraints} 
\label{sec:global_conserv_discrete}
In this section, we first describe common discretization schemes $G$ for the integral operator $\mathcal{G}$ in \autoref{eq:linear_constraint} given as:
\begin{equation}
    \mathcal{G}u(t,x) 
    = \int_{\Omega} u(t,x)d\Omega 
    = b(t),
    \label{eq:linear_constraint_app}
\end{equation}
to form a linear matrix constraint equation $Gu=b$. Then, we show how to incorporate other types of linear constraints into our framework \probconservnosp.
In particular, we consider artificial diffusion, which is a common numerical technique to smooth numerical artifacts through the matrix $\tilde G$ arising from the second order central finite difference scheme 
of the second derivative.

\subsection{Discretizations of the Integral Operator $\mathcal{G}$}
\label{app:integral_discret}
Here, we provide examples of the discrete matrix $G \in \mathbb{R}^{T\times MT}$, which approximates the continuous integral operator $\mathcal{G}$ in \autoref{eq:linear_constraint_app}.
We use $M$ to denote the number of spatial points, $T$ to denote the number of time points, and we set $N=MT$.

We form a discrete linear system from the continuous integral conservation law, i.e,. $Gu=b$, where each row $i$ of $G$ acts as a Riemann approximation to the integral $\mathcal{G}u(t,x)$ at time $t_i$.  At inference time, we assume we have an ordered output grid $\{(t_1, x_1), \dots, (t_1, x_M), \dots, (t_T, x_1), \dots, (t_T, x_M)\}$ with spatial grid spacing $\Delta x_j = x_{j+1}-x_j$ for $j = 1, \dots, M-1$. We want to compute the solution at these corresponding grid points given as: 
\[
u = [u(t_1, x_1), \dots, u(t_1, x_M), \dots, u(t_T, x_1), \dots, u(t_T, x_M)]^T \in \mathbb{R}^{MT}.
\]
The known right-hand side is given as:
\[
b = [b(t_1), \dots, b(t_T)]^T \in \mathbb{R}^T.
\]

We now proceed to provide examples of specific matrices $G$ corresponding to common numerical spatial integration schemes \citep{burden_num_anal}. 

\paragraph{Left Riemann Sum.} For $G$ arising from the common first-order left Riemann sum 
\[
\sum_{j=1}^{M-1}u(t_i,x_j)\Delta x_j,
\]
at time $t_i$, we have the following expression:
\[
    G_{ij} =
     \begin{dcases}
     \Delta x_j, \hspace{0.1cm} (i-1)M+1 \le j \le iM-1, \\
    0, \hspace{0.1cm} \text{otherwise}.
    \end{dcases}
\]
In other words, it uses the left function value $u(t,x_j)$ on the interval $[x_j, x_{j+1}]$.  
 The right Riemann sum ($\sum_{j=2}^{M}u(t,x_j)\Delta x_{j-1}$ at time $t$) is a simple extension that shifts the column indices by 1  to $(i-1)M+2 \le j \le iM$ to use the right value $u(t_i, x_{j+1})$ on the interval $[x_j, x_{j+1}]$.

\paragraph{Trapezoidal Rule.} For $G$ arising from the second order trapezoidal rule 
\[
Gu = \sum_{j=1}^{M-1}\frac{u(t_i,x_j) +u(t_i,x_{j+1})}{2}\Delta x_j,
\] at time $t_i$, we have the following expression: 
\begin{equation}
    G_{ij} =
     \begin{dcases}
     \frac{\Delta x_j}{2}, \hspace{0.1cm} j=(i-1)M+1, \\
     \frac{\Delta x_{j-1} + \Delta x_{j}}{2}, \hspace{0.1cm} (i-1)M+2 \le j \le iM-1, \\
     \frac{\Delta x_{j-1}}{2}, \hspace{0.1cm}  j = iM, \\
    0, \hspace{0.1cm} \text{otherwise}.
    \end{dcases}
    \label{eqn:trap_rule_G}
\end{equation}
We use the trapezoidal discretization of $G$ in \autoref{eqn:trap_rule_G} in our experiments. Note that higher order schemes, e.g., Simpson's Rule may also be used, as well as more advanced numerical techniques.
These can help to reduce the error in the spatial integration approximation, including shock tracking schemes in \citet{maddix2018temp_oscill} on the more challenging sharper problems with shocks that we see for high values of $m$ in the PME and Stefan.

\subsection{Adding Artificial Diffusion into the Discretization}
\label{app:artificial_diff} 
In addition to various discretization schemes to compute the integral operator $\mathcal{G}$, our \probconserv framework can incorporate other inductive biases based on the knowledge of the underlying PDE, e.g.,  to bypass undesirable numerical artifacts. One common technique that has been used widely in numerical methods for this purpose is adding artificial diffusion \citep{maddix2018_harmonic_avg}. This artificial diffusion can act locally at sharp corners such as shock interfaces, where numerical methods tend to suffer from high frequency oscillations. Other common numerical methods to avoid numerical oscillations include total variation diminishing (TVD), i.e., $\text{TV}(u(t_{i+1}, x)) \le \text{TV}(u(t_i, x))$, $\forall i$, or  total variation bounded (TVB), i.e., $\text{TV}(u(t_{i+1}, x)) \le C$, $C > 0$, $\forall i$, where $\text{TV}(u) = \int_\Omega |\frac{\partial u}{\partial x}| d\Omega$ and is approximated as $\sum_{j=1}^{M-1} |u(t_i, x_{j+1})-u(t_i, x_j)|$ \citep{leveque1990numerical, osti_1395816}.  Note that enforcing these inequality constraints is a direction of future work.

In machine learning, artificial diffusion is analogous to adding a regularization penalty on the $L_2$ norm of the second derivative $\int \{\frac{\partial^2}{\partial x^2}u (t_i, x) \}^2dx$ \citep{hastie2013elements}.
This can be written as the $L_2$ norm of a linear operator applied to $u$, $\|\tilde {\mathcal G} (u)\|_2^2$, where $\tilde {\mathcal G} (u) (t_i) \coloneqq \frac{\partial^2}{\partial x^2} u(t_i, x)$.
Thus, we can incorporate this penalty term into \probconserv in the same manner as the integral operators by discretizing $\tilde {\mathcal G}$ via a matrix $\tilde G$.
Let $\tilde G$ be the second order central finite difference three-point stencil at time $t_i$ over $M$ spatial points:
\[
\begin{split}
    (\tilde G u)_j &= \left( \frac {u(t_i, x_{j+2}) - u(t_i, x_{j+1})}{\Delta x_{j+1}} \right)  - \left( \frac{u(t_i, x_{j+1}) - u(t_i, x_{j})}{\Delta x_{j}} \right). \\
\end{split}
\]
for $j = 1, \dots, M-2$.
For simplicity of notation, we assume $\Delta x_j \coloneqq \Delta x$ for all $x_j$, though this need not be the case in general. 
This results in the following three-banded matrix: 
\begin{equation} 
\begin{split}
\tilde G &= \frac 1 {\Delta x}  \begin{bmatrix}
1 & -2 & 1 & 0 & \dots \\
0 & 1 & -2 & 1 & \dots \\
\vdots & \vdots & \vdots & \vdots & \vdots \\
\end{bmatrix}.
\end{split}
\label{eqn:diffusion_matrix_APPENDIX}
\end{equation}
Since our goal is to penalize large differences in the solution, we set the constraint value $b$ to zero:
\[
\tilde G u + \sigma_{\tilde G} \epsilon = 0,
\]
where $\sigma_{\tilde G} > 0$ denotes the constraint value for the artificial diffusion. 
Since the mechanism is exactly the same with a linear constraint, artificial diffusion can be applied using \autoref{eqn:updated_mean_var_MEAN} with $b=0$, where $\tilde \mu$ and $\tilde \Sigma$ are the mean and covariance after applying the conservation constraint as follows:
\[
\begin{split}
\tilde \mu_{\text{diffusion}} &= \tilde \mu - \tilde \Sigma \tilde G^T (\sigma^2_{\tilde G} I + G \tilde \Sigma G^T)^{-1}(\tilde G \tilde \mu), \\
\tilde \Sigma_{\text{diffusion}} &= \tilde \Sigma - \tilde \Sigma \tilde G^T (\sigma^2_{\tilde G} I + G \tilde \Sigma G^T)^{-1}(\tilde G \tilde \Sigma).
\end{split}
\]
Moreover, the guarantees of Theorem \autoref{thm:constraint_to_zero} still hold.
Smaller values of $\sigma_{\tilde G}$ lead to smaller values of $\|\tilde G \tilde \mu_{\text{diffusion}}\|_2^2$, which results in a smoother solution. 
 
Unlike the case of enforcing conservation, it is typically not desirable when applying artificial diffusion to set $\sigma_{\tilde G}$ to zero, as this will lead to a simple line fit \citep{hastie2013elements}.
We set the variance for each row of $\tilde{G}$ as follows:
Let $\sigma_i^2 \coloneqq \text{Var}(u_n)$ be the variance of target value $u_n$ from the Step~1 procedure:
\[
\begin{aligned}
\sigma_{\tilde G, i}^2 \coloneqq \text{Var}((\tilde G u)_i) = \text{Var}(u_i - 2 u_{i+1} + u_{i+2}) = \sigma_{i}^2 + 4 \sigma_{i+1}^2 + \sigma_{i+2}^2 - 4\rho \left(\sigma_i \sigma_{i+1} + \sigma_{i+1} \sigma_{i+2} \right) + 2 \rho^2 \sigma_i \sigma_{i+2} ,
\end{aligned}
\]
where $\rho \in [0, 1]$ determines the level of auto-correlation between neighboring points.
Higher values of $\rho$ lead to lower values of $\sigma^2_{\tilde G, i}$, and hence a higher penalty.

\section{Control on Conservation Constraint}
\label{app:noise_param}
\begin{figure}[H]
    \centering
    \vskip 0.1in
    \subfigure[Norm of the conservation error:  $\text{CE}^2(\tilde \mu) = \|G \tilde \mu - b \|_2^2$.]{\label{subfig:cons_wrt_precision}
    \includegraphics[width=0.65\linewidth]{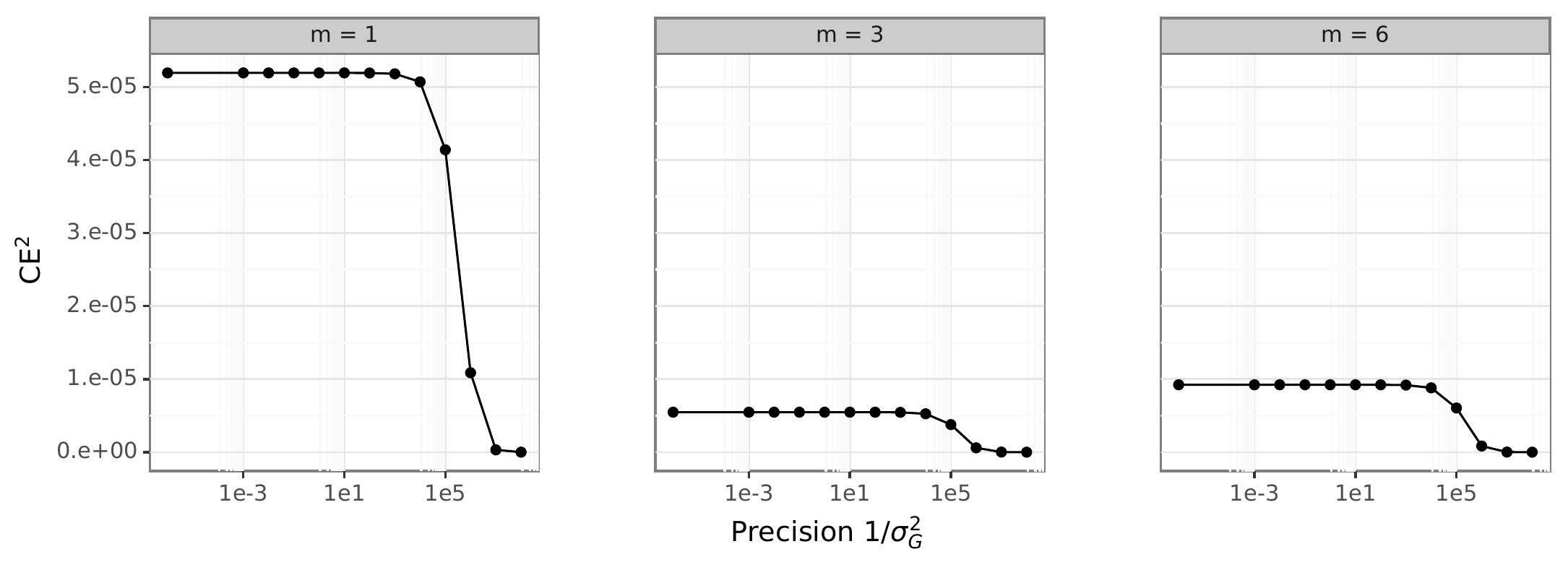}}
    \subfigure[Log-likelihood: $\text{LL}(u; \tilde \mu, \tilde \Sigma) = {-\frac 1 {2M} \|u_{t_j, \cdot} -  \tilde \mu_{t_j, \cdot}\|_{\tilde \Sigma^{-1}_{t_j}} - \frac{1}{2M} \sum_i \log \tilde \sigma^2_{t_j, i} - \log 2 \pi}$.]{\label{subfig:ll_wrt_precision}
    \includegraphics[width=0.65\linewidth]{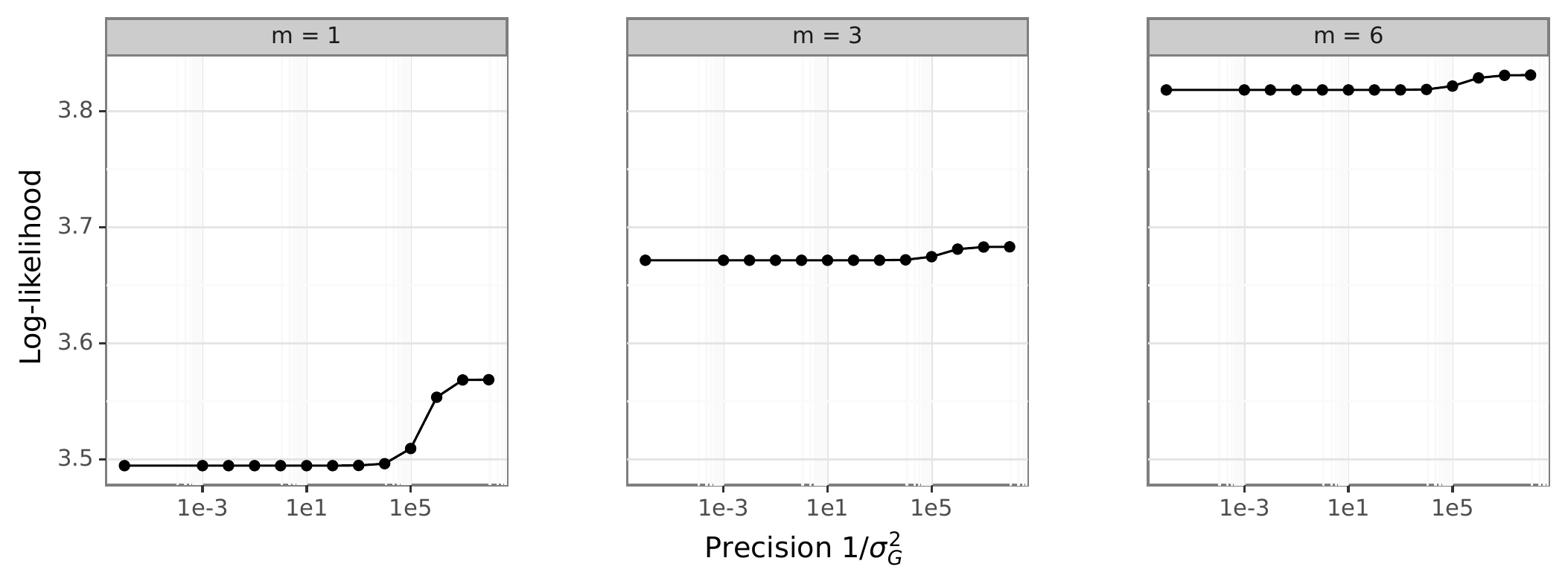}}
    \subfigure[Mean-squared error: $\text{MSE}(u, \tilde \mu) =  \frac{1}{M} \|u_{t_j, \cdot} - \tilde \mu_{t_j, \cdot}\|^2_2$.]{\label{subfig:mse_wrt_precision}
    \includegraphics[width=0.65\linewidth]{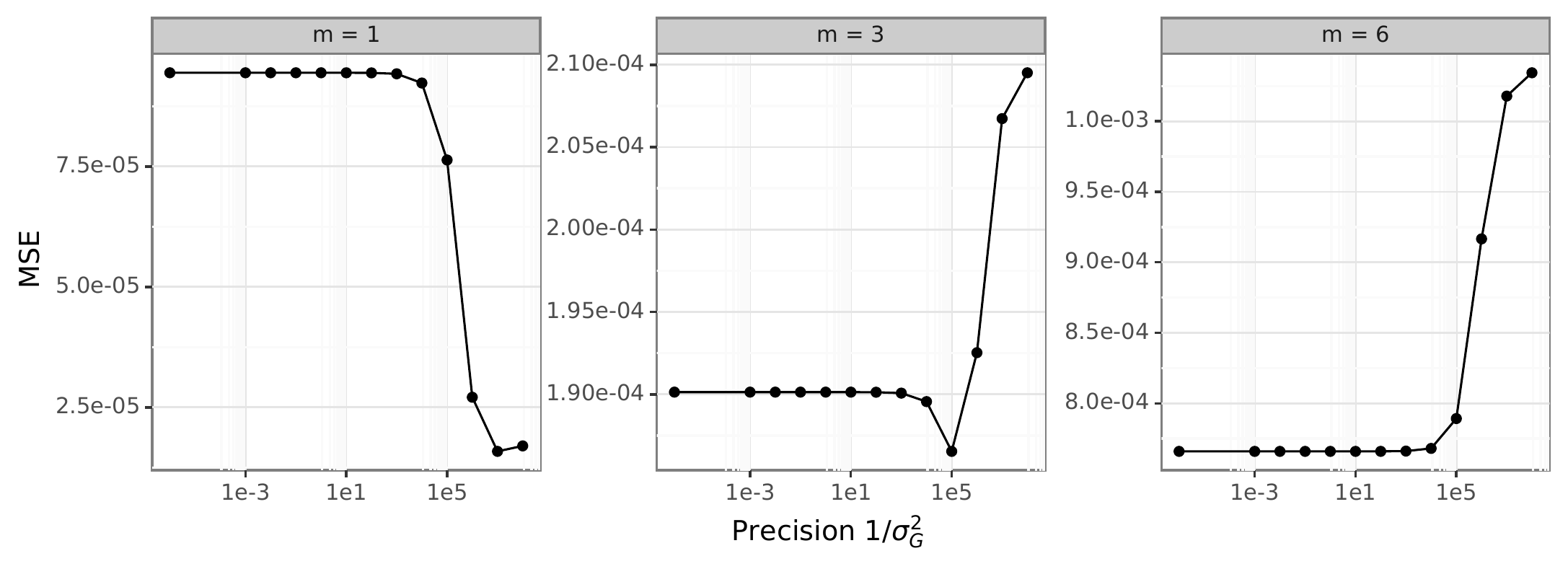}}
    \caption{Illustration of the norm of the conservation error $\text{CE}^2$ (lower is better) in the top row, the predictive log likelihood (LL) in the middle row (higher is better), and the mean-squared error (MSE) (lower is better) in the bottom row, as a function of the constraint precision $\frac 1 {\sigma_G^2}$ for \physnp on the PME in \autoref{scn:empirical_medium}, where $M$ denotes the number of spatial points, $\tilde \sigma^2_{t_j, \cdot}$ denotes the diagonal of the covariance $\tilde \Sigma_{t_j} \in \mathbb{R}^{M\times M}$ and $t_j$ denotes the time-index in the training window at which the metrics are reported. Each column indicates results for a different values of PDE parameter $m \in \{1, 3, 6\}$,
    corresponding to ``easy'', ``medium'', and ``hard'' scenarios, respectively. 
    In all three cases, $\text{CE}^2$ monotonically decreases to zero and LL monotonically increases as $\sigma_G^2 \to 0$ ($1 / \sigma_G^2 \rightarrow \infty$), illustrating Theorem \ref{thm:constraint_to_zero}.
    The biggest gains in log-likelihood are for $m=1$, where conservation was also violated the most.
    In contrast, the relationship between MSE and $\frac 1 {\sigma_G^2}$ is not guaranteed to be monotonic, and it qualitatively changes, depending on the value of $m$.
    }\label{fig:pme_pred_wrt_precision}
    \vskip -0.1in
\end{figure}

\autoref{fig:pme_pred_wrt_precision} illustrates that Theorem \ref{thm:constraint_to_zero} holds empirically for \physnp on the PME in \autoref{scn:empirical_medium}, where both the norm of the conservation error ($\text{CE}^2$) monotonically decreases to zero and the predictive log likelihood (LL) monotonically increases as the constraint precision $\sigma_G^2 \rightarrow 0$ ($1 / \sigma_G^2 \rightarrow \infty$). For the MSE, the trend depends on the difficulty of the problem. For ``easy'' scenarios, where $m=1$, the MSE also monotonically improves (decreases) as $\sigma_G^2 \rightarrow 0$ ($1 / \sigma_G^2 \rightarrow \infty$). For ``medium'' difficulty problems, where $m=3$, we see that there is an optimal value for $\sigma_G^2$ around $10^{-5}$, and enforcing the constraint exactly does not result in the lowest MSE. For the ``harder'' $m=6$ case, we see that a looser tolerance on the constraint results in better MSE. In this case the solution is non-physical since it does not satisfy conservation. Note that in the sharper $m=6$ case, the accuracy may be able to be improved by using more advanced approximations for the integral operator $\mathcal{G}$ that take the sharp corners in the solution into account \citep{maddix2018temp_oscill}.

\section{Derivation of Constrained Mean and Covariance} 
\label{app:deriv_mean_cov_limit}

In this section, 
we provide two interpretations for the Step~2 procedure of \probconserv from \autoref{eqn:updated_mean_var} given as:
\begin{subequations}
\label{eqn:updated_mean_var_app}
\begin{align}
   \label{eqn:updated_mean_var_MEAN_app}
   \tilde \mu &= \mu - \Sigma G^T (\sigma_G^2 I + G \Sigma G^T)^{-1} (G\mu - b), \\
   \label{eqn:updated_mean_var_VAR_app}
   \tilde \Sigma &= \Sigma - \Sigma G^T (\sigma_G^2 I + G \Sigma G^T)^{-1} G \Sigma.
\end{align}
\end{subequations}
While \autoref{eqn:updated_mean_var} is well-defined in the case that $\sigma_G^2 = 0$, for simplicity we assume $\sigma_G^2 > 0$ throughout this section. 
In Lemma \ref{lemma:step2bayesian}, we show how Step 2 is justified as a Bayesian update of the unconstrained normal distribution from Step 1 by adding information about the conservation constraint contained in \autoref{eqn:normal_constraint}, i.e., $b = Gu + \sigma_G \epsilon$ in Step 2.
In Lemma \ref{lemma:step2numerical}, we show how the posterior mean $\tilde \mu$ and $\tilde \Sigma$ can be re-expressed in a numerically stable and computationally efficient form given in \autoref{eqn:updated_mean_var_app}.
Finally, Lemma \ref{lemma:step2constrained} shows that this is equivalent to a least-squares optimization with an upper bound on the conservation error.

Note:  $\mu \in \mathbb{R}^{MT}, \Sigma \in \mathbb{R}^{MT \times MT}, G \in \mathbb{R}^{T \times MT}, b \in \mathbb{R}^T$, where $N=MT$ denotes the number of spatio-temporal output points, $M$ denotes the number of spatial points and $T$ denotes number of constraints or in this case time steps to enforce the conservation constraint. 

\begin{lemma}[Step 2 as Bayesian update] \label{lemma:step2bayesian}
    Assume the predictive distribution of $u$ conditioned only on observed data $D$ is normal with mean $\mu$ and covariance $\Sigma$. 
    Let $b$ be the known conservation quantity that follows a normal distribution with mean $G u$ and covariance $\sigma_G^2 I$, where $\sigma_G^2 > 0$.
    Then the posterior distribution of $u$ conditional on both data $D$ and conservation quantity $b$ is normal with mean $\tilde \mu$ and covariance $\tilde \Sigma$ given as:
    \[
    \begin{split}
    u | b, D &\sim \mathcal N(\tilde \mu, \tilde \Sigma), \\
    \tilde \Sigma &= A^{-1} \Sigma, \\
    \tilde \mu &= A^{-1} (\mu + \frac{1}{\sigma_G^2} \Sigma G^T b),
    \end{split}
    \]
\end{lemma}
where  $A = I + \frac{1}{\sigma_G^2}\Sigma G^TG.$

\paragraph*{Proof}
This follows the same logic as a standard multivariate normal model with known covariance; see Chapter 3.5 of \citet{gelmanBayesianDataAnalysis2015}.
We outline the derivation below.
Note that we mark the terms that are independent of the unknown $u$ as constants.
\begin{equation*}
    \begin{aligned}
        \log p(u | b, D) &= 
        \log \underbrace{p (u \vert D)}_{\text{Step 1}}\underbrace{p(b \vert u)}_{\text{Step 2}} - \log \int p(b | u) dp(u | D) \hspace{.1cm}\text{(Bayes' Rule)}
        \\ 
        & = \log p(u \vert D) + \log p(b \vert u) + C_1 
 \\
        &=\log \mathcal{N} (u; \mu, \Sigma) + \log \mathcal{N}(b; Gu, \sigma_G^2 I) + C_1 \\
        \\
        &= - \frac 1 2 \left (\|u - \mu \|_{\Sigma^{-1}}^2 + \frac{1}{\sigma_G^2} \|Gu - b\|_{2}^2 \right) + C_2 
        \\
        &= - \frac 1 2 \left( u^T \Sigma^{-1} u - 2 u^T \Sigma^{-1} \mu + u^T (\frac{1}{\sigma_G^2}G^T G) u - 2 u^T \frac{1}{\sigma_G^2}G^T b \right) + C_3
        \\
        &= - \frac 1 2 \left(u^T (\Sigma^{-1} + \frac{1}{\sigma_G^2} G^T G) u - 2 u^T (\Sigma^{-1} \mu + \frac{1}{\sigma_G^2} G^Tb) \right) + C_3 \\
        &= - \frac 1 2 \bigg(u^T \underbrace{(\Sigma^{-1} + \frac{1}{\sigma_G^2} G^T G)}_{\tilde \Sigma ^{-1}} u  - 2 u^T \underbrace{(\Sigma^{-1} + \frac 1 {\sigma_G^2} G^T G)}_{\tilde \Sigma ^{-1}} \underbrace{(\Sigma^{-1} + \frac 1 {\sigma_G^2} G^T G)^{-1} (\Sigma^{-1} \mu + \frac 1 {\sigma_G^2} G^Tb)}_{\tilde \mu} \bigg) + C_3 \\
        & = - \frac 1 2 \left (u^T \tilde \Sigma^{-1} u - 2 u^T \tilde \Sigma^{-1} \tilde \mu \right) + C_3 \\
        &= \log \mathcal{N} (u; \tilde \mu, \tilde \Sigma) + C_4,
    \end{aligned}
\end{equation*}
where 
\begin{subequations}
\begin{align}
\tilde \Sigma &= (\Sigma^{-1} + \frac{1}{\sigma_G^2}G^TG)^{-1} = (I + \frac{1}{\sigma_G^2}\Sigma G^TG)^{-1} \Sigma = A^{-1} \Sigma,
\vspace{0.1cm}
\\
\tilde \mu &= (\Sigma^{-1} + \frac{1}{\sigma_G^2}G^T G)^{-1}(\Sigma^{-1} \mu + \frac{1}{\sigma_G^2}G^T b)
 = \tilde \Sigma  (\Sigma^{-1} \mu + \frac{1}{\sigma_G^2}G^Tb) \\
 &= A^{-1} \Sigma  (\Sigma^{-1} \mu + \frac{1}{\sigma_G^2} G^T b)
= A^{-1} (\mu + \frac{1}{\sigma_G^2}\Sigma G^T b),
\label{eqn:tidle_mu}\\
C_1 &= -\log \int p(b \vert u) dp (u \vert D), \\
C_2 &= C_1 -\frac{1}{2} \left(MT \log 2 \pi + \log \det \Sigma + T \log \pi + \log \sigma_G^2 \right), \\
C_3 &= C_2 -\frac{1}{2} \left( \mu^T \Sigma^{-1} \mu + \frac 1 {\sigma_G^2} b^T b \right), \\
C_4 &= 0. 
\end{align}
\end{subequations}
Note that $C_4 = 0$ since the left-hand side and right-hand side are log-probability densities, so we have the desired expression. $\square$

\begin{lemma}[Numerically stable form for Step 2]\label{lemma:step2numerical}
Assume that $\sigma_G^2 > 0$. The posterior mean and covariance $\tilde \mu$ and $\tilde \Sigma$ can be written in a numerically stable form as:
    \[
    \begin{split}
   \tilde \mu &= \mu - \Sigma G^T (\sigma_G^2 I + G \Sigma G^T)^{-1} (G\mu - b), \\
   \tilde \Sigma &= \Sigma - \Sigma G^T (\sigma_G^2 I + G \Sigma G^T)^{-1} G \Sigma.
    \end{split}
    \]
\end{lemma}
\paragraph*{Proof}
We use the following two Searle identities (corollaries of the Woodbury identity) \citep{petersen2008matrix}:
\begin{subequations}
\begin{align}
(I + CB)^{-1} &= I - C(I + BC)^{-1} B, \label{eqn:searle1}\\
(C + B B^T)^{-1}B &= C^{-1}B(I + B^T C^{-1} B)^{-1}, \label{eqn:searle2}
\end{align}
\end{subequations}
for some matrices $B, C$.
Using \autoref{eqn:searle1}, we re-write $A^{-1}$:
\begin{subequations}
\begin{align}
   A^{-1} &= (I + \frac{1}{\sigma^2_G} \Sigma G^T G)^{-1} \\
   &= I - \Sigma G^T (I + \frac {1}{\sigma^2_G} G \Sigma G^T)^{-1} \frac {1}{\sigma_G^2} G \\
   &= I - \Sigma G^T (\sigma^2_G I + G \Sigma G^T)^{-1} G.
   \label{eqn:a_inv_stable}
\end{align}
\end{subequations}
The desired expression for $\tilde \Sigma$ immediately follows by combining \autoref{eqn:a_inv_stable} with Lemma \ref{lemma:step2bayesian}.
For $\tilde \mu$, we break the expression into two parts, and then use the Searle identity shown in \autoref{eqn:searle2} as follows:
\begin{subequations}
\begin{align}
   \tilde \mu &= A^{-1} (\mu + \frac{1}{\sigma^2_G} \Sigma G^T b) \\
   &= A^{-1} \mu + A^{-1} \frac{1}{\sigma^2_G} \Sigma G^T b,  \\
A^{-1} \mu &= (I - \Sigma G^T (\sigma^2_G I + G \Sigma G^T)^{-1} G) \mu, \label{eqn:final_mu0} \\
A^{-1} \frac 1 {\sigma^2_G} \Sigma G^T b &= (\Sigma^{-1} + \frac 1 {\sigma^2_G} G^T G)^{-1} \frac{1}{\sigma_G^2}G^T b \label{eq:nonzero_before_searle} \\
&=  \frac{1}{\sigma^2_G} \Sigma G^T (I + \frac{1}{\sigma_G^2} G \Sigma G^T)^{-1} b \label{eq:nonzero_after_searle} \\
&= \Sigma G^T (\sigma^2_G I + G \Sigma G^T)^{-1} b. \label{eqn:final_b}
\end{align}
\end{subequations}
Adding the expressions in \autoref{eqn:final_mu0} and \autoref{eqn:final_b} yields the desired form for $\tilde \mu$.
$\square$

Observe that the matrix $\sigma^2_G I + G \Sigma G^T \in \mathbb{R}^{T \times T}$ is invertible for all values of $\sigma^2_G$ (including zero), since it is square in the smaller dimension and has full rank $T$.
In addition, inverting $\sigma^2_G I + G \Sigma G^T \in \mathbb{R}^{T \times T}$ has reduced computational complexity compared to inverting $A$.

\begin{lemma}[Solution to constrained optimization] \label{lemma:step2constrained}
The expression for the posterior mean $\tilde \mu$ with $\sigma_G^2 > 0$ is equivalent to solving the following constrained least-squares problem for some value of $c > 0$:
\[
\tilde \mu = \mathrm{argmin}_{y} \frac 1 2 \|y - \mu\|_{\Sigma^{-1}}^{2},
\] subject to $\frac 1 2 \| Gy - b \|_2^2 < c$, where $c < \frac 1 2 \| G\mu - b \|_2^2$.
\end{lemma}
\paragraph{Proof} 
This is a standard result from ridge regression \citep{hastie2013elements}. 

Since $c < \frac 1 2 \| G\mu - b \|_2^2$, the complementary slackness condition requires that $c = \frac 1 2 \| Gy - b \|_2^2$.
Thus, we get the following Lagrangian:
$$
L(y, \lambda) = \frac 1 2 \|y - \mu\|_{\Sigma^{-1}}^{2} + \lambda \left( \frac 1 2 \| Gy - b \|_2^2 - c \right).
$$
Observe that, if we re-label $y \coloneqq u$ and $\lambda \coloneqq 1/\sigma_G^2$, then $L(y, \lambda)$ is equal to $- \log p(u | b, D) + C_2$, where $C_2$ is a constant with respect to $y$.
Thus, the optimal value of $y$ is the posterior mean from \autoref{eqn:updated_mean_var_MEAN_app}, i.e.,
$$
\nabla_y L(y, \lambda) = 0 \iff y = \tilde \mu,
$$
where
$$
\tilde \mu = \mu - \Sigma G^T (\frac{1}{\lambda} I + G \Sigma G^T)(G\mu - b).
$$
Next, we substitute the above expression for $\tilde \mu$ into the remaining feasibility condition:
\[
\begin{split}
c &= \frac 1 2 \| G \tilde \mu - b \|_2^2 \\ 
&= \|G \left(\mu - \Sigma G^T (\frac 1 \lambda I + G \Sigma G^T)^{-1} (G \mu - b) \right) - b \|_2^2 \\
&= \|G \mu - G\Sigma G^T (\frac 1 \lambda I + G \Sigma G^T)^{-1} (G \mu - b) - b \|_2^2 \\
&= \|\left( I - G\Sigma G^T (\frac 1 \lambda I + G\Sigma G^T)^{-1}\right) (G \mu - b) \|_2^2.
\end{split}
\]
The eigenvalues of matrix \( I - G\Sigma G^T [(1/\lambda)I + G\Sigma G^T]^{-1}\) shrink to $0$ as $1/ \lambda \to 0$.
This establishes that $c$ and $1/\lambda$ have a monotonic relationship.
Hence, one can find a value of $c$ such that $\lambda = 1/\sigma^2_G$.

\section{Proof of Theorem \ref{thm:constraint_to_zero}}

\label{app:limiting_sol}
In this section, we provide the proof for Theorem \ref{thm:constraint_to_zero}.  We begin by first restating Theorem \ref{thm:constraint_to_zero}.
\constrainttozero*

\noindent
For the proof of Theorem \ref{thm:constraint_to_zero},
recall the following expression for the posterior mean from \autoref{eqn:updated_mean_var_MEAN}:
\[
\tilde \mu_n = \mu - \Sigma G^T (\sigma^2_{G, n} I + G \Sigma G^T)^{-1} (G \mu - b).
\]
\paragraph*{Proof of 1.}
Define $\tilde \mu^{\star} \equiv \mu - \Sigma G^T (G \Sigma G^T)^{-1} (G \mu - b)$.
We show that $\tilde \mu_n$ converges monotonically to $\tilde \mu^{\star}$ as follows:
\newcommand{\sigmaG}{\sigma_{G, n}}
\begin{subequations}
\begin{align}
\tilde \mu_{n} - \tilde \mu^{\star} &= \Sigma G^{T} \left [ (G \Sigma G^{T})^{-1} - (\sigmaG^{2} I + G\Sigma G^{T})^{-1} \right] (G\mu - b) \label{eqn:searle_part1} \\
&= \Sigma G^{T} \left [ (G \Sigma G)^{-1} (-\sigmaG^2 I) (-\sigmaG^2 I - G\Sigma G^T)^{-1}\right] (G\mu - b) \label{eqn:searle_part2}\\
&= \sigmaG^2 \Sigma G^{T} \left [ (G \Sigma G^T)^{-1} (\sigmaG^2 I + G\Sigma G^T)^{-1}\right] (G\mu - b) \\
&= 
\sigmaG^2 \Sigma G^{T} \left [(\sigmaG^2 I + G\Sigma G^T)(G \Sigma G^T)\right]^{-1} (G\mu - b) \\
&= 
\sigmaG^2 \Sigma G^{T} \left [ \sigmaG^2 G \Sigma G^T + (G \Sigma G^T)^2 \right]^{-1} (G\mu - b).
\end{align}
\end{subequations}
The derivation from \autoref{eqn:searle_part1} to \autoref{eqn:searle_part2}  follows from the Searle identity: 
$$C^{-1} + B^{-1} = C^{-1}(C+B)B^{-1} , $$ 
where $C=G \Sigma G^{T}$, $B =  - (\sigmaG^{2} I + G\Sigma G^{T})$, and $C+B = -\sigmaG^{2} I$.
Then,
\begin{equation}
\|\tilde \mu_{n} - \tilde \mu^{\star}\|^2_{\Sigma^{-1}} = (G\mu - b)^T Q_n (G \mu - b), 
\end{equation}
where
\begin{subequations}
\begin{align}
Q_n&\coloneqq\left [ \sigma_{G,n}^2(G\Sigma G^T) + (G \Sigma G^T)^2  \right]^{-1} \sigma^2_{G,n} G \Sigma \Sigma^{-1} \Sigma G^T \sigma^2_{G,n} \left [ \sigma_{G,n}^2(G\Sigma G^T) + (G \Sigma G^T)^2  \right]^{-1} \\
&= \left [ \sigma_{G,n}^2(G\Sigma G^T) + (G \Sigma G^T)^2  \right]^{-1} \sigma^2_{G,n} G \Sigma G^T \sigma^2_{G,n} \left [ \sigma_{G,n}^2(G\Sigma G^T) + (G \Sigma G^T)^2  \right]^{-1} \\
&= \sigma^4_{G,n} \left [ \sigma_{G,n}^2(G\Sigma G^T) + (G \Sigma G^T)^2 \right]^{-1} G \Sigma G^T \left [ \sigma_{G,n}^2(G\Sigma G^T) + (G \Sigma G^T)^2  \right]^{-1}. \label{eq:inner_matrix}
\end{align}
\end{subequations}
Let $\lambda_i, v_i$ be an eigenvalue and associated eigenvector of $G \Sigma G^T$, respectively.
First, $\lambda_i > 0$ because $G \Sigma G^T$ is symmetric positive definite. This follows from the fact that $\Sigma$ is positive definite and $G^T$ is not rank deficient.
Next, the associated eigenvector $v_i$ is also an eigenvector of matrix $\sigma_{G, n}^2 (G \Sigma G^T) + (G \Sigma G^T)^2$ with eigenvalue $\sigma_{G, n}^2 \lambda_i + \lambda_i^2$.
Therefore, $v_i$ is an eigenvector of $Q_n$ with eigenvalue:
\[
\sigma_{G, n}^4 \left(\frac{1}{\sigma_{G, n}^2 \lambda_i + \lambda_i^2} \right) \lambda_i \left(\frac{1}{\sigma_{G, n}^2 \lambda_i + \lambda_i^2} \right) = \frac{\lambda_i}{\left(\lambda_i + \sigma_{G, n}^{-2} \lambda_i^2 \right)^2}.
\]
Since all the eigenvalues are strictly decreasing as $\sigmaG \to 0$, the value $\|\tilde \mu_n - \tilde \mu^\star \|^2_{\Sigma^{-1}} = (G \mu - b)^T Q_n (G \mu - b) \downarrow 0$, as required.
$\square$

\paragraph*{Proof of 2.}
Now, we show that $\tilde \mu^{\star} = \mathrm{argmin}_{y} \|y - \mu\|_{\Sigma^{-1}}^{2}$ subject to $Gy = b$.
This constrained least-squares problem can be cast into the following constrained least-norm problem:
\[
  \mathrm{minimize} \|u\|_{2}^{2}, \text{ subject to } G\Sigma^{1/2} u = b - \Sigma^{-1/2} \mu,
\]
with the transformation
$u = \Sigma^{-\frac 1 2} (y - \mu)$ or $y = \mu + \Sigma^{\frac 1 2} u$.

The final solution is
\[
\mu - \Sigma G^T (G \Sigma G^T)^{-1} (G \mu - b),
\]
which equals $\tilde \mu^\star$.
$\square$

\paragraph*{Proof of 3.}
We show that the $L_2$ norm between the predicted conservation value and the true value, $\|G \tilde \mu_n - b\|_2^2$, converges monotonically to $0$ as $\sigma_{G, n}^2 \to 0$.
We start by substituting the expression for \autoref{eqn:updated_mean_var_MEAN}:
\begin{equation} \label{eqn:theory_cons_constraint}
\begin{split}
    G \tilde \mu_n - b &= G \mu - G \Sigma G^T (\sigma^2_{G, n} I + G \Sigma G^T)^{-1} (G \mu - b) - b \\
    &= (I -  G \Sigma G^T (\sigma^2_{G, n} I + G \Sigma G^T)^{-1}) (G \mu - b).
\end{split}
\end{equation}
Let $v_i$ be an eigenvector of $G \Sigma G^T$ and $\lambda_i$ the associated eigenvector.
Then $v_i$ is also an eigenvector of $(I -  G \Sigma G^T (\sigma^2_{G, n} I + G \Sigma G^T)^{-1})$ with eigenvalue $1 - \lambda_i/(\sigma_{G, n}^2 + \lambda_i)$.
Since all the eigenvalues are monotonically decreasing to zero as $\sigma^2_{G, n} \to 0$ monotonically, $\|G \tilde \mu_n - b \|_2^2 \downarrow 0$. 
For $\sigma^2_{G,n} = 0$, $G \tilde \mu_n - b = 0$.
$\square$
\paragraph{Proof of 4.}
Define $P \coloneqq \Sigma G^T (G \Sigma G^T)^{-1} G$, which is an oblique projection matrix since 
$$ P^2 = \Sigma G^T (G \Sigma G^T)^{-1} G \Sigma G^T (G \Sigma G^T)^{-1} G = P $$ 
and
\[
\begin{split}
    \langle x, Py \rangle_{\Sigma^{-1}} = x^T \Sigma^{-1} P y = x^T  G^T (G \Sigma G^T)^{-1} G y = x^T P^T \Sigma^{-1} y =  \langle Px, y \rangle_{\Sigma^{-1}}.
\end{split}
\]
The norm $\|\tilde \mu_n - u\|_{\Sigma^{-1}}$ can be decomposed into two parts:
\[
\begin{split}
\|\tilde \mu_n - u\|_{\Sigma^{-1}} &= \|P(\tilde \mu_n - u)\|_{\Sigma^{-1}} + \|(I-P)(\tilde \mu_n - u)\|_{\Sigma^{-1}} .
\end{split}
\]
First, we show that the second term $\|(I-P)(\tilde \mu_n - u)\|_{\Sigma^{-1}}$ equals $\|\tilde \mu^\star - u\|_{\Sigma^{-1}}$ for all $n$ as follows:
\[
\begin{split}
(I-P)\tilde \mu_n &= (I - P) \mu - (I - P) \Sigma G^T (\sigmaG^2 I + G \Sigma G^T)^{-1} (G \mu - b) \\
&= (I - P) \mu - \Sigma G^T (\sigmaG^2 I + G \Sigma G^T)^{-1} (G \mu - b) + P \Sigma G^T (\sigmaG^2 I + G \Sigma G^T)^{-1} (G \mu - b) \\
&= (I - P) \mu - \Sigma G^T (G \Sigma G^T)^{-1} (G \mu - b) + \Sigma G^T (G \Sigma G^T)^{-1} G \Sigma G^T (\sigmaG^2 I + G \Sigma G^T)^{-1} (G \mu - b) \\
&= (I - P) \mu \\
&= \tilde \mu^\star - \Sigma G^T (G \Sigma G^T)^{-1} b, \\
(I-P) u &= u - \Sigma G^T (G \Sigma G^T)^{-1}G u\\
&=  u - \Sigma G^T (G \Sigma G^T)^{-1}b, 
\end{split}
\]
Therefore, 
$$
(I-P)\tilde \mu_n - (I-P) u = \tilde \mu^\star - u .
$$
Next, we show that the first term $\|P(\tilde \mu_n - u)\|_{\Sigma^{-1}}$ is equal to the distance between $\tilde \mu_n$ and $\tilde \mu^\star$. We first compute:
\begin{subequations}
\begin{align}
P \tilde \mu_n &= P \mu - P \Sigma G^T (\sigmaG^2 I + G \Sigma G^T)^{-1} (G \mu - b) 
\label{subeq:P_u_tilde_n}\\
 &= \Sigma G^T (G \Sigma G^T)^{-1} G \mu - \Sigma G^T (\sigmaG^2 I + G \Sigma G^T)^{-1} (G \mu - b), \\
P u &= \Sigma G^T (G \Sigma G^{-1}) G u\\
&=  \Sigma G^T (G \Sigma G^{-1}) b.
\label{subeq:P_u}
\end{align}
\end{subequations}
Then subtracting \autoref{subeq:P_u} from \autoref{subeq:P_u_tilde_n} gives:
\begin{subequations}
\begin{align}
P \tilde \mu_n - P u &= \Sigma G^T (G \Sigma G^T)^{-1} G \mu - \Sigma G^T (\sigmaG^2 I + G \Sigma G^T)^{-1} (G \mu - b) - \Sigma G^T (G \Sigma G^T)^{-1} b \\
&= \left(\Sigma G^T (G \Sigma G^T)^{-1} - \Sigma G^T (\sigmaG^2 I + G \Sigma G^T)^{-1} \right) (G\mu - b) \\
&= \tilde \mu_n - \tilde \mu^\star.
\end{align}
\end{subequations}
From part 1, $\|\tilde \mu_n - \tilde \mu^\star\|_{\Sigma^{-1}} \downarrow 0$ monotonically as $\sigma^2_{G, n} \downarrow 0$.
Thus,
$$\|\tilde \mu_n - u\|_{\Sigma^{-1}}^2 = \|\tilde \mu_n - \tilde \mu^\star\|_{\Sigma^{-1}}^2 + \|\tilde \mu^\star - u\|_{\Sigma^{-1}}^2 \downarrow \|\tilde \mu^\star - u\|_{\Sigma^{-1}}^2.
$$
 $\square$
 \paragraph{Proof of 5.} 
Recall that the predictive log-likelihood (LL) is defined as:
\[
\text{LL}(u; \tilde \mu_n, \tilde \Sigma_n) = -\frac 1 {2M} \|u - \tilde\mu_n\|_{\tilde \Sigma_n^{-1}} - \frac{1}{2} \sum_i \log \tilde \Sigma_{n, i, i} - \frac{1}{2M} \log 2 \pi,
\]
where $M$ denotes the total number of points.
Also recall that the precision is well-defined as:
\[
\tilde \Sigma_n^{-1} = \Sigma^{-1} + \frac{1}{\sigma_{G, n}^2} G^T G,
\]
so the first term of the predictive likelihood can be further decomposed as:
\[
\begin{split}
\|\tilde \mu_n - u\|_{\tilde \Sigma^{-1}_n}^2 &= (\tilde \mu_n - u)^T \tilde \Sigma^{-1}_n (\tilde \mu_n - u) = (\tilde \mu_n - u)^T \Sigma^{-1} (\tilde \mu_n - u) + (\tilde \mu_n - u)^T \frac{1}{\sigmaG^2} G^T G (\tilde \mu_n - u) \\
&= \|\tilde \mu_n - u\|_{\Sigma^{-1}}^2 + \frac{1}{\sigmaG^2} \|G \tilde \mu_n - b\|_2^2 \\
&= \|\tilde \mu_n - u\|_{\Sigma^{-1}}^2 + \|\frac{1}{\sigmaG} (G \tilde \mu_n - b)\|_2^2.
\end{split}
\]
Substituting the expression from \autoref{eqn:theory_cons_constraint}, we get:
\begin{equation}
\begin{split}
    \frac{1}{\sigmaG} (G \tilde \mu_n - b) &=  \frac{1}{\sigmaG}(I - G \Sigma G^T (\sigma^2_{G, n} I + G \Sigma G^T)^{-1}) (G \mu - b).
\end{split}
\end{equation}
Let $v_i$ be an eigenvector of $G \Sigma G^T$ and $\lambda_i$ the associated eigenvector.
Then $v_i$ is also an eigenvector of ${\frac{1}{\sigmaG}(I - G \Sigma G^T (\sigma^2_{G, n} I + G \Sigma G^T)^{-1}) (G \mu - b)}$ with eigenvalue:
$$
\frac{1}{\sigmaG} \bigg(1 - \frac{\lambda_i}{\sigma_{G, n}^2 + \lambda_i}\bigg) = \frac{\sigma_{G, n}}{\sigma_{G, n}^2 + \lambda_i} = \frac{1}{\sigma_{G, n} + \lambda_i \sigma_{G, n}^{-1}}.
$$
For sufficiently small $\sigma_{G, n}$, the eigenvalues are monotonically decreasing to zero as $\sigma^2_{G, n} \to 0$.

Finally, $\log (\tilde \Sigma_n)_{i, i}$ is non-increasing with respect to $\sigma^2_{G, n}$. From \autoref{eqn:updated_mean_var_VAR},
\[
\begin{split}
\tilde \Sigma_n &= \Sigma - \Sigma G^T (\sigma^2_{G, n} I + G \Sigma G^T)^{-1} G \Sigma, \\
(\tilde \Sigma_n)_{i, i} &= \Sigma_{i, i} - e_i^T \Sigma G^T (\sigma^2_{G, n} I + G \Sigma G^T)^{-1} G \Sigma e_i,
\end{split}
\]
where $e_i$ denotes the $i$-th elementary vector.
Since $\Sigma G^T (\sigmaG^2 I + G \Sigma G^T)^{-1} G \Sigma$ is positive definite with positive diagonal entries, and the eigenvalues of $(\sigmaG^2 I + G \Sigma G^T)^{-1}$ increase monotonically as $\sigma_{G, n} \to 0$, the entry $(\tilde \Sigma_n)_{i, i}$ decreases as $\sigma_{G, n} \to 0$.

\section{Additional Details on the Generalized Porous Medium Equation} 
\label{app:overview_gpme}

In this section, we discuss in more detail the parametric Generalized Porous Medium Equation (GPME).
The GPME is a \emph{family} of conservation equations, parameterized by a nonlinear coefficient $k(u)$, and it has been used in several applications ranging from underground flow transport to nonlinear heat transfer to water desalination and beyond. 
Among other things, it has the parametric ability to represent pressure, diffusivity, conductivity, or permeability, in these and other applications \citep{vazquez2007}.
From the ML/SciML methods perspective, it has additional advantages, including closed-form self-similar solutions, structured nonlinearities, and the ability to choose the parameter $k(u)$ to interpolate between ``easy'' and ``hard'' problems (analogous to but distinct from the properties of elliptical versus parabolic versus hyperbolic~PDEs). 

\paragraph{The GPME Equation.}
The basic GPME is given as: 
\begin{equation}
u_t - \nabla \cdot (k(u) \nabla u) = 0,
\label{eqn:gpme_IN_APP}
\end{equation}
where $F(u) = -k(u)\nabla u$ is a nonlinear flux function, and where the parameter $k=k(u)$ can be varied (to model different physical phenomena, or to transition between ``easy'' PDEs and ``hard'' PDEs).   
Even though the equation appears to be parabolic, for small values of $k(u)$ in the nonlinear case, it exhibits degeneracies, and it is is called ``degenerate parabolic.''  
By varying $k$, solutions span from ``easy'' to ``hard,'' exhibiting many of the qualitative properties of smooth/nice parabolic to sharp/hard hyperbolic PDEs.
Among other things, this includes discontinuities associated with self-sharpening occurring over time, even for smooth initial conditions. 

\begin{figure*}[h]
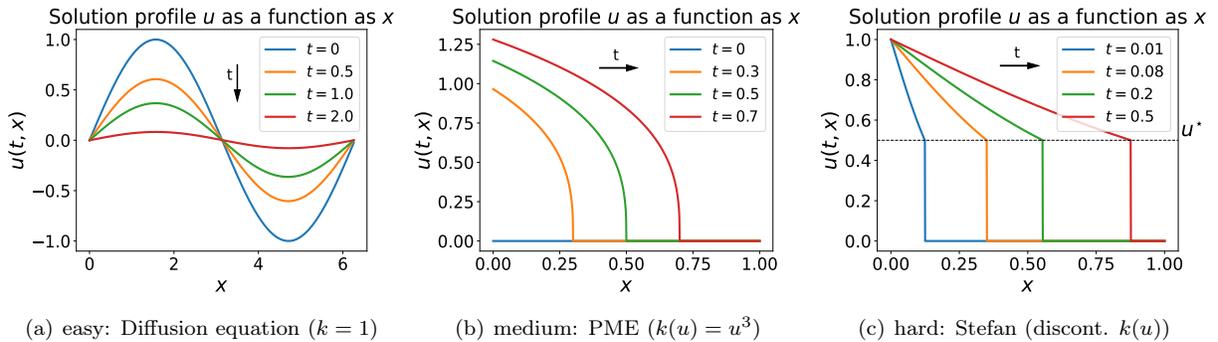

    \centering
    \vskip 0.1in
    \subfigure[easy: Diffusion equation ($k=1$)]{\label{subfig:heat_eqn_app}\includegraphics[width=0.32\linewidth, height=4cm]{fig/heat_eqtn_arrow_t.pdf}}
    \centering
    \subfigure[medium: PME ($k(u)=u^3$)]{\label{subfig:PME_app}\includegraphics[width=0.32\linewidth, height=4cm]{fig/pme_arrow_t.pdf}}
    \subfigure[hard: Stefan (discont. $k(u)$)]{\label{subfig:Stefan_app}\includegraphics[width=0.32\linewidth, height=4cm]{fig/stefan_arrow_t.pdf}}
    \caption{Illustration of the ``easy-to-hard'' paradigm for PDEs, for the GPME family of conservation equations: 
    (a) ``easy'' parabolic smooth (diffusion equation) solutions, with constant parameter $k(u)=k\equiv1$; 
    (b) ``medium'' degenerate parabolic PME solutions, with nonlinear monomial coefficient $k(u) = u^m$, with parameter $m=3$ here; and  
    (c) ``hard'' hyperbolic-like (degenerate parabolic) sharp solutions (Stefan equation) with nonlinear step-function coefficient 
    $k(u) = \1_{u \ge u^{\star}}$, where $\1_{\mathcal{E}}$ is an indicator function for event $\mathcal{E}$.
    }
    \label{fig:para_to_degenerate_app}
    \vskip -0.1in
\end{figure*}

\autoref{fig:para_to_degenerate_app} (\autoref{fig:para_to_degenerate} repeated here) provides an illustration of 
this ``easy-to-hard'' paradigm for PDEs
for the three classes of the GPME considered in the main text.
In particular, Figure \ref{subfig:heat_eqn_app} illustrates an ``easy'' situation, with $k(u) \equiv 1$, where we have a simple parabolic solution to the linear heat/diffusion equation, where a sine initial condition is gradually smoothed over time.
Figure \ref{subfig:PME_app} illustrates a situation with ``medium'' difficulty, namely the degenerate parabolic Porous Medium Equation (PME) with nonlinear differentiable monomial coefficient $k(u)=u^m$.
Here, for $m=3$, a constant zero initial condition self-sharpens, and it develops a sharp gradient that does not dissipate over time \citep{maddix2018_harmonic_avg}.
Finally, Figure \ref{subfig:Stefan_app} illustrates an example of the ``hard'' Stefan problem, where the coefficient $k(u)$ is a nonlinear discontinuous step-function of the unknown $u$ defined by the unknown value $u^{\star} = u(t, x^{\star}(t))=0.5$ at the discontinuity location $x^{\star}(t)$.
In this case, the solution evolves as a rightward moving shock or moving interface over time \citep{maddix2018temp_oscill}.  

Here, we provide more details on these and other classes of the GPME.

\paragraph{Heat/Diffusion Equation.}
Perhaps the simplest non-trivial form of the GPME, where the conductivity or diffusivity coefficient 
$$k(u) = k > 0,$$
is a constant, corresponds to the heat (or diffusion) equation. 
In this case, \autoref{eqn:gpme} reduces to the linear parabolic equation, $u_t= k \Delta u$, where $\Delta$ denotes the Laplacian operator.
Solutions of this equation are smooth due to the diffusive nature of the Laplacian operator, and even sharp initial condition are smoothed over time. 

\paragraph{Variable Coefficient Problem.} 
The linear variable coefficient problem 
$$k(u,x) = k(x),$$
is also a classical parabolic equation. The variable coefficient problem is commonly used in reservoir simulations to model the interface between permeable and impermeable materials, where $k(u)$ denotes the step-function permeabilities that depends on the spatial position $x$.

\paragraph{Porous Medium Equation (PME).}
Another subclass of the GPME, in which the coefficient 
is nonlinear but smooth, is known as the Porous Medium Equation (PME).
The PME is known to be degenerate parabolic, and it becomes more challenging as $m$ increases.
The PME with $m=1$ has been widely used to model isothermal processes, e.g., groundwater flow and population dynamics in biology. For $m>1$, the PME results in sharp solutions, and it has been used to describe adiabatic processes and nonlinear phenomena such as heat transfer of plasma (ionized gas).

\paragraph{Super-slow Diffusion Problem.} 
Another subclass of the GPME, known as super-slow diffusion, occurs when 
$$k(u) = \exp(-1/u) .$$ 
Here, 
the diffusivity $k(u)\rightarrow0$ as $u\rightarrow0$ faster than any power of $u$. 
This equation models the diffusion of solids at different absolute temperatures $u$. 
The coefficient $k(u)$ represents the mass diffusivity in this case, and it is connected with the Arrhenius law in thermodynamics.

\paragraph{Stefan Problem.} 
The most challenging case of the GPME is when the coefficient $k(u)$ is a discontinuous nonlinear step function:
\begin{equation}
k(u) = 
\begin{cases}
k_{\max}, \hspace{0.1cm} u \ge u^{\star} \\
k_{\min}, \hspace{0.1cm} u < u^{\star},
\end{cases}
\label{eqn:stefan_app}
\end{equation}
for given constants $k_{\max}$, $k_{\min}$ and $u^{\star} \in \mathbb{R}$, in which case it is known as the Stefan problem.  
The Stefan problem has been used to model two-phase flow between water and ice, crystal growth, and more complex porous media such as foams \citep{vandermeer2016}.

We conclude by noting that, even though the GPME is nonlinear in general, for specific initial and boundary conditions, it has closed form self-similar solutions.  
For details, see \citet{vazquez2007, maddix2018_harmonic_avg, maddix2018temp_oscill}.
This enables ease of evaluation by comparing each competing method to the ground truth.

\section{Detailed Experiment Settings} 
\label{sec:detailed_experiments}

\begin{figure}[h]
\centering
\vskip 0.1in
\includegraphics[width=0.8\linewidth]{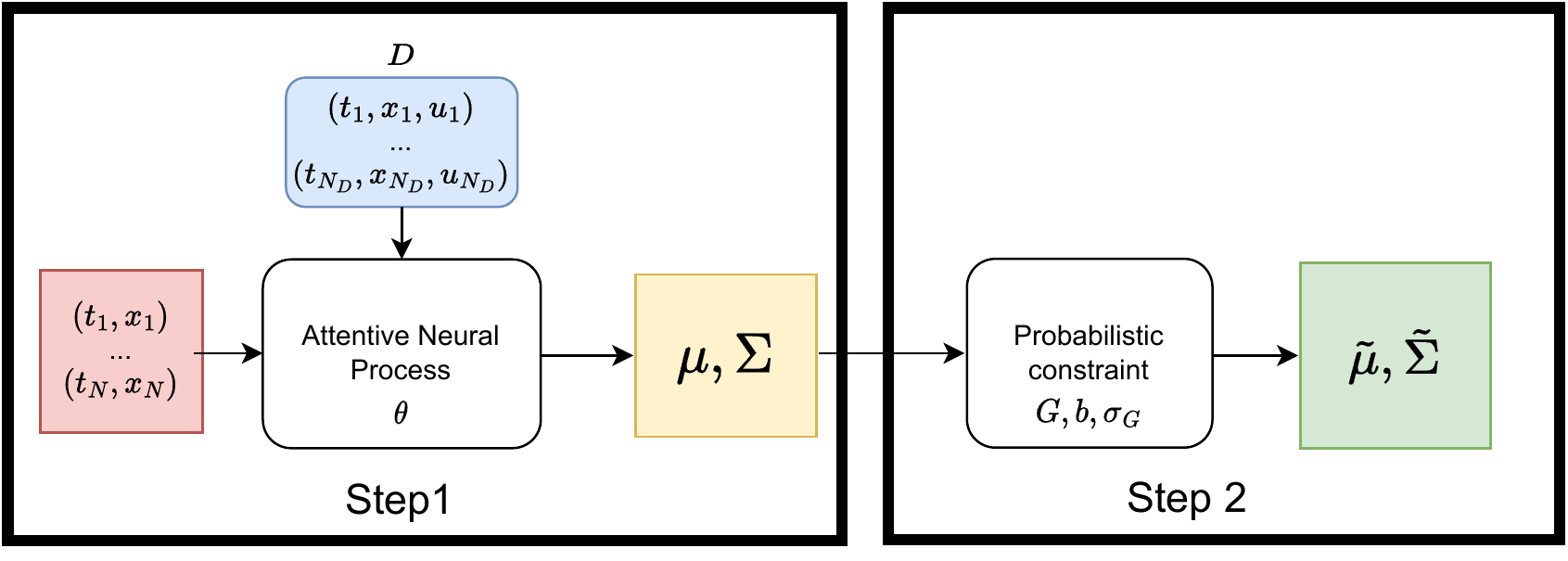}
\caption{Schematic for the instantiation of our framework \probconserv with the \ANP (\physnpnosp) as the data-driven black box model in Step 1  that is used in the empirical results.
In Step~1, the ANP outputs a mean $\mu$ and covariance $\Sigma$ (yellow) of the solution profile $u$ evaluated at the $N$ target points (red). 
The ANP takes as input the context set $D$ that comprises $N_D$ labelled points (blue). 
The parameter $\theta$ encapsulates the neural network weights within the \ANPnosp.
In Step~2, the probabilistic constraint in \autoref{eqn:updated_mean_var} is applied yielding an updated mean $\tilde \mu$ and covariance $\tilde \Sigma$ (green).
The probabilistic constraint is determined by the matrix $G$, value $b$, and variance $\sigma_G^2$ in \autoref{eqn:normal_constraint}.
}
\vskip -0.1in
\label{fig:schematic}
\end{figure}

In this section, we review the basics of the Attentive Neural Process (ANP) \citep{kimAttentiveNeuralProcesses2019} that we use as the black-box deep learning model in Step 1 of our model \physnp in the empirical results \autoref{sec:experiments}. \autoref{fig:schematic} illustrates a schematic for \physnp that shows how in the first step the mean and covariance estimates $\mu, \Sigma$ from the \ANP are fed into our probabilistic constraint in the second step to output the updated mean and covariance estimates $\tilde \mu, \tilde \Sigma$.

\paragraph{Model Training.} 
\label{sec:training}
The model from Step 1 is data-driven, with parameter $\theta$ that needs to be learned from data.
Given an empirical data distribution, written as $(u, b, D) \sim p$,
we maximize the expected joint likelihood of the function $u$ and the constraint $b$, conditioned on data $D$, as a function of the Step 1 parameter $\theta$ and Step 2 parameters $\sigma_G$ and $G$ as follows:
\begin{equation} 
\label{eqn:training_objective}
\begin{split}
     L(\theta, \sigma_G, G) &= \mathbb E_{u, b, D \sim p} \log p (u, b \vert D) \\
     &=\underbrace{\mathbb E_{u, D \sim p}\log p_{\theta} (u \vert D)}_{\text{Step 1}} + \underbrace{\mathbb E_{u, b}\log p_{\sigma_G, G}(b \vert u)}_{\text{Step 2}} .
\end{split}
\end{equation}
This follows because the joint probability can be broken into conditionals ${p(u, b \vert D) = p_\theta(u \vert D) p_{\sigma_G, G} (b \vert u)}$, using Bayes' Rule. The Step 2 constraint only depends on the value $u$.

The Step~1 parameter $\theta$ is only present in the first term of the summation in \autoref{eqn:training_objective}. Then, the optimal value for $\theta^\star$ is found by optimizing the unconstrained log-likelihood from Step 1 over the empirical data distribution and is given as follows:
\begin{equation} \label{eqn:step1_training_objective}
\begin{split}
     \theta^\star &= \argmax_\theta L(\theta, \sigma_G, G) \\ &=
 \argmax_\theta \mathbb E_{u, D \sim p}\log p_{\theta} (u \vert D).
\end{split}
\end{equation}
\autoref{eqn:step1_training_objective} is simply the optimization target of several generative models, e.g., Gaussian processes and the ANP.
This justifies training the Step~1 black-box model with its original training procedure before applying our Step~2.

\paragraph{Data Generation.}
For each PDE instance, we first generate training data for the data-driven model in Step 1.
We generate these samples, indexed by $i$, by randomly sampling $n_{\text{train}}$ values of the PDE parameters $\alpha_i$ from an interval $\mathcal A$. 
To create the input data $D_i$, the solution profile corresponding to $\alpha_i$ is evaluated on a set of $N_D$ points uniformly sampled from the spatiotemporal domain $[0, t] \times \Omega$.
Then, the reference solution for $u$ with parameter $\alpha_i$, denoted $u_i$, is evaluated over another set of $N_{\text{train}}$ uniformly-sampled points.
The Step~1 model (\ANPnosp) is then trained on these supervised input-output pairs, $(D_i, u_i)$.  
Using \autoref{eq:linear_constraint}, the conservation value $b$ in Step 2 is calculated given the parameter $\alpha_i$. 
At inference time, we fix specific values of the PDE parameters $\alpha$ that are of interest and generate new input-output pairs to evaluate the predictive performance.
The settings are the same as those at training, except that the reference solution is evaluated on a fixed grid that evenly divides the time domain $[0, t]$ into $T_{\text{test}}$ points and the spatial domain $\Omega$ into $M_{\text{test}}$ points for a spatio-temporal grid of $N_{\text{test}} = T_{\text{test}} \times M_{\text{test}}$ points.
For consistent results, we repeat this procedure over $n_{\text{test}}$ independent datasets for each $\alpha$.

\autoref{tbl:experiment_domain_app_train} 
provides the training settings and \autoref{tbl:experiment_domain_app_test} provides the cor    responding test settings.
\begin{table}[h]
\centering
 \caption{Training details for each instance of the GPME (Diffusion, PME, Stefan) used in the experiments. }\label{tbl:experiment_domain_app_train}
 \vskip 0.1in
\begin{tabular} {lllllllllllll}
\toprule
PDE & Parameter & $\mathcal A$ & Time domain $[0,t]$ & Spatial domain $\Omega$  &$n_{\text{train}}$ & $N_D$ & $N_{\text{train}}$  \\ 
\midrule
Diffusion & $k$& $[1, 5]$ &$[0, 1]$ & $[0, 2\pi]$ & 10,000 & $100$ & $100$ \\
\midrule
PME & $m$ & $[1, 6]$ & $[0, 1]$ & $[0, 1]$ & 10,000 & $100$ & $100$  \\
\midrule
Stefan & $u^\star$ & $[0.55, 7]$ & $[0, 0.1]$ & $[0, 1]$ & 10,000 & $100$ & $100$\\ 
\bottomrule
\end{tabular}
\vskip -0.1in
\end{table}

\begin{table}[h]
\centering
 \caption{Testing details for each instance of the GPME (Diffusion, PME, Stefan) used in the experiments. 
}\label{tbl:experiment_domain_app_test}
\vskip 0.1in
\begin{tabular} {lllllllllllll}
\toprule
PDE & Parameter values & Test time & Spatial domain $\Omega$  &$n_{\text{test}}$ & $N_D$ &  $T_{\text{test}}$ & $M_{\text{test}}$ &
$N_{\text{test}}$ \\ 
\midrule
Diffusion & $k \in \{1, 5\}$ & $0.5$ & $[0, 2\pi]$ & 50 & $100$ & $201$ & $201$ & $40,401$ \\
\midrule
PME & $m \in \{1, 3, 6\}$ & $0.5$ & $[0, 1]$  & 50 & $100$   & $201$ & $201$ & $40,401$ \\
\midrule
Stefan & $u^\star \in \{0.6\}$ & $0.05$ & $[0, 1]$ & 50 & $100$ & $201$ & $201$ & $40,401$\\ 
\bottomrule
\end{tabular}
\vskip -0.1in
\end{table}
We describe here how the input data $D$; input points $(t_{1}, x_{1}), \dots, (t_{N}, x_{N})$; and solution $u$ are created for a particular draw of PDE parameter $\alpha \in \mathcal A$.
The input data (a.k.a. the context set) $D$ is generated as follows.
First, draw samples from the spatiotemporal domain $(t_n, x_n) \sim \text{Uniform} ([0, t] \times \Omega)$, for $n = 1, \dots, N_D$.
For each sample $(t_n, x_n)$, evaluate the reference solution $u_n \coloneqq u(t_n, x_n)$ for $\alpha$.
Then $D = \{(t_n, x_n, u_n)\}_{n=1, \dots, N_D}$.

We create input points $(t_{1}, x_{1}), \dots, (t_{N}, x_{N})$ differently depending on whether we are training or testing.
At train-time, the input points are sampled uniformly from the spatiotemporal domain \[{(t_n, x_n) \sim \text{Uniform} ([0, t] \times \Omega)},\] for ${n = 1, \dots, N_{\text{train}}}$.
At test-time, we divide up the time domain $[0, t]$ into $T_{\text{test}}$ evenly-spaced points and the spatial domain $\Omega$ into $M_{\text{test}}$ evenly-spaced points.
We then take the cross product of these as the set of input points, whose size is $N_{\text{test}} = T_{\text{test}} \times M_{\text{test}}$.

Finally, over the set of input points, we evaluate the reference solution for $\alpha$ as: $u = [u(t_n, x_n)]_{n=1, \dots, N_{\text{train}}}$.

\paragraph{Attentive Neural Processes (ANP).}
The Attentive Neural Process (ANP) \citep{kimAttentiveNeuralProcesses2019} models the conditional distribution of a function $u$ at target input points $\{x_n\} \coloneqq x_1, \dots, x_N$ for $x_i \in \mathbb R^{D+1}$ given a small set of context points $D \coloneqq \{x_i, u_i\}_{i \in C}$.
The function values at each target point $x_n$, written as $u_n$, are conditionally independent given the latent variable $z$ with the following distribution for $u_n$:
\begin{equation} \label{anp}
\begin{split}
        p_\theta(u_n| D) &= \int_z p_\theta(u_n \vert z, D) p_\theta(z \vert D) dz, \\
        p_\theta(u_n \vert z, D) &= \dnorm(u_n| \mu_n, \sigma^2_n), \\
        p_\theta(z | \mu_z, \Sigma_z) &= \dnorm (z | \mu_z, \Sigma_z), \\
        \mu_n, \sigma_n &= f_\theta^u(x_n, z, f_\theta^r(x_n, D)), \\
        \mu_z, \Sigma_z &= f_\theta^z(D). \\
\end{split}
\end{equation}
Here,  $\dnorm(u | \mu, \sigma^2) \coloneqq (2 \pi \sigma^2)^{-1/2} \exp\left(- \frac 1 {2\sigma^2} (x - \mu)^2\right)$ denotes the univariate normal distribution with mean $\mu$ and variance $\sigma^2$ and
$f_\theta^z$, $f_\theta^u$, and $f_\theta^r$ are neural networks whose architecture is described in more detail~below.

As standard in variational inference, the attentive neural process (ANP) is trained to maximize the evidence lower bound (ELBO), which is a tractable lower bound to the marginal likelihood $\mathbb E_{u, D \sim p}\log p_{\theta} (u \vert D)$ that we want to maximize in \autoref{eqn:step1_training_objective}:
\begin{equation} \label{eqn:anp_elbo}
\begin{split}
             \mathbb E_{u, D \sim p}\log p_{\theta} (u \vert D) &\ge \mathbb E_{u, D\sim p} \mathbb E_{z \sim q_{\phi}} \log p_{\theta} (u, z \vert D) - \log q_{\theta}(z \vert u, D), \\
             q_{\theta}(z \vert u, D) &= \dnorm (z | \mu_z^q, \Sigma_z^q),\\
             \mu_z^q, \Sigma_z^q &= f_\theta^z(D \cup \{(t_1, x_1, u_1), \dots (t_N, x_N, u_N)\}).
\end{split}
\end{equation}
By concatenating the context set $D$ with the target set, the \ANP can use the same networks for both the generative model $p_\theta$ and the variational model $q_\theta$. 
This differs from methods such as the variational auto-encoder (VAE) that train a separate network for the variational model.

In the experiments, we train the ELBO in \autoref{eqn:anp_elbo} using stochastic gradient descent over random mini-batches of the supervised pairs $(u, D)$ and a sample of the latent variable $z$ (using the reparameterization trick for an unbiased gradient estimate).
Specifically, we use the ADAM optimizer with a learning rate of $1 \times 10^{-4}$ and a batch size of $250$. 

\paragraph{Architectural details.} 
Here, we briefly describe the architecture of the ANP used in experiments; a more thorough description of the ANP in general can be found in the original paper \citep{kimAttentiveNeuralProcesses2019}. 

\begin{figure}[H]
    \vskip 0.1in
      \centering
\includegraphics[width=\figsizeapp\linewidth]{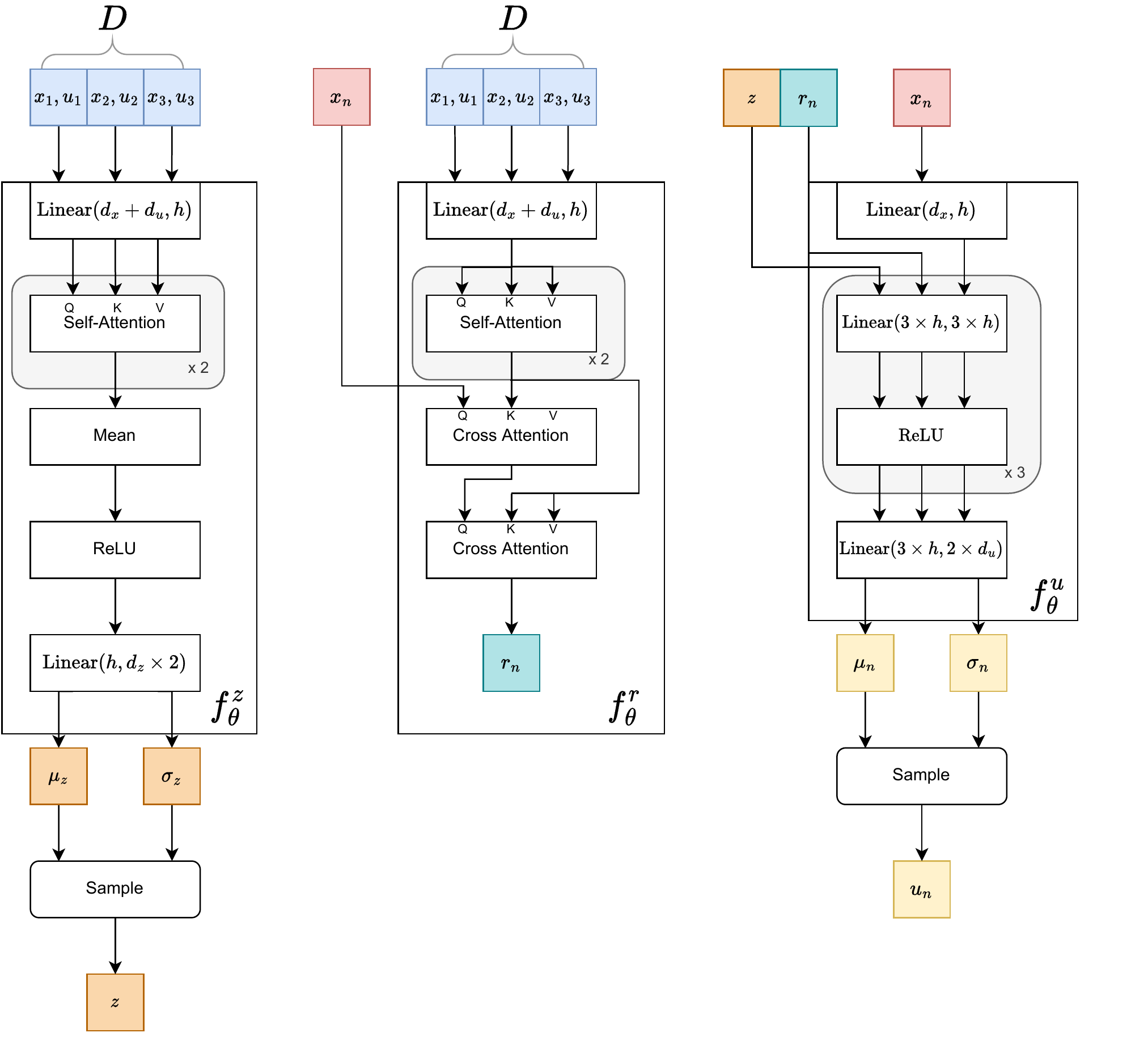}
    \caption{Architectural diagram of the three main networks that make up the Attentive Neural Process (ANP) from \citet{kimAttentiveNeuralProcesses2019} that is used in the experiments as the Step~1 black-box model.} 
    \vskip -0.1in
\label{fig:anp_architecture}
\end{figure}

The ANP consists of three distinct networks:
\begin{enumerate}

    \item The \textit{latent encoder} $f_\theta^z$ takes the context set $D = \{x_i, u_i\}_{i \in C}$ as input and outputs a mean $\mu_z$ and diagonal covariance $\Sigma_z$ for the latent representation $z$. Note that $f_\theta^z$ is invariant to the order of the context set inputs in $D$.
    \item The \textit{deterministic encoder} $f_\theta^r$ takes the context set $D = \{x_i, u_i\}_{i \in C}$ and the target points $\{x_n\}$ as input, and outputs a set of deterministic representations $\{r_n\}$ corresponding to each target point. Note that $f_\theta^r$ is permutation-invariant to the order of the context set inputs in $D$, and is applied pointwise across the target inputs $\{x_n\}$.
    \item The \textit{decoder} $f_\theta^u$ takes the outputs from the latent encoder, deterministic encoder, and the target points $\{x_n\}$ as input, and outputs a set of mean and variances $\{\mu_n, \sigma_n\}$ corresponding to each target point. The decoder is applied pointwise across the target inputs $\{x_n\}$ and deterministic representation $\{r_n\}$.
\end{enumerate}
\begin{table}[H]
    \centering
    \caption{ANP hyperparameters.}
\label{tab:anp_hyperparameters}
    \vskip 0.1in
    \begin{tabular}{lll}
         Symbol & Value & Description    \\
         \midrule
         $d_x$ & 2 & Input dimension \\
         $d_u$ & 1 & Output dimension \\
         $d_z$ & 128 & Latent dimension \\
         $h$ & 128 & Size of hidden layer \\
         $n_{\text{heads}}$ & 4 & Number of heads in MultiHead \\
         $d_h$ & 128 & Column dimension in MultiHead layers \\
    \end{tabular}
    \vskip -0.1in
\end{table}
For reproducibility, \autoref{fig:anp_architecture} shows how each network is constructed and Table \ref{tab:anp_hyperparameters} shows the ANP hyperparameters.
Each building blocks is also briefly described below:
\begin{itemize}
    \item $\text{Linear}(d_{\text{in}}, d_\text{out})$: dense linear layer $xA + b$.
    \item $\text{Mean}$: Averages the inputs of the input set; i.e., $\text{Mean} (\{s_i\}) = \frac 1 {|\{s_i\}|} \sum_i s_i$.
    \item $\text{ReLU}$: Applies ReLU activation pointwise.
    \item Cross-Attention and Self-Attention. These are multi-head attention blocks first introduced in \citet{vaswani2017attention}. The three inputs to the multi-head attention block are the queries $Q=[q_1 | \dots | q_{d_q}]^\top$, keys $K=[k_1 | \dots | q_{d_k}]^\top$, and values $V=[v_1 | \dots | v_{d_k}]^\top$. The hyperparameters are the number of heads, $n_{\text{heads}}$ and the number of columns of the matrices $W_i^Q, W_i^K, W_i^V$, denoted as $d_h$. We summarize the notations below:
        \[
          \begin{split}
          \text{Self-Attention}(Q) &\coloneqq \text{MultiHead}(Q, Q, Q), \\
          \text{Cross-Attention}(Q, K, V), &\coloneqq \text{MultiHead}(Q, K, V), \\
            \text{MultiHead}(Q, K, V) &\coloneqq [H_1 | \dots | H_{n_{\text{heads}}}] W^O, \\
            H_i &\coloneqq \text{Attention}(QW_i^Q, KW_i^K, VW_i^V), \\
            \mathrm{Attention}(Q, K, V) &\coloneqq \mathrm{softmax} \left( \frac{QK^\top}{\sqrt{d_k}} \right)V, \\
            \mathrm{softmax} \left( \begin{bmatrix} x_{1,1} & \dots & x_{1, n} \\ \vdots & \ddots & \vdots \\ x_{m, 1} & \dots & x_{m,n} \end{bmatrix} \right) &\coloneqq \begin{bmatrix} \frac{\exp(x_{1,1})}{\sum_{i=j}^n \exp(x_{1, j})} & \dots & \frac{\exp(x_{1, n})}{\sum_{j=1}^n\exp(x_{1, j})} \\ \vdots & \ddots & \vdots \\ \frac{\exp(x_{m, 1})}{\sum_{i=1}^m \exp(x_{m, j})} & \dots & \frac{\exp(x_{m,n})}{\sum_{j=1}^n \exp(x_{m, j})} \end{bmatrix}.\\
        \end{split}
        \]
\end{itemize}
\vspace{-.1cm}

\section{Additional Empirical Results}
\label{sxn:app-additional-empirical-results}

In this section, we provide additional empirical results for the degenerate parabolic Generalized Porous Medium (GPME) family of conservation laws as well as for hyperbolic conservation laws.

\subsection{GPME Family of Conservation Laws}
Here, we include additional solution profiles and conservation profiles over time for the GPME family of equations, ranging from the ``easy" diffusion (heat), ``medium" PME, to the ``hard" Stefan equations.
\subsubsection{Diffusion (Heat) Equation}
\label{app:heat_extra}
\begin{figure}[H]
\centering
\includegraphics[width=0.85\linewidth]{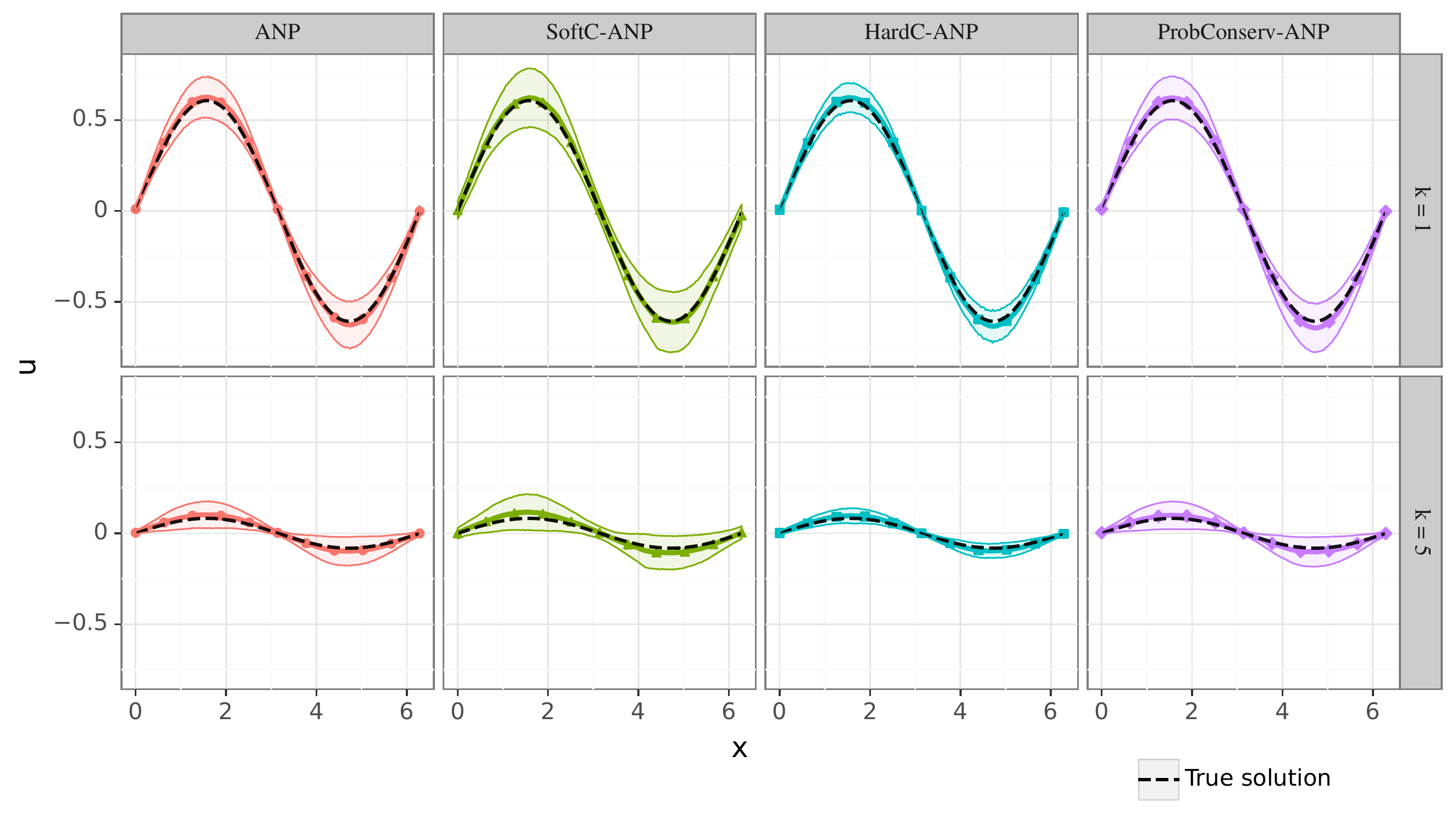}
    \caption{
    Solution profiles for the diffusion (heat) equation at time $t=0.5$ for diffusivity (conductivity) test-time parameter $k=1$ in the top row and $k=5$ in the bottom row. 
    Each model is trained on samples of $k \in \mathcal A = [1,5]$.
    The shaded region illustrates $\pm 3$ standard deviation uncertainty intervals. \physnp and \HCANP both display tighter uncertainty bounds than the baseline \ANPnosp, while \SCANP is more diffuse. The uncertainty is relatively homoscedastic on this ``easy'' case.}
    \label{fig:heat_time_plot}
\end{figure}

\paragraph{Solution Profiles.} \autoref{fig:heat_time_plot} shows the solution profiles for the ``easy" diffusion equation, at time $t=0.5$, where a sine curve is damped over time for test-time parameter $k=1, 5 \in \mathcal{A} = [1, 5]$. \autoref{tab:heat_vary_k_5} shows the corresponding metrics. 

\begin{table}[h]
\caption{
     Mean and standard error for CE $\times 10^{-3}$ (should be zero), LL (higher is better) and MSE $\times 10^{-4}$ (lower is better) over $n_{\text{test}}=50$ for the (``easy") diffusion equation at time $t=0.5$ with variable diffusivity constant $k$ parameter in the range $\mathcal{A}=[1,5]$ and test-time parameter values $k=1,5$.
     }
     \label{tab:heat_vary_k_5}
     \vskip 0.1in
    \centering
    \begin{tabular}{c|lll|lll}
        & $k=1$       &                        &                         & $k=5$  \\
        & CE          & LL                     & MSE                     & CE         & LL                     & MSE                       \\
\midrule
\ANP    & 4.68 (0.10)        & $2.72$ (0.02)          & $1.71$ (0.41)           & 1.76 (0.04)       & $3.28$ (0.02)          & $0.547$ (0.08)            \\
\SCANP  & 3.47 (0.17)       & $2.40$ (0.02)          & $2.24$ (0.78)           & 2.86  (0.05)     & $2.83$ (0.02)          & $1.75$ (0.24)             \\
\HCANP  & \textbf{0} (0.00) & $\textbf{3.08}$ (0.04) &  $\mathbf{1.37}$ (0.33) & \textbf{0} (0.00) & $\textbf{3.64}$ (0.03) & $\mathbf{0.461}$ (0.07)   \\
\physnp & \textbf{0} (0.00) & \textbf{2.74} (0.02)   & \textbf{1.55} (0.33)    & \textbf{0} (0.00) & \textbf{3.30} (0.02)   & \textbf{0.485} (0.07)     \\
    \end{tabular}
    \vskip -0.1in
\end{table}

\subsubsection{Porous Medium Equation (PME)}
\paragraph{Results for Different $\lambda$ for \SCANPnosp.} \label{app:scanp_lambda}
As is the case with PINNs \citep{Raissi19}, the \SCANP method has a hyper-parameter $\lambda$ that controls the balance in the training loss between the reconstruction and differential term.
A higher value of $\lambda$ places more emphasis on the residual of the PDE term and less emphasis on the evidence lower bound (ELBO) from the \ANPnosp.

To investigate whether tuning $\lambda$ will lead to significantly different results, we report results for different values of $\lambda$ for the \SCANP on the Porous Medium Equation (PME).
Since these results are presented on the same test dataset used in \autoref{tab:pme_vary_m}, it provides an optimistic case on how tuning $\lambda$ could improve the results for \SCANPnosp.
\autoref{tab:pinns_hpo} shows that the predictive performance is roughly the same across different values of $\lambda$, with both MSE and LL worse than the original ANP across the board and the conservation error (CE) $G\mu - b$ at the final time worse for $m=6$.

\begin{table}[h!]
\centering
\caption{Investigation of the effect of the soft constraint penalty parameter $\lambda$ in the \SCANP baseline. The metrics CE $\times 10^{-3}$ (should be zero), LL (higher is better) and MSE $\times 10^{-4}$ (lower is better) are reported for the (``medium") PME at time $t=0.5$ with variable $m$ parameter in the range $\mathcal{A}=[0.99,6]$ and test-time parameters $m \in \{1,3,6\}$. We see that the performance is not significantly changed as a function of $\lambda$, and, surprisingly, that the unconstrained \ANP ($\lambda=0)$ performs better in most metrics than \SCANPnosp.}
\vskip 0.1in
\begin{small}
\begin{tabular}{l|lll|lll|lll}
      &        $m=1$  & & &     $m=3$ & & &$m=6$ \\
          & CE &LL & MSE & CE & LL & MSE & CE & LL & MSE \\
\midrule
\ANP $(\lambda = 0)$     & $6.67$ & $\bf{3.49}$ &$\bf{0.94}$  & $-1.23$ & $\bf{3.67}$ & $\bf{1.90}$  & $\bf{-2.58}$ & $\bf{3.81}$& $\bf{7.62}$ \\
\midrule
\SCANP $(\lambda=0.01)$ & $5.58$ & $3.11$ & $1.11$ & $-0.61$ &  $3.46$ & $2.03$  & $-3.00$ & $3.49$ & $7.76$   \\
\SCANP $(\lambda=0.1)$ & $5.58$ & $3.11$ & $1.11$  & $-0.67$  & $3.46$ & $2.07$  & $-3.01$ &  $3.49$ & $7.87$  \\
\SCANP $(\lambda=1)$ & $5.62$ & $3.11$ & $1.11$ &  $-0.65$ & $3.46$ & $2.06$  & $-3.03$ & $3.49$ & $7.82$   \\
\SCANP $(\lambda=10)$ & $\bf{5.52}$ &  $3.11$ & $1.08$  & $\bf{-0.56}$ & $3.46$ & $2.04$  & $-3.02$ & $3.49$ &  $7.76$  \\
\SCANP $(\lambda=100)$ & $5.62$ & $3.11$ & $1.11$  & $-0.59$ & $3.46$ & $2.03$  & $-3.03$ & $3.49$ & $7.69$  \\
\end{tabular}
\end{small}
\vskip -0.1in
\label{tab:pinns_hpo}
\end{table}

\begin{figure}[H]
\vskip 0.1in
\centering 
\includegraphics[width=\figsizepme\linewidth]{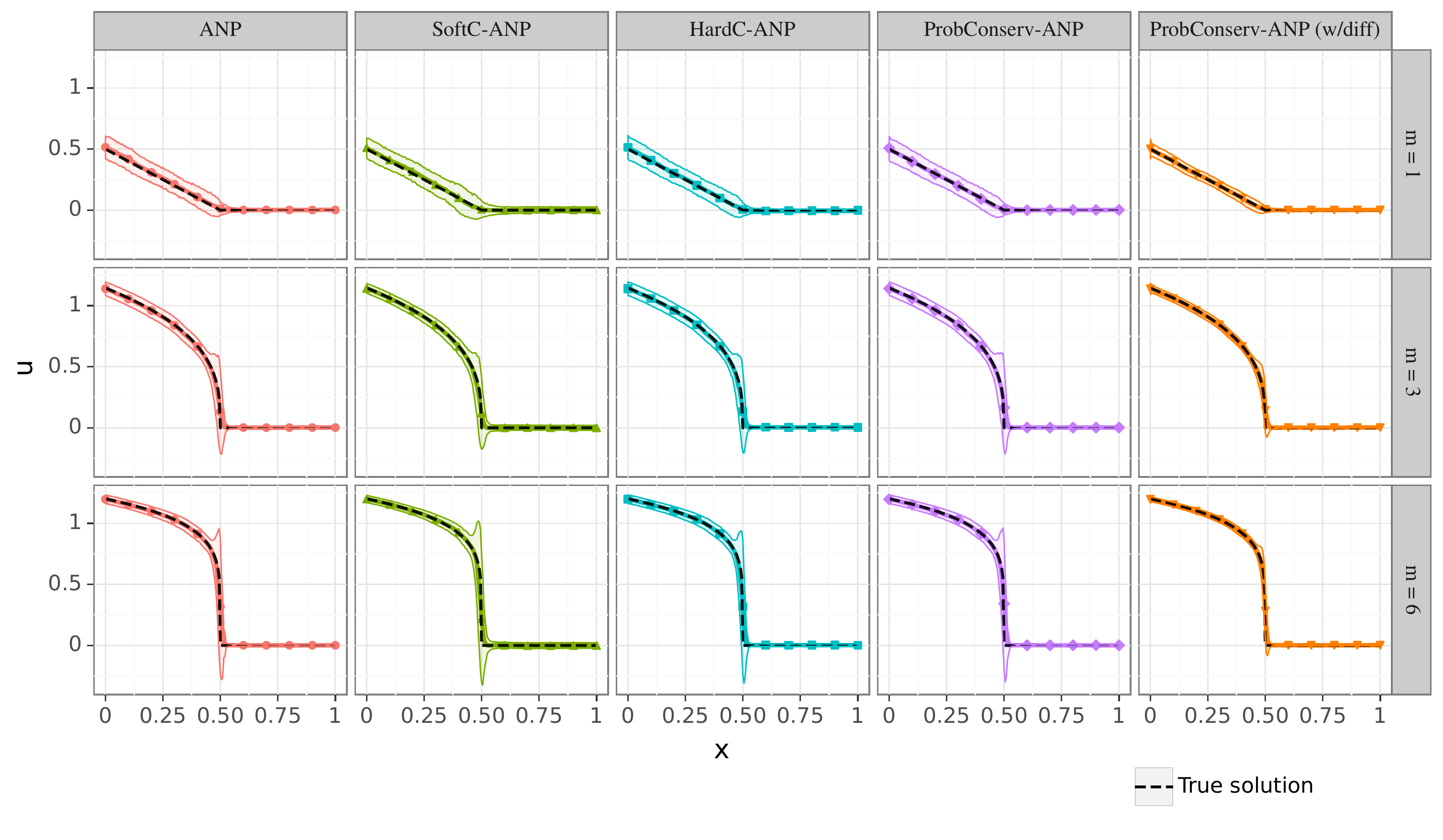}
    \caption{Solution profiles and uncertainty intervals for the PME predicted by our \physnp and other baselines. The solutions are obtained for three scenarios with increasing sharpness in the profile as $m$ is increased from $m=1$ to $m=6$ from left to right, respectively. The \HCANP model, which assumes constant variance for the whole domain, results in too high uncertainty in the zero (degenerate) region, unlike our proposed \physnp approach that incorporates the variance information to effectively handle this heteroscedasticity.  Adding diffusion to \physnp removes the oscillations locally at the degeneracy, as desired.}
    \label{fig:pme_solution_profile_app}
    \vskip -0.1in
\end{figure}

\begin{figure}[H]
\vskip 0.1in
\begin{center}
\includegraphics[width=\figsizepme\linewidth]{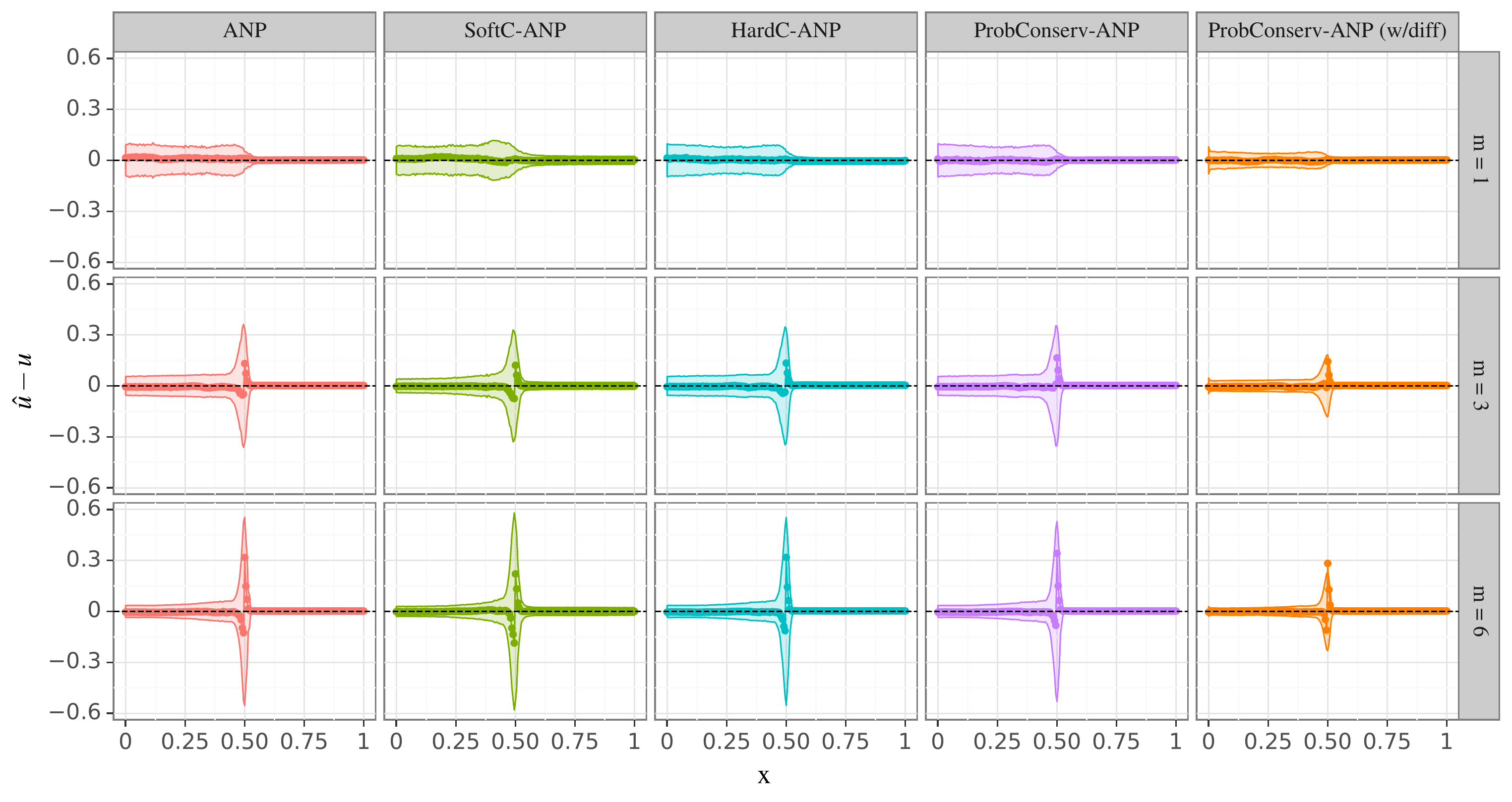}
\end{center}
\caption{Solution errors with uncertainty bounds as a function of $x$ for our \physnp and other baselines for the PME with parameter $m\in \{1, 3, 6\}$ after training on $m \in \mathcal A = [0.99, 6]$.
The shaded region indicates $\pm 3$ standard deviations as estimated by each model. 
For $m=1$, both \physnp with and without diffusion result in solutions with smaller errors. While \HCANP model reduces the error scale, it underestimates the zero portion of the solution, which is nonphysical, as the solution quantity cannot be negative.
For $m\in\{3, 6\}$, while the error magnitude becomes dominant at the shock position for all methods, \physnp with diffusion provides the lowest errors with the tightest confidence interval.}
\label{fig:pme_error}
\vskip -0.1in
\end{figure}

\paragraph{\physnp with Diffusion.}
As described in \autoref{app:artificial_diff}, we explore adding numerical diffusion for eliminating artificial small-scale noises when enforcing conservation. \autoref{tab:pme_vary_m_app} shows that adding artificial diffusion
improves both MSE and LL compared to the conservation constraint alone.
Figures \ref{fig:pme_solution_profile_app}-\ref{fig:pme_error} illustrate that by removing such artificial noises, \physnp with diffusion leads to tighter uncertainty bounds as well as higher LL than the other baselines. 

\begin{table*}[h]
    \centering
     \caption{Mean and standard error for CE $\times 10^{-3}$ (should be zero),  LL (higher is better) and MSE $\times 10^{-4}$ (lower is better) over $n_{\text{test}}=50$ runs  for the (``medium'') PME at time $t=0.5$ with variable $m$ parameter in the range $\mathcal{A} = [0.99, 6]$.  
     We see that \physnp (w/diff) improves the performance on \physnp by applying smoothing at the sharp boundary as the test-time parameter $m$ is increased.}
       \vskip 0.1in
    \resizebox{1\linewidth}{!}{
    \begin{tabular}{c|lll|lll|lll}
      &        $m=1$  & & &     $m=3$ & & &$m=6$ \\
                      & CE         & LL                   & MSE                  & CE         & LL                   & MSE                  & CE         & LL                   & MSE \\                   
\midrule
\ANP                  & $6.67$ (0.39)     & $3.49$ (0.01)        & $0.94$ (0.09)        & $-1.23$ (0.29)   & $3.67$  (0.00)       & $1.90$ (0.04)        & $-2.58$ (0.23)    & $3.81$ (0.01)        & \textbf{7.67} (0.09) \\         
\SCANP                & $5.62$ (0.35)    & $3.11$ (0.01)        & $1.11$ (0.14)        & $-0.65$ (0.30)   & $3.46$  (0.00)       & $2.06$ (0.03)        & $-3.03$ (0.26)    & $3.49$ (0.00)        & $7.82$ (0.09)    \\      
\HCANP                & \textbf{0} (0.00) & 3.16  (0.04)         & 0.43   (0.04)        & \textbf{0} (0.00) & $3.44$  (0.03)       & 1.86 (0.03) & \textbf{0} (0.00)  & 3.40 (0.05)          & \textbf{7.61} (0.09)   \\
\physnp               & \textbf{0} (0.00)  & 3.56 (0.01) & \textbf{0.17} (0.02) & \textbf{0} (0.00)  & 3.68 (0.00) & 2.10 (0.07)          & \textbf{0} (0.00) & 3.83 (0.01) & 10.4 (0.04)  \\      
 \physnp (w/diff) & \textbf{0} (0.00) 
 & \textbf{4.04} (0.02) & \textbf{0.15} (0.02) & \textbf{0} (0.00)  & \textbf{3.96} (0.00) & \textbf{1.43} (0.05) & \textbf{0} (0.00)  & \textbf{4.03} (0.01) & 7.91 (0.03)  \\ 
    \end{tabular}
}
     \label{tab:pme_vary_m_app}
     \vskip -0.1in
\end{table*}

\paragraph{Solution and Error Profiles.} Figures \ref{fig:pme_solution_profile_app}-\ref{fig:pme_error} illustrate the differing solution profiles and errors for the PME for various values of $m\in\{1,3,6\}$, respectively. As expected, we see a gradient for $m > 1$ that becomes sharper and approaches infinity for $m=6$.  Increasing $m$ results in smaller values of the PDE parameter denoting the pressure $k(u) = u^m$, which increases the degeneracy for smaller values of $k(u)$, i.e., larger values of $m$. In this case the problem also becomes more challenging.  For $m=1$, we have a piecewise linear solution, and for $m=3,6$ we see sharper oscillatory uncertainty bounds at the front or free boundary, resulting in some negative values at this boundary as well. We see the value of the uncertainty quantification to reflect that the model is certain in the parabolic regions to the left and right of the sharp boundary especially in the zero (degeneracy) region, and is most uncertain at the boundary (degeneracy) point. 

\subsubsection{Stefan}
\begin{figure}[H]
    \centering
    \vskip 0.1in \includegraphics[width=0.7\linewidth]{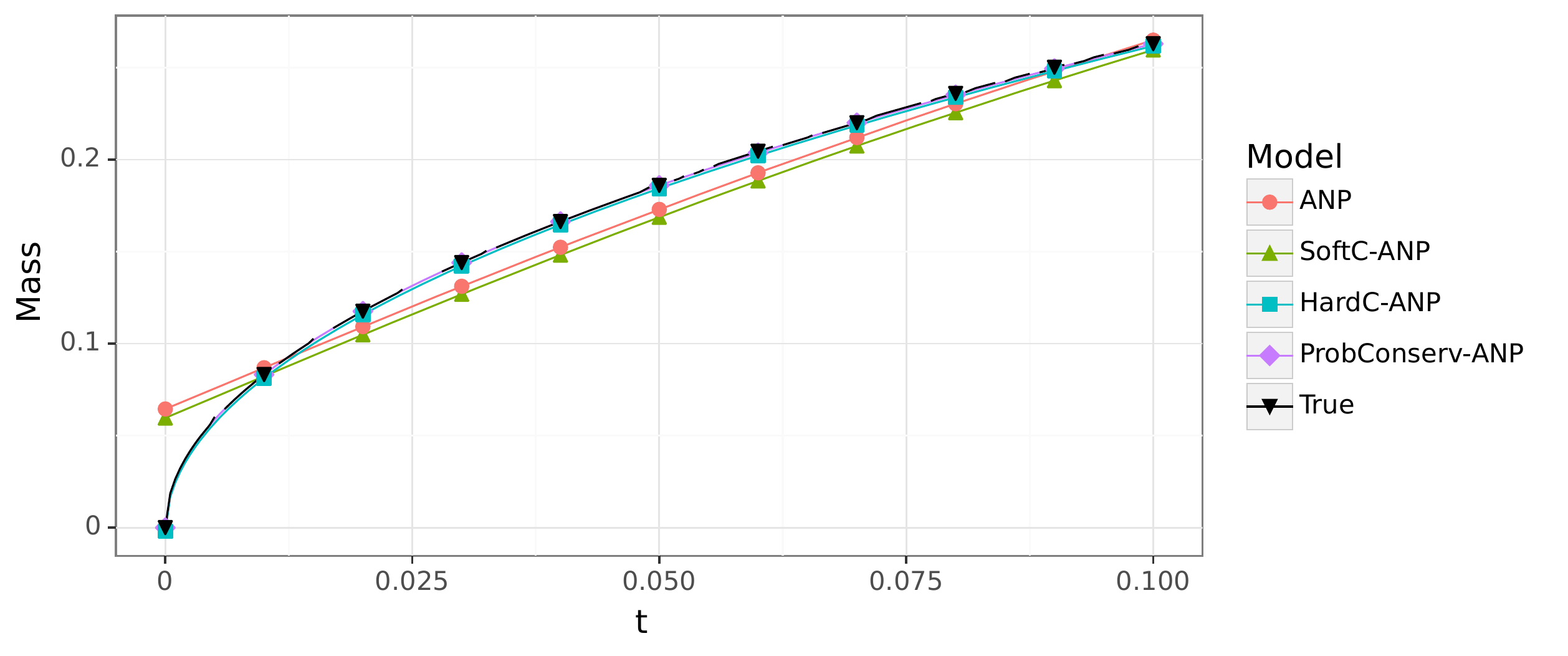}
    \caption{True mass over time for each model. The true mass conservation profile over time is matched exactly by our proposed method \physnp and the hard-constrained \HCANP by design. The unconstrained \ANP and surprisingly even the differential form soft-constrained \SCANP have a non-physical linear mass profile over time.}
    \label{fig:stefan_cons}
    \vskip -0.1in 
\end{figure}
\label{app:stefan_extra}
 \autoref{fig:stefan_cons} shows \physnp follows the true profile of conserved mass in the system over time by design. We also see that the unconstrained \ANP and surprisingly the soft-constrained \SCANP that applied the differential form as a soft constraint does not result in conservation being satisfied since it does not enforce it exactly.  For these baselines, the mass profile over time is linear and does not match the true profile which is proportional to $\sqrt{t}$.

\subsection{Hyperbolic Equations}
Here, we demonstrate that our approach \physnp also works for hyperbolic conservation laws by considering the linear advection problem (``medium'') and Burgers' equation (``hard''), which are both introduced in \autoref{tbl:classification_pde_laws} in \autoref{app:exact_sol}. 

\subsubsection{Linear Advection}
\begin{figure}[H]
     \centering
     \vskip 0.1in \includegraphics[width=0.6\linewidth]{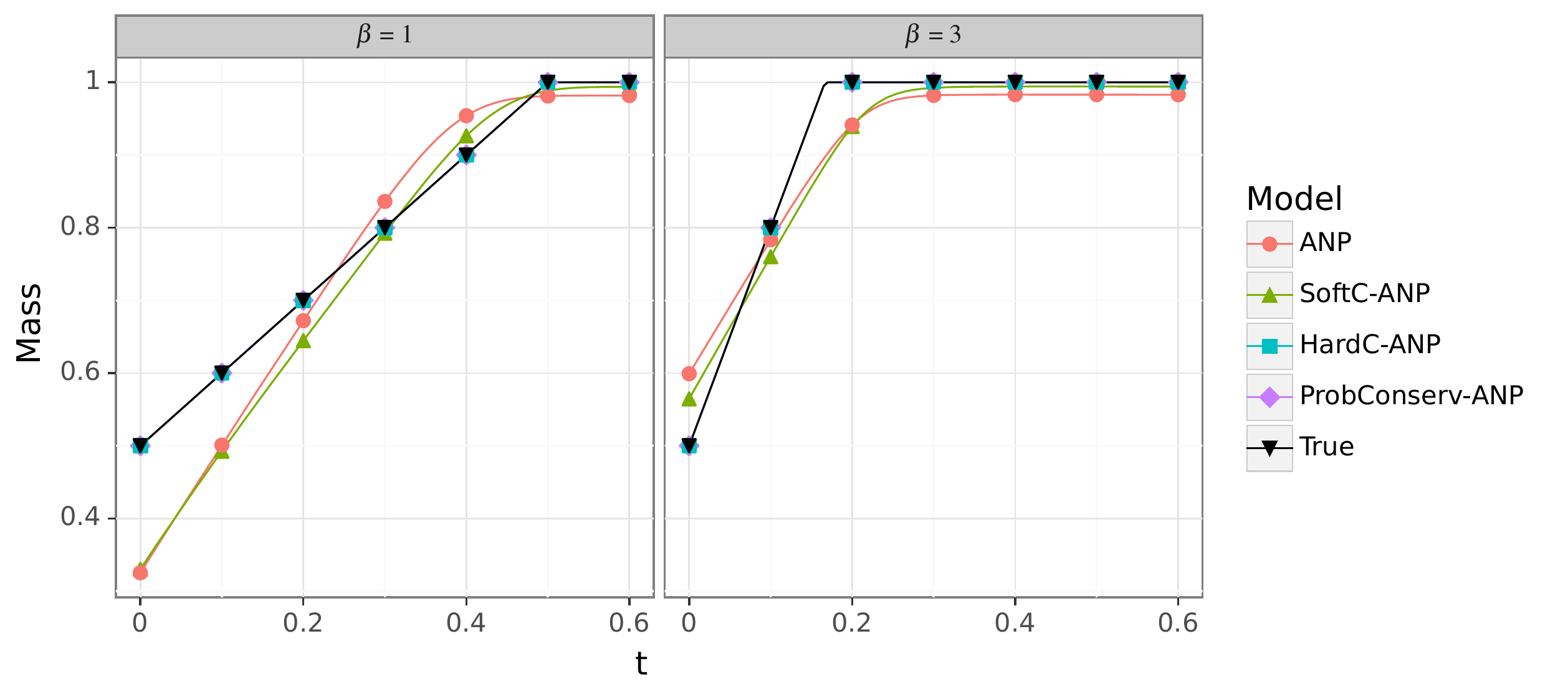}
     \caption{System total mass as a function of time $t$ for the linear advection problem with test-time parameter $\beta=1, 3$ and training parameter range $\mathcal{A} = [1,5]$. Both \physnp and \HCANP satisfy conservation of mass while the unconstrained \ANP and soft-constrained \SCANP baselines deviate from the actual trend completely at all times.}
    \label{fig:linear_advection_cons}
     \vskip -0.1in 
 \end{figure}
 \autoref{fig:linear_advection_cons} displays the system total mass, $U(t)=\int_\Omega u(t,x) d\Omega$ as a function of time, obtained by our \physnp model and the other baselines and compared against the true curve. The results are obtained for two values of test-time parameter $\beta=1, 3$ denoting the velocity with training range $\beta \in \mathcal{A} = [1,5]$. The unconstrained \ANP contradicts the system true mass at all times including $t=0$. By proper incorporation of the conservation constraint, both \physnp and \HCANP methods are able to predict the system mass and capture the actual trend over time exactly while the soft-constrained differential form \SCANP baseline results in little improvement.

\begin{figure}[h]
     \centering
     \vskip 0.1in \includegraphics[width=\figsizeadvection\linewidth]{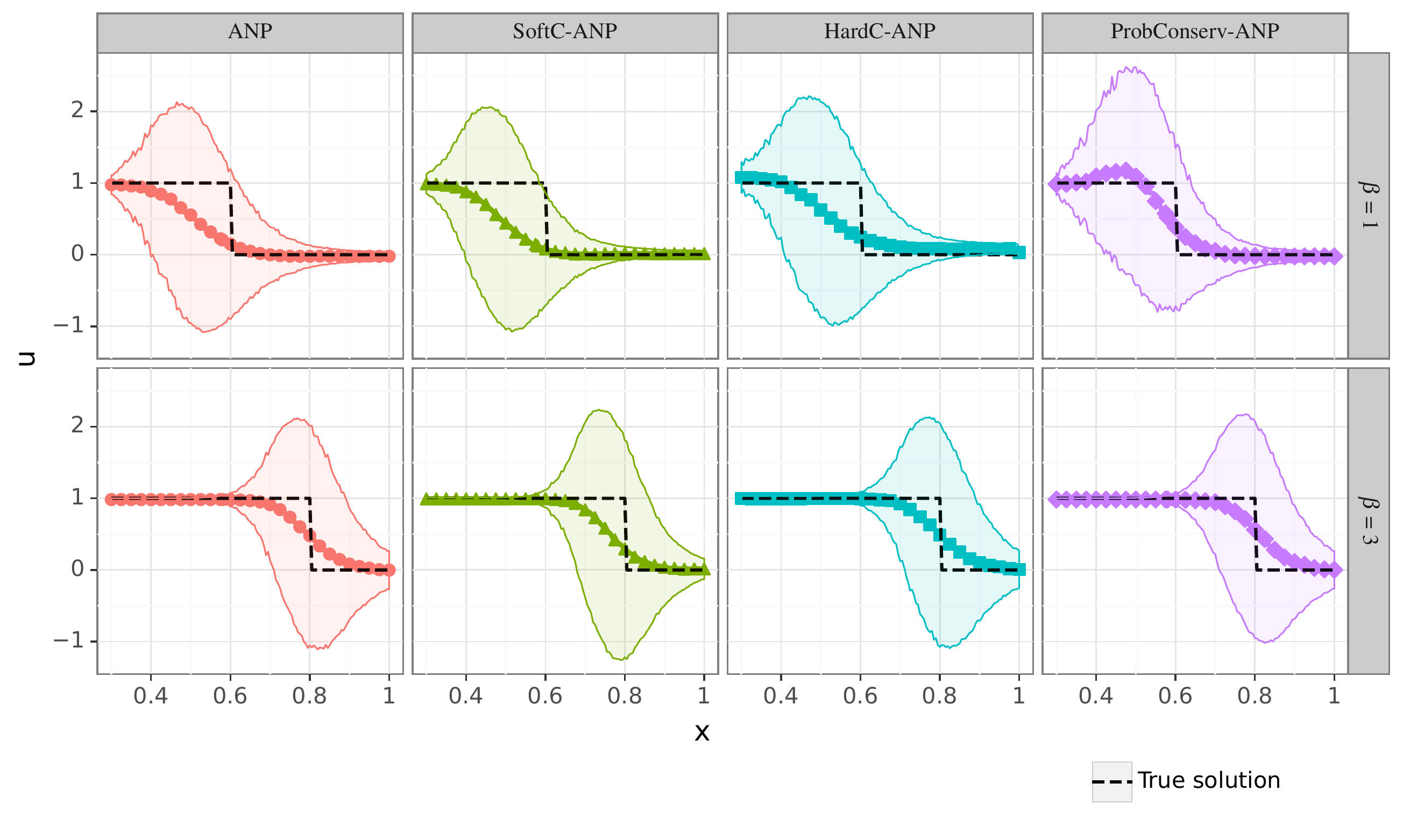}
     \caption{Solution profiles and uncertainty intervals for linear advection problem at time $t=0.1$ for test-time parameter $\beta=1, 3$ and training parameter range $\mathcal{A} = [1,5]$.  Despite satisfaction of conservation constraint, \HCANP  predicts a highly diffusive profile and remarkable underestimation of shock interface region especially for $\beta=1$. The prediction error is even higher for the unconstrained \ANP model which does not enforce the conservation, and the shock interface is shifted further away from the true solution. \physnp results in a sharper profile than other the baselines and the predicted shock interface is around the actual shock position leading to more accurate shock position estimation.}
     \vskip -0.1in \label{fig:linear_advection_solution}
 \end{figure}

 \begin{figure}[H]
     \centering
     \vskip 0.1in \includegraphics[width=\figsizeadvection\linewidth]{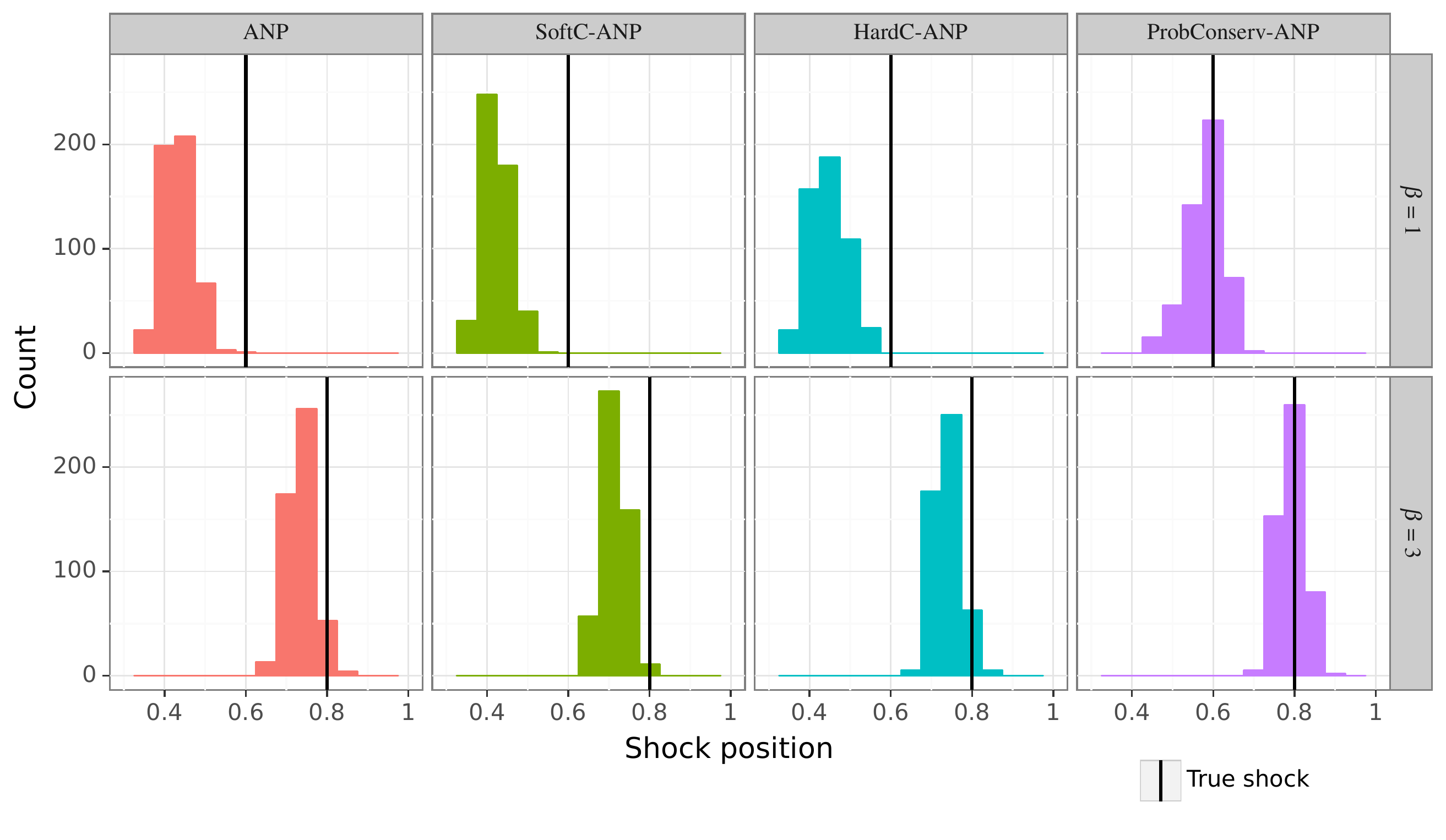}
     \caption{The histogram of shock position for the linear advection problem, computed as the mean plus or minus 3 standard deviations.  Due to the shift in the shock interface, both the \ANP and \HCANP models underestimate the position of the shock, and the underestimation is more significant for $\beta=1$. The \physnp model provides a histogram distributed almost symmetrically around the true shock interface and thus leads to an accurate estimate of the shock position.}
     \vskip -0.1in \label{fig:linear_advection_shock}
 \end{figure}

 \autoref{fig:linear_advection_solution} shows the predicted solution profiles and corresponding uncertainty intervals for time $t=0.1$ and test-time parameter $\beta=1, 3$. Our \physnp model predicts sharper shock profile centered around the actual shock position. On the contrary, both \ANP and \HCANP  lead to highly diffusive profiles which are shifted toward the left of actual shock interface, leading to the under-estimation of the shock position on this downstream task. This under-estimation becomes more evident in \autoref{fig:linear_advection_shock}, which indicates the corresponding histograms of shock position. The histograms associated with the \ANP and \HCANP models are skewed to the left and both result in the averaged shock positions which are lower than the actual value depicted by the solid vertical line. By proper leveraging of our finite volume based physical constraint, our \physnp results in proper uncertainty quantification which leads to accurate prediction of shock location compared to the other baseline models. \autoref{tab:linear_advection} also shows this accuracy improvement with a maximum improvement of $2.86\times$ in MSE for $\beta=1$.
 
 \begin{table}[h]
\caption{
     Mean and standard error for CE (should be zero),  LL (higher is better) and MSE $\times 10^{-2}$ (lower is better) over $n_{\text{test}}=50$ runs  for the hyperbolic linear advection problem at time $t=0.1$ with variable $\beta$ parameter in the range $\mathcal{A} = [1, 5]$.
     }
     \label{tab:linear_advection}
     \vskip 0.1in
\resizebox{1\linewidth}{!}{
    \centering
    \begin{tabular}{c|lll|lll}
        & $\beta=1$       &                        &                         & $\beta=3$  \\
        & CE          & LL                     & MSE                     & CE         & LL                     & MSE                       \\
\midrule
\ANP   & -0.136 (0.004) & 0.96 (0.01) & 5.72 (0.25) & 0.042 (0.003) & 0.51 (0.01) & 2.03 (0.01) \\
\SCANP & -0.137 (0.004) & \textbf{1.58} (0.03) & 7.64 (0.34) & 0.013 (0.003) & \textbf{2.31} (0.02) & 2.87 (0.20) \\
\HCANP & \textbf{0} (0.00) & -2.96 (0.34) & 4.59 (0.17) & \textbf{0} (0.00) & 1.34 (0.21) & 1.93 (0.07) \\
\physnp & \textbf{0} (0.00) & 1.06 (0.01) & \textbf{2.00} (0.06) & \textbf{0} (0.00) & 0.52 (0.01) & \textbf{1.62} (0.01) \\
    \end{tabular}
   }
    \vskip -0.1in
\end{table}

\subsubsection{Burgers' Equation}
\begin{figure}[H]
     \centering
     \vskip 0.1in \includegraphics[width=0.75\linewidth]{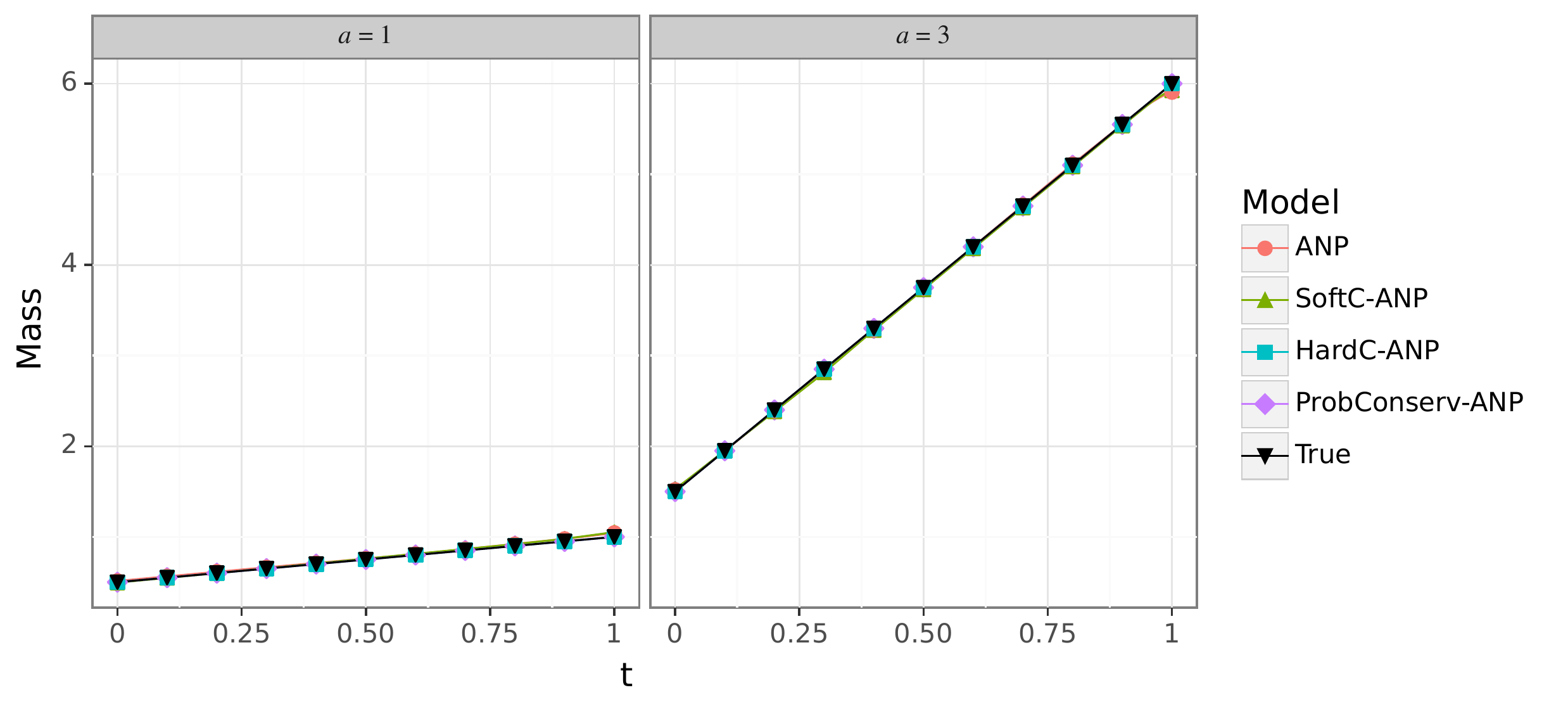}
     \caption{True mass as a function of time $t$ for the Burgers' equation with test-time parameter $a=1, 3$ and training parameter range $\mathcal{A} = [1,4]$.}
     \label{fig:burgers_cons}
     \vskip -0.1in 
 \end{figure}

\begin{figure}[h]
     \centering
     \vskip 0.1in \includegraphics[width=0.8\linewidth]{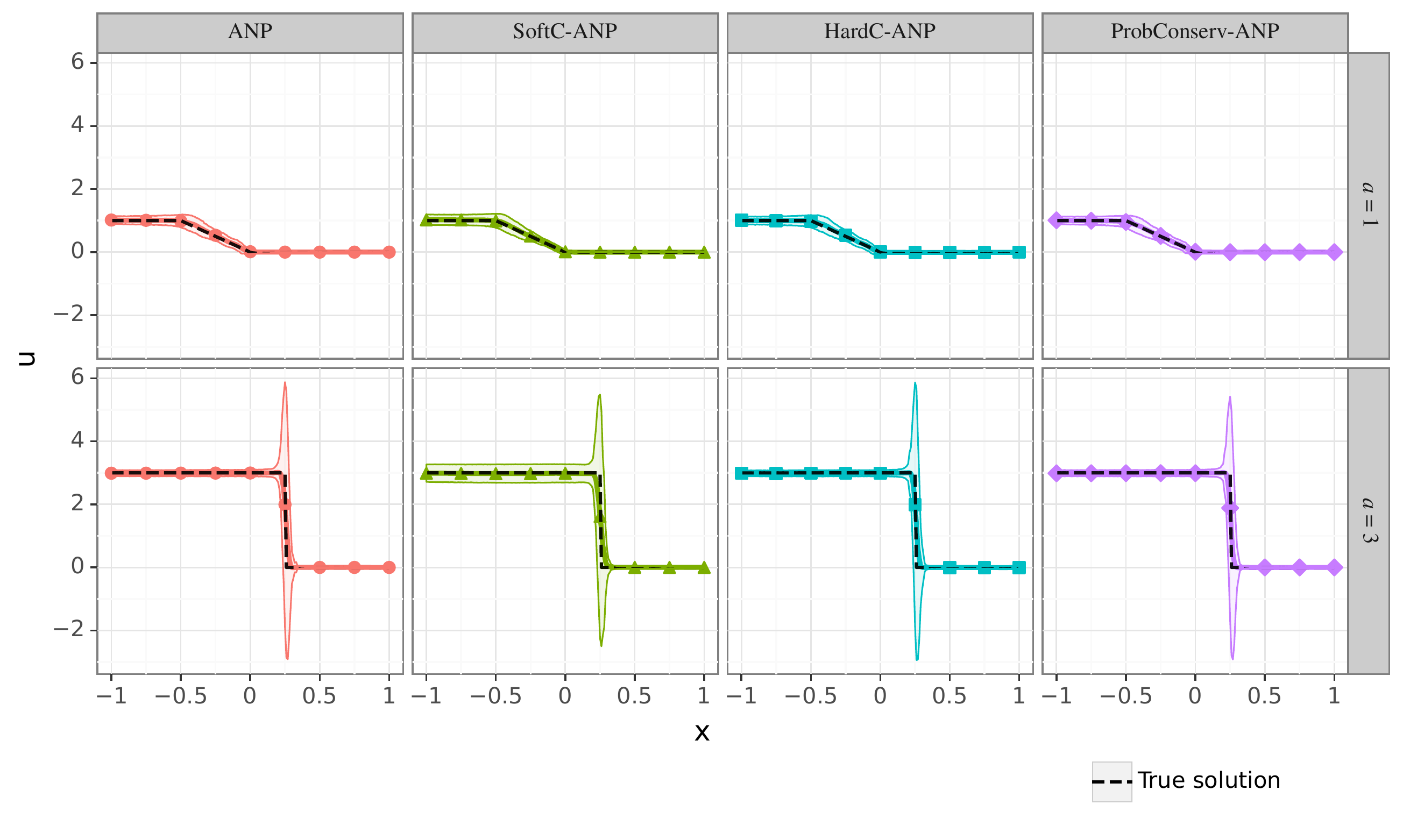}
     \caption{Solution profiles and uncertainty intervals for Burgers' equation at time $t=0.5$ for test-time parameter $a=1, 3$ and training parameter range $\mathcal{A} = [1,4]$.}
     \vskip -0.1in \label{fig:burgers_solution}
 \end{figure}

\autoref{fig:burgers_cons} illustrates that the total mass is linear over time, and in this case is approximately satisfied by our \physnp and the baselines. \autoref{fig:burgers_solution} shows the waiting time phenomenon, where the piecewise linear initial condition self-sharpens until the breaking time $t_b=1/a$, where it forms a rightward moving shock. We see that the breaking time is inversely proportional to the slope, and that the shock forms sooner for larger values of $a$.

\end{document}